%% file: hoda.tex
\RequirePackage[2020-02-02]{latexrelease}
\documentclass[twocolumn, switch]{article} 

\usepackage{preprint}

\usepackage{amsmath, amsthm, amssymb, amsfonts}

\usepackage[numbers,square]{natbib}
\usepackage[utf8]{inputenc}	
\usepackage[T1]{fontenc}	
\usepackage{xcolor}		
\usepackage[colorlinks = true,
            linkcolor = purple,
            urlcolor  = blue,
            citecolor = cyan,
            anchorcolor = black]{hyperref}	
\usepackage{booktabs} 		
\usepackage{nicefrac}		
\usepackage{microtype}		
\usepackage{lineno}		
\usepackage{float}			

\usepackage{microtype}
\usepackage{graphicx}
\usepackage{booktabs} 

\usepackage{hyperref}

\usepackage{pgfplots}
\usepackage{tikz}
\usepgfplotslibrary{groupplots}
\usepgfplotslibrary{fillbetween}
\pgfplotsset{compat=newest}
\pgfplotsset{width=7.5cm,compat=1.12}

\usepackage{amsfonts}
\usepackage{amsmath}
\usepackage{graphicx}
\usepackage{grffile}
\usepackage[export]{adjustbox}
\usepackage{epstopdf} 

\usetikzlibrary{plotmarks}
\usepackage{amsmath}
\usepackage{caption}
\usepackage{subcaption}
\usepackage{url}                
\usepackage[]{hyperref}         
\hypersetup{                    
	colorlinks,
	linkcolor={green!80!black},
	citecolor={red!70!black},
	urlcolor={blue!70!black}
}

\usepackage{filecontents}
\usepackage{multirow}

\usepackage{hyperref}
\usepackage{url}

\usepackage{algorithm,algpseudocode}

\usepackage{lipsum}
\newcommand{\changecolor}{black} 

\newcommand{\comment}[1]{}

\usepackage{newfloat}
\DeclareFloatingEnvironment[name={Supplementary Figure}]{suppfigure}
\usepackage{sidecap}
\sidecaptionvpos{figure}{c}

\usepackage{titlesec}
\titlespacing\section{0pt}{12pt plus 3pt minus 3pt}{1pt plus 1pt minus 1pt}
\titlespacing\subsection{0pt}{10pt plus 3pt minus 3pt}{1pt plus 1pt minus 1pt}
\titlespacing\subsubsection{0pt}{8pt plus 3pt minus 3pt}{1pt plus 1pt minus 1pt}

\usepackage{tikz,xcolor,hyperref}

\definecolor{lime}{HTML}{A6CE39}
\DeclareRobustCommand{\orcidicon}{
	\begin{tikzpicture}
	\draw[lime, fill=lime] (0,0) 
	circle [radius=0.16] 
	node[white] {{\fontfamily{qag}\selectfont \tiny ID}};
	\draw[white, fill=white] (-0.0625,0.095) 
	circle [radius=0.007];
	\end{tikzpicture}
	\hspace{-2mm}
}
\foreach \x in {A, ..., Z}{\expandafter\xdef\csname orcid\x\endcsname{\noexpand\href{https://orcid.org/\csname orcidauthor\x\endcsname}
			{\noexpand\orcidicon}}
}

\title{HODA: Hardness-Oriented Detection of Model Extraction Attacks}

\usepackage{xwatermark}

\usepackage{authblk}

\author[1]{Amir~Mahdi~Sadeghzadeh}
\author[2]{Amir~Mohammad~Sobhanian}
\author[3]{Faezeh~Dehghan}
\author[4]{Rasool~Jalili}

\affil[1,2,3,4]{Data and Network Security Lab (DNSL), Department of Computer Enginnering, Sharif University of Technology, Iran.}
\affil[$ $]{E-mail:  \{amsadeghzadeh, amsobhanian, dehghanniri\}@ce.sharif.edu; jalili@sharif.edu.}

\begin{document}

\twocolumn[ 
  \begin{@twocolumnfalse} 
  
\maketitle

\begin{abstract}
Model Extraction attacks exploit the target model's prediction API to create a surrogate model in order to steal or reconnoiter the functionality of the target model in the black-box setting. Several recent studies have shown that a data-limited adversary who has no or limited access to the samples from the target model's training data distribution can use synthesis or semantically similar samples to conduct model extraction attacks. 
In this paper, we define the hardness degree of a sample using the concept of learning difficulty. The hardness degree of a sample depends on the epoch number that the predicted label of that sample converges.
We investigate the hardness degree of samples and demonstrate that the hardness degree histogram of a data-limited adversary's sample sequences is distinguishable from the hardness degree histogram of benign users' samples sequences. We propose Hardness-Oriented Detection Approach (HODA) to detect the sample sequences of model extraction attacks. The results demonstrate that HODA can detect the sample sequences of model extraction attacks with a high success rate by only monitoring 100 samples of them, and it outperforms all previous model extraction detection methods.
\end{abstract}
\vspace{0.35cm}

  \end{@twocolumnfalse} 
] 



\section{Introduction}
\input{content/intro.tex}

\section{Model Extraction Attacks}
\input{content/mea.tex}

\section{Related Work}

\input{content/inrelatedwork.tex}

\section{Our Proposal: Hardness-Oriented detection Approach}
\subsection{Hardness Degree of Samples}
\label{sec:hsbase}
\input{content/hs.tex}

\subsection{Model Extraction Attacks Setup}
\input{content/attackanalysis.tex}

\subsection{Hardness of Model Extraction Attack Samples}
\input{content/hbdm.tex}

\section{Setup and Evaluation}
\label{sec:seteva}
\input{content/evaluation.tex}

\subsection{Transfer Learning}

\input{content/transferlearning.tex}

\section{Discussion on Adaptive Adversary}

\input{content/duscussion.tex}

\section{Conclusions}
\input{content/conc.tex}



\normalsize
\bibliography{references}

\newpage
\appendix

\section{Datasets}
\input{content/datasets.tex}
\section{Relationship Between the Accuracy of Classifiers and Hardness Degree of Samples}
\label{secmisrate}

\input{content/conf.tex}

\section{Hardness Transferability}
\input{content/classifiers.tex}
\section{Details of Model Extraction Attacks}
\input{content/detailattack.tex}

\section{DenseNet121 Target Classifiers }
\input{content/secattackhistdensenet.tex}
\section{Visualization of CIFAR100 samples and Their Hardness }
\label{sec:viscifar100}
\input{content/viscifar100.tex}
\section{HODA Algorithm}
\input{content/hodaalgorithm.tex}

\section{Pearson Distance Histogram}
\label{secpearsondisthist}
\input{content/disthist.tex}

\section{The Accuracy of Surrogate Classifiers for defended target models}
\input{content/accdefsur.tex}

\section{Adversarial Examples (AEs)}
\input{content/interae.tex}

\section{Performance Analysis of Model Extraction Attacks}

\input{content/hardnessanalysisofattacks.tex}

\section{Comprehensive Related Work}
\input{content/rw.tex}


\end{document}

%% file: content/intro.tex
Deep Neural Networks (DNNs) have shown impressive performance in various tasks in recent years that have encouraged the industry to deploy DNN-based models in a variety of real-world applications. Since the training process of DNNs and collecting training data is an expensive and tedious process, models are considered the intellectual property of organizations, and they must be kept secure. Therefore, models are often securely deployed on cloud servers, and only the creators can access the model parameters. Users are only allowed to query the model via a prediction API and receive predictions.
Recent studies \cite{DBLP:conf/uss/TramerZJRR16,10.1145/3052973.3053009,DBLP:conf/eurosp/JuutiSMA19,DBLP:conf/cvpr/OrekondySF19,DBLP:conf/uss/JagielskiCBKP20} demonstrate that an adversary can exploit the prediction API of a target model to create a surrogate model in order to steal or reconnoiter the functionality of the target model. Such attacks are called \textit{model extraction attacks}, and they violate the intellectual property of model owners. Furthermore,  the surrogate model can be leveraged to conduct other attacks on the target model in black-box setting, such as adversarial example attacks \cite{10.1145/3052973.3053009,DBLP:conf/eurosp/JuutiSMA19} or membership inference attacks \cite{7958568}. 

Most of the model extraction attacks use the target model's prediction API to label an unlabeled dataset to create the surrogate model's training set. 
In most real-world settings, the adversary has no or limited access to samples from the target model's training data distribution, which is called \textit{normal} or in-distribution samples. Hence, most proposed attacks in the previous studies use some form of Out-Of-Distribution (OOD) samples, such as synthesis \cite{10.1145/3052973.3053009,DBLP:conf/eurosp/JuutiSMA19} or semantically similar samples to the target model's training set \cite{DBLP:conf/cvpr/OrekondySF19,pal2019framework} to conduct model extraction attacks. We focus on such attacks in this paper. 
There are two main approaches to defend against model extraction attacks, manipulating the target model outputs to prevent adversary from producing high-quality surrogate model \cite{DBLP:conf/sp/LeeEMS19,DBLP:conf/iclr/OrekondySF20,DBLP:conf/cvpr/KariyappaQ20,kariyappa2021protecting} and detecting the sample sequences of model extraction attacks \cite{10.1145/3274694.3274740,DBLP:conf/eurosp/JuutiSMA19,DBLP:journals/corr/abs-2107-05166,10.1145/3474369.3486863}. 
We propose Hardness-Oriented Detection Approach (HODA), a new approach to detect sample sequences of model extraction attacks, which outperforms PRADA \cite{DBLP:conf/eurosp/JuutiSMA19} and VarDetect \cite{DBLP:journals/corr/abs-2107-05166} by a large margin and it is more scalable.

\textcolor{\changecolor}{In this paper, we use the concept of learning difficulty \cite{DBLP:conf/iclr/TonevaSCTBG19,DBLP:conf/icml/HacohenCW20} to define our hardness measure. We expect the predicted label of easy samples to converge sooner than the predicted label of hard samples during training. Hence, the hardness Degree of a sample depends on the predicted label convergence speed for that sample during training.
We consider a DNN-based classifier training process as a sequence of subclassifiers so that each subclassifier is created at the end of an epoch. HODA uses a subsequence of subclassifiers to compute the hardness degree of samples.
Since attack samples do not lie on the data manifold that is well supported by the target classifier training data \cite{DBLP:conf/icml/JiangZTM21}, attack samples have a very small number of easy samples, unlike normal samples.
Normal samples are independent and identically distributed (iid) data from the target classifier's training data distribution.}
We demonstrate that the hardness degree histogram of benign user's samples is distinguishable from the hardness degree histogram of model extraction attack samples. 
HODA uses this observation to detect sample sequences of model extraction attacks. 
For each user, HODA computes the distance between the hardness degree histograms of that user's samples and normal samples, and if the distance exceeds a threshold, the user is detected as an adversary.
HODA can detect JBDA \cite{10.1145/3052973.3053009}, JBRAND \cite{DBLP:conf/eurosp/JuutiSMA19}, and Knockoff Net \cite{DBLP:conf/cvpr/OrekondySF19} attacks with a high success rate by only monitoring 100 samples of attack. We demonstrate that HODA is also highly effective when the target classifier is trained using transfer learning. 

\textbf{Contributions}. 
(i) We demonstrate that the hardness degree of a sample for a classifier pertains to the training data distribution of that classifier.
(ii) We indicate that the hardness degree histogram of normal samples is distinct from the hardness degree histograms of model extraction attack samples.
(iii) We propose HODA to detect the sample sequences of model extraction attacks.

%% file: content/mea.tex
The model extraction attack is one of the most serious threats against machine learning-based classifiers on remote servers, such as Machine Learning as a Service (MLaaS).
	 The adversary's goal is to create a surrogate classifier $f_{s}$ that imitates target classifier $f_{t}$ on task $T$. 
	 Most model extraction attacks exploit target model $f_t$ to label unlabeled samples to create the surrogate model's training set. The adversary sends sample $x_i$ to the target model and receives its output $f_t(x_i)$, and then she uses pair $(x_i,f_t(x_i))$ to train surrogate classifier $f_s$.
	 The output type of target model can be label, label confidence, top-k values in probability vector, or the entire probability vector. We only consider label $\bar{f_t}(x_i)$ and the entire probability vector $f_t(x_i)$ as the output type of target classifiers in our experiments.
	 There are two primary intents for adversaries to conduct model extraction attacks, \textit{stealing} and \textit{reconnaissance}. 

\textbf{Stealing}:
Producing a high performance classifier is an expensive and time-consuming process and requires computational resources and experts. Therefore, adversaries are motivated to take advantage of a target classifier to reduce the cost of creating a new classifier. 
The adversary's goal in stealing is to maximize the accuracy of surrogate model on data distribution $\mathcal{D}_T$. Hence, the adversary's goal is:
\begin{equation}
\text{Maximize} \quad P_{(x,y)\sim \mathcal{D}_T} \bar{f_{s}}(x) = y
\end{equation}

\textbf{Reconnaissance}: The model extraction attacks can be used to conduct other attacks in the black-box setting, such as adversarial example attacks \cite{10.1145/3052973.3053009,DBLP:journals/corr/GoodfellowSS14} and membership inference attacks \cite{7958568}. The adversary's goal in reconnaissance is to maximize the \textit{fidelity} among surrogate and target classifiers in order to increase the success rate of black-box attacks. Similar to \cite{DBLP:conf/uss/JagielskiCBKP20}, we consider label agreement among surrogate and target classifiers as the fidelity metric on data distribution $\mathcal{D}_T$. Hence, the adversary's goal is:
\begin{equation}
\text{Maximize}\quad P_{(x,y) \sim \mathcal{D}_T} \bar{f_s}(x) = \bar{f_t}(x)
\end{equation} 


Proposed model Extraction attacks create the surrogate classifier training set $\mathbb{X}_s=\{(x_i,f_t(x_i))\}_{i=1}^B$ by various methods, where $B$ is the attack budget. The attack budget determines the number of samples that an adversary is allowed to send to the target classifier and receive their associated predictions.
After creating $\mathbb{X}_s$, the adversary trains surrogate classifier $f_s$ to minimize empirical loss on $\mathbb{X}_s$. We suppose that the adversary knows the architecture and hyperparameters of the target classifier and uses them to train the surrogate classifier. It is noteworthy that our proposed defense is independent of surrogate classifiers' training process.  

%% file: content/inrelatedwork.tex
\textbf{Model Extraction Attacks}:
Primary model extraction attacks try to extract the exact value of parameters \cite{10.1145/1081870.1081950,DBLP:conf/uss/TramerZJRR16} and hyperparameters \cite{DBLP:conf/sp/WangG18} of shallow models. In recent years, the proposed attacks mainly aimed to steal or reconnoiter the functionality of deep neural networks by querying them in the black-box setting. It is often sensibly assumed in the literature that the adversary has no or limited access to samples from the training set distribution of target classifier. In order to overcome this issue, attacks use some form of out-of-distribution samples, such as synthesis or semantically similar samples, to create the surrogate classifier's training set. Knockoff Net \cite{DBLP:conf/cvpr/OrekondySF19}, ActiveThief \cite{pal2019framework}, and Copycat CNN \cite{DBLP:conf/ijcnn/SilvaBBSO18} use semantically similar datasets to the target model’s training set to train a surrogate classifier. In another line of studies, \cite{10.1145/3052973.3053009,DBLP:conf/eurosp/JuutiSMA19,DBLP:conf/ndss/YuYZTHJ20,Truong_2021_CVPR,kariyappa2020maze,NEURIPS2020_e8d66338} use synthetic data to create the surrogate classifier's training set.

\textbf{Model Extraction Defenses}:
Existing defense methods against model extraction attacks generally distribute into two branches: perturbation-based and detection-based. 
Perturbation-based defenses \cite{DBLP:conf/sp/LeeEMS19,DBLP:conf/iclr/OrekondySF20,DBLP:conf/cvpr/KariyappaQ20} attempt to prevent adversaries from producing high-quality surrogate classifiers by adding perturbation to the target classifier outputs. Recently, \citet{kariyappa2021protecting} proposed a new defense with the same goal as perturbation-based defenses, which does not perturb the target classifier outputs. Their approach employs an ensemble of diverse models to produce discontinuous predictions for OOD samples.
Detection-based defenses attempt to detect the occurrence of model extraction attacks by observing successive input queries to the target classifier. \citet{10.1145/3274694.3274740} present a method to detect extraction attacks against Decision Tree models. PRADA \cite{DBLP:conf/eurosp/JuutiSMA19} is the first proposed detection-based defense for DNN models. 
\citet{10.1007/978-3-030-62144-5_4} demonstrate that several OOD detection approaches, such as Baseline \cite{DBLP:conf/iclr/HendrycksG17} and ODIN \cite{DBLP:conf/iclr/LiangLS18}, have poor performance in detecting Knockoff Net attack samples. Hence, they propose a new OOD detection approach that use a detector to detect OOD samples. However, the OOD detector is trained on samples from the same distribution used by the adversary to conduct Knockoff Net attacks, which is an unrealistic assumption in practice.  SEAT \cite{10.1145/3474369.3486863} aims to detect model extraction attacks that use several similar samples to extract a target model, such as jacobian-based attacks \cite{10.1145/3052973.3053009,DBLP:conf/eurosp/JuutiSMA19}. Hence,  SEAT is ineffective when an adversary uses natural samples that are not similar to each other, such as Knockoff Net attack. 
VarDetect \cite{DBLP:journals/corr/abs-2107-05166} uses Variational Autoencoders (VAs) and Maximum Mean Discrepancy (MMD) to detect model extraction attacks. 
HODA can detect both jacobian-based and Knockoff Net attacks, and it performs well on high-dimensional datasets, such as Caltech256 and CUB200. Furthermore, unlike other work \cite{10.1007/978-3-030-62144-5_4,DBLP:conf/cvpr/KariyappaQ20,kariyappa2021protecting}, HODA only needs access to in-distribution samples to detect model extraction attacks.

\textbf{Sample Hardness}: 
The sample hardness has attracted attention from several machine learning domains, such as curriculum learning, identifying important or most informative examples, and detecting mislabeled samples \cite{DBLP:conf/iclr/TonevaSCTBG19,carlini2019prototypical,DBLP:conf/icml/JiangZTM21}. 
We review only the most relevant work to our study.
\citet{DBLP:conf/icml/HacohenCW20} and \citet{Mangalam2019DoDN} show that DNNs learn samples that are learnable by shallow models in early epochs of training before learning harder ones.
The work of  \citet{DBLP:conf/icml/HacohenCW20} that inspired our definition of hardness demonstrates that DNNs learn samples in both training and test sets in a similar order.  \citet{DBLP:journals/corr/abs-2105-08997} discuss the correlation between learning order of samples with image statistics like segment count, edge strengths, image intensity entropy, and DCT coefficient matrix. \textcolor{\changecolor}{\citet{carlini2019prototypical} propose several measures for identifying \textit{prototypical samples} representing the behavior to be learned. They show prototypical samples are easy to learn, and training on hard samples can improve accuracy on many datasets and tasks. \citet{DBLP:conf/iclr/TonevaSCTBG19} investigate the learning dynamics of neural networks and define a \textit{forgetting event} to have occurred when a training sample transitions from being classified correctly to incorrectly over the course of learning. They show hard samples are forgotten with higher frequency than easy samples. Two works, \citet{DBLP:conf/icml/JiangZTM21} and \citet{DBLP:journals/corr/abs-2106-09647}, are particularly relevant to our work. \citet{DBLP:conf/icml/JiangZTM21} introduce the \textit{consistency score} that measures the structural consistency of an example with the underlying data distribution $\mathcal{P}_{data}$. They demonstrate that easy samples have a higher consistency score, which means they lie in a region on the data manifold that is well supported by other regular instances. The authors examine several proxies for consistency score and indicate, in contrast to distance-based proxies, learning-speed-based proxies correlate very well with the consistency score. We demonstrate most model extraction attack samples are hard, which means that they are not well supported by the samples in the target model's training data. In independent and concurrent work, \citet{DBLP:journals/corr/abs-2106-09647} introduce two measures of sample hardness, \textit{prediction depth} and \textit{learning difficulty}. The learning difficulty measure is the same as our measure of hardness, except that we use epoch rather than iteration. The authors demonstrate a negative correlation between prediction depth and learning difficulty. They show that samples learned in later epochs have higher prediction depth and confirm that neural networks learn easy samples first. }

%% file: content/hs.tex
\input{content/figure/hardnesshist.tex}

We use the concept of learning difficulty \cite{DBLP:conf/iclr/TonevaSCTBG19,DBLP:conf/icml/HacohenCW20} to define our hardness measure.
We expect that the predicted label of easy samples converges in early epochs and the predicted label of hard samples converges in the later epochs during training.
The training process of a DNN-based classifier can be considered a sequence of subclassifiers so that each subclassifier is created at the end of an epoch. Suppose that classifier $f_t$ is trained for $m$ epochs. The training process of $f_t$ can be represented as the following sequence of subclassifiers:
\begin{equation}
F = <f_t^0,f_t^1, f_t^2, ..., f_t^{m-1}>
\end{equation}
where subclassifier $f_t^i$ is created at the end of the $i^{th}$ epoch. 
To compute the hardness degree of samples, we select a \textit{subsequence} of $F$ called $F_{subclf}$. For example, if $m=100$, $F_{subclf}$ can be $<f_t^{19},f_t^{39},f_t^{59},f_t^{79},f_t^{99}>$.
The \textbf{\textit{hardness degree}} of sample $x_i$ is $h$ if the $h^{th}$ subclassifier in $F_{subclf}$ is the first subclassifier that the predicted label of all subsequent subclassifier in $F_{subclf}$ is equal to its predicted label. Therefore, the hardness degree of sample $x_i$ for classifier $f_t$, which is displayed by $\phi_{f_t}(x_i)$  is defined as follows:
\begin{equation}
\begin{split}
  &\phi_{f_t}(x_i) = h \\
  \text{s.t.} \;  \forall \; j > h, \; &\bar{F}_{subclf}[h](x_i) = \bar{F}_{subclf}[j](x_i) ,\;  \\
   &\bar{F}_{subclf}[h](x_i) \neq \bar{F}_{subclf}[h-1](x_i).
 \end{split}
\end{equation}
where $\bar{F}_{subclf}[k](x_i)$ is the predicted label by $k^{th}$ subclassifier in $F_{subclf}$ for sample $x_i$. It is supposed that $\bar{F}_{subclf}[-1](x_i) = \emptyset$.
Based on the hardness degree definition, the hardness degree of a sample is in the range $[0,|F_{subclf}|-1]$, where $|F_{subclf}|$ is the size of $F_{subclf}$. 
Since we want to calculate the hardness degree of samples at \textbf{\textit{inference time}} for target model users' samples, we need to save subclassifiers in $F_{subclf}$ during training to use them at inference time. When a new sample arrives, it is fed to all loaded subclassifiers in $F_{subclf}$, and using their predictions, the hardness degree of that sample is computed. It is important to note that we do not use the true label of samples to calculate their hardness degree. Algorithm \ref{alg:hd} in Appendix \ref{sec:hodaalg} describes how the hardness of samples is computed using $F_{subclf}$ at inference time.

\begin{table}
	\caption{The accuracy of classifiers on CIFAR10 and CIFAR100 test sets.}
	\renewcommand{\arraystretch}{1.3}
	\vspace{0\baselineskip}
	\label{tab:clsacc}
	\centering
	\resizebox{0.6\linewidth}{!}{
		\begin{tabular}{cccc}
			\hline
			& \multicolumn{3}{c}{Acc(\%)}        \\ \cline{2-4} 
			& ResNet18 & DenseNet121 & MobileNet \\ \cline{2-4} 
			CIFAR10  & 94.36    & 94.92       & 93.59     \\
			CIFAR100 & 76.38    & 77.57       & 73.47     \\ \hline
		\end{tabular}
	}
\end{table} 

\begin{figure}
	\begin{subfigure}{\linewidth}
	\resizebox{\linewidth}{!}{
	\begin{tabular}{cccccccccc}
		Airplane & Automobile & Bird & Cat & Deer & Dog & Frog & Horse & Ship & Truck \\
{\includegraphics[width = 0.5\linewidth,valign=m]{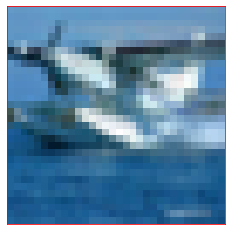}} &
{\includegraphics[width = 0.5\linewidth,valign=m]{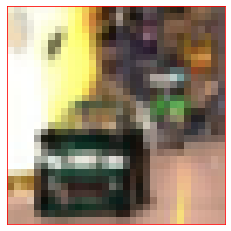}} &
{\includegraphics[width = 0.5\linewidth,valign=m]{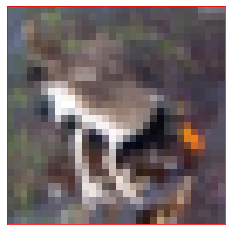}} &
{\includegraphics[width = 0.5\linewidth,valign=m]{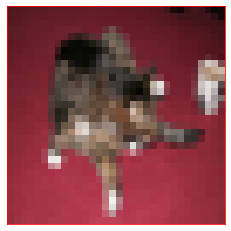}} &
{\includegraphics[width = 0.5\linewidth,valign=m]{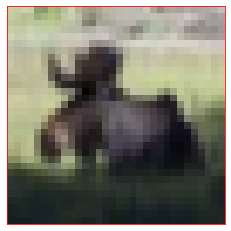}} &
{\includegraphics[width = 0.5\linewidth,valign=m]{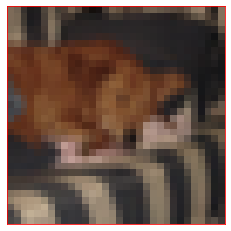}} &
{\includegraphics[width = 0.5\linewidth,valign=m]{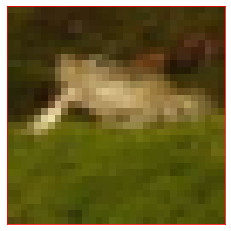}} &
{\includegraphics[width = 0.5\linewidth,valign=m]{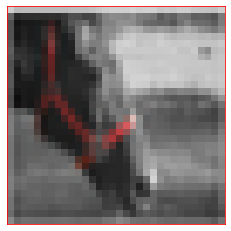}} &
{\includegraphics[width = 0.5\linewidth,valign=m]{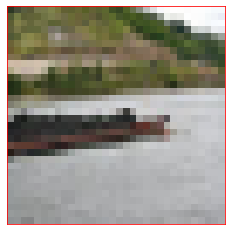}} &
{\includegraphics[width = 0.5\linewidth,valign=m]{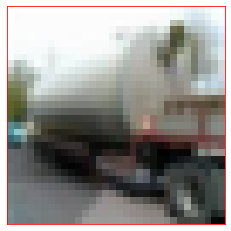}} \\
		HD=99 & HD=98 & HD=99 & HD=99 & HD=99 & HD=99 & HD=99 & HD=99 & HD=99 & HD=99 \\
		{\includegraphics[width = 0.5\linewidth,valign=m]{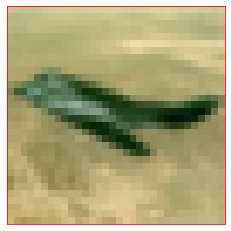}} &
		{\includegraphics[width = 0.5\linewidth,valign=m]{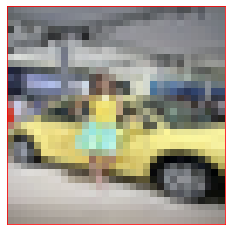}} &
		{\includegraphics[width = 0.5\linewidth,valign=m]{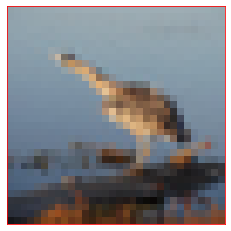}} &
		{\includegraphics[width = 0.5\linewidth,valign=m]{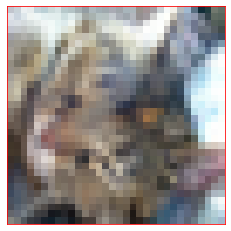}} &
		{\includegraphics[width = 0.5\linewidth,valign=m]{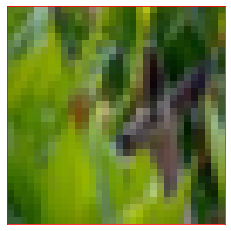}} &
		{\includegraphics[width = 0.5\linewidth,valign=m]{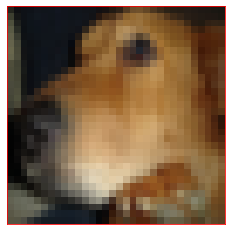}} &
		{\includegraphics[width = 0.5\linewidth,valign=m]{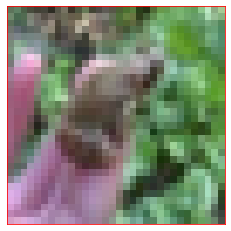}} &
		{\includegraphics[width = 0.5\linewidth,valign=m]{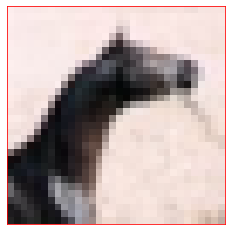}} &
		{\includegraphics[width = 0.5\linewidth,valign=m]{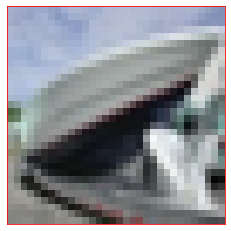}} &
		{\includegraphics[width = 0.5\linewidth,valign=m]{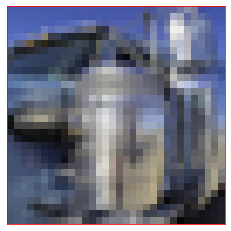}} \\
		HD=99 & HD=98 & HD=99 & HD=99 & HD=99 & HD=99 & HD=99 & HD=99 & HD=99 & HD=98 \\
		{\includegraphics[width = 0.5\linewidth,valign=m]{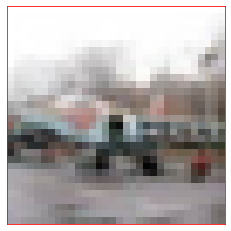}} &
		{\includegraphics[width = 0.5\linewidth,valign=m]{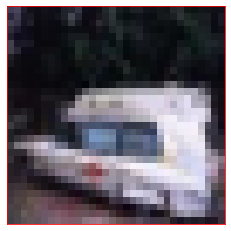}} &
		{\includegraphics[width = 0.5\linewidth,valign=m]{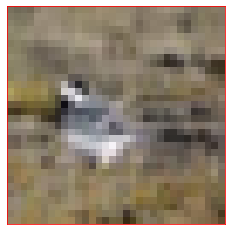}} &
		{\includegraphics[width = 0.5\linewidth,valign=m]{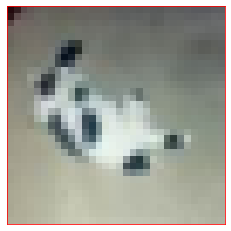}} &
		{\includegraphics[width = 0.5\linewidth,valign=m]{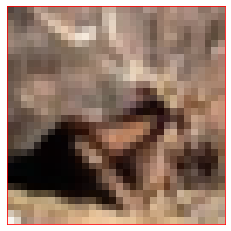}} &
		{\includegraphics[width = 0.5\linewidth,valign=m]{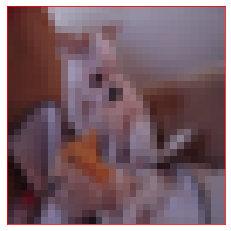}} &
		{\includegraphics[width = 0.5\linewidth,valign=m]{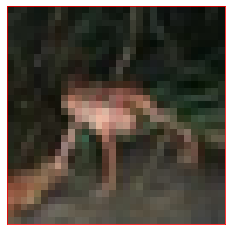}} &
		{\includegraphics[width = 0.5\linewidth,valign=m]{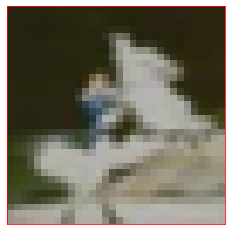}} &
		{\includegraphics[width = 0.5\linewidth,valign=m]{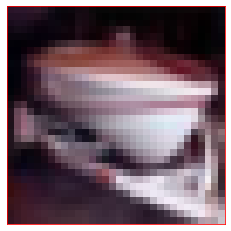}} &
		{\includegraphics[width = 0.5\linewidth,valign=m]{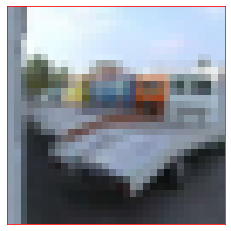}} \\
		HD=99 & HD=97 & HD=99 & HD=99 & HD=99 & HD=99 & HD=99 & HD=99 & HD=98 & HD=98 \\
	\end{tabular}
}
\caption{ \small{Hard Samples}}
\end{subfigure}
\begin{subfigure}{\linewidth}
		\resizebox{\linewidth}{!}{
		\begin{tabular}{cccccccccc}
			Airplane & Automobile & Bird & Cat & Deer & Dog & Frog & Horse & Ship & Truck \\
			{\includegraphics[width = 0.5\linewidth,valign=m]{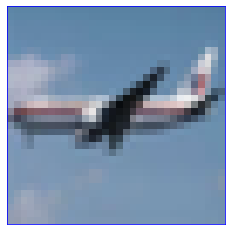}} &
			{\includegraphics[width = 0.5\linewidth,valign=m]{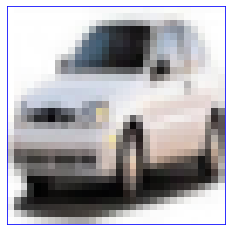}} &
			{\includegraphics[width = 0.5\linewidth,valign=m]{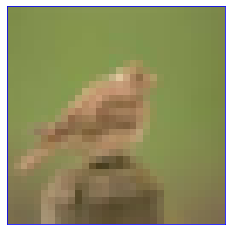}} &
			{\includegraphics[width = 0.5\linewidth,valign=m]{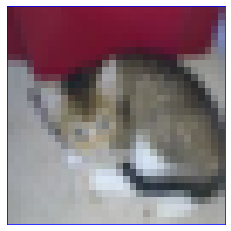}} &
			{\includegraphics[width = 0.5\linewidth,valign=m]{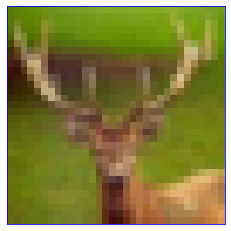}} &
			{\includegraphics[width = 0.5\linewidth,valign=m]{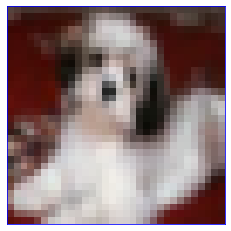}} &
			{\includegraphics[width = 0.5\linewidth,valign=m]{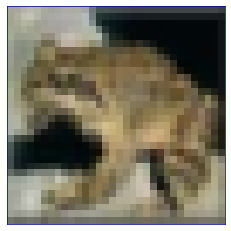}} &
			{\includegraphics[width = 0.5\linewidth,valign=m]{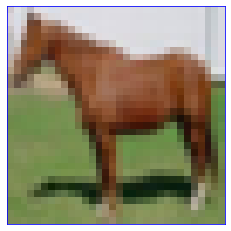}} &
			{\includegraphics[width = 0.5\linewidth,valign=m]{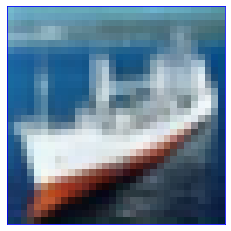}} &
			{\includegraphics[width = 0.5\linewidth,valign=m]{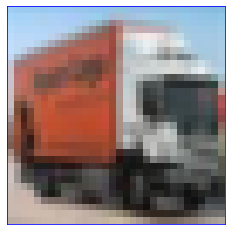}} \\
			HD=0 & HD=0 & HD=0 & HD=0 & HD=0 & HD=0 & HD=0 & HD=0 & HD=0 & HD=0 \\
			{\includegraphics[width = 0.5\linewidth,valign=m]{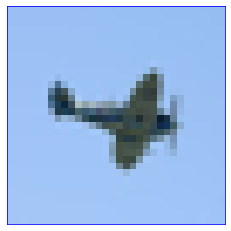}} &
			{\includegraphics[width = 0.5\linewidth,valign=m]{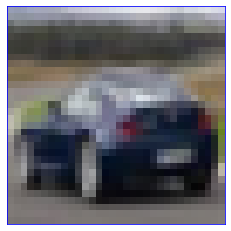}} &
			{\includegraphics[width = 0.5\linewidth,valign=m]{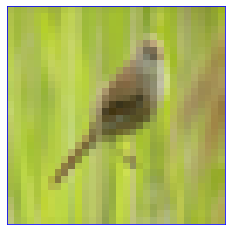}} &
			{\includegraphics[width = 0.5\linewidth,valign=m]{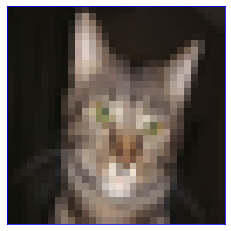}} &
			{\includegraphics[width = 0.5\linewidth,valign=m]{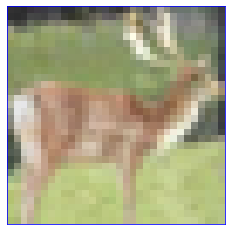}} &
			{\includegraphics[width = 0.5\linewidth,valign=m]{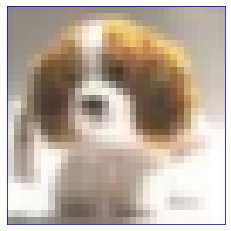}} &
			{\includegraphics[width = 0.5\linewidth,valign=m]{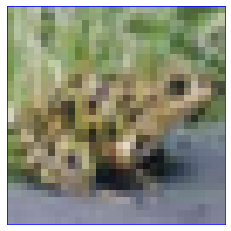}} &
			{\includegraphics[width = 0.5\linewidth,valign=m]{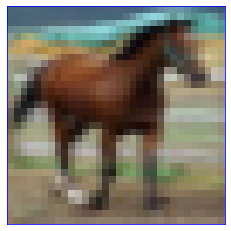}} &
			{\includegraphics[width = 0.5\linewidth,valign=m]{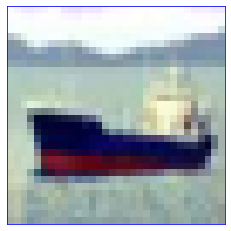}} &
			{\includegraphics[width = 0.5\linewidth,valign=m]{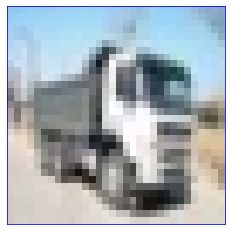}} \\
			HD=0 & HD=0 & HD=0 & HD=0 & HD=0 & HD=0 & HD=0 & HD=0 & HD=0 & HD=0 \\
			{\includegraphics[width = 0.5\linewidth,valign=m]{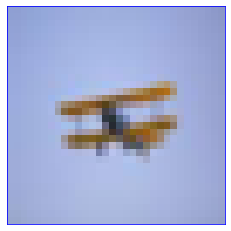}} &
			{\includegraphics[width = 0.5\linewidth,valign=m]{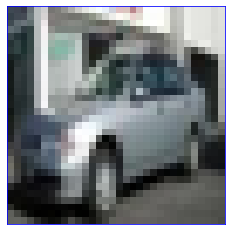}} &
			{\includegraphics[width = 0.5\linewidth,valign=m]{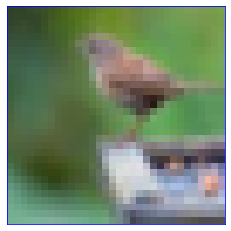}} &
			{\includegraphics[width = 0.5\linewidth,valign=m]{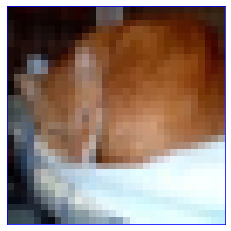}} &
			{\includegraphics[width = 0.5\linewidth,valign=m]{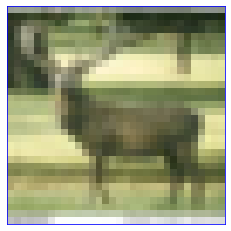}} &
			{\includegraphics[width = 0.5\linewidth,valign=m]{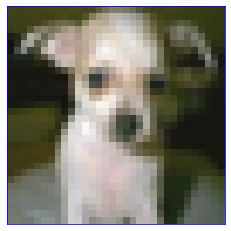}} &
			{\includegraphics[width = 0.5\linewidth,valign=m]{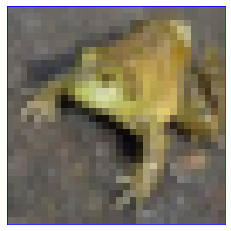}} &
			{\includegraphics[width = 0.5\linewidth,valign=m]{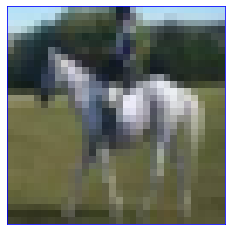}} &
			{\includegraphics[width = 0.5\linewidth,valign=m]{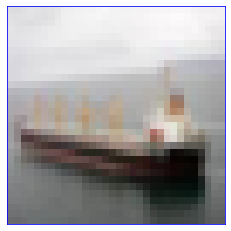}} &
			{\includegraphics[width = 0.5\linewidth,valign=m]{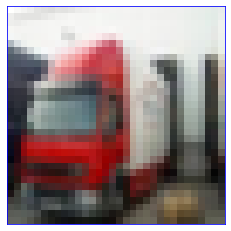}} \\
			HD=0 & HD=0 & HD=0 & HD=0 & HD=0 & HD=0 & HD=0 & HD=0 & HD=0 & HD=0 \\
		\end{tabular}
	}
\caption{ \small{Easy Samples}}
	\end{subfigure}

\caption{Some of the easiest and hardest CIFAR10 test samples based on the hardness degree for each class.}
	\label{fig:cifarvishe}
\end{figure}

We train three various types of classifiers, including DenseNet121 \cite{8099726}, ResNet18 \cite{7780459}, and MobileNet \cite{8578572}, on CIFAR10 and CIFAR100 training sets for 100 epochs (details of datasets in Appendix \ref{sec:appa}). All classifiers are trained using stochastic gradient descent with momentum 0.9 and batch size 128.
The learning rate is $0.1$ and it is scheduled to be decreased in each epoch by a constant factor of $0.955$.
The accuracy of classifiers is presented in Table \ref{tab:clsacc}. We save all 100 subclassifiers in the training phase of each classifier and use them to calculate the hardness degree of samples ($|F_{subclf}|=100$).
Figure \ref{loginMock} shows the hardness degree histogram of CIFAR10 and CIFAR100 test samples for various classifiers. The figure demonstrates that a large fraction of CIFAR10 test samples is easy, and CIFAR100 test set has more number of hard samples than CIFAR10 test set. Figure \ref{fig:cifarvishe} shows some examples of the easiest and hardest CIFAR10 test samples for each class to sanity check our hardness measure. The figure indicates that for easy samples, it is clear that they belong to their true label class, but for hard samples, it is not clear. Also, Figure \ref{fig:trust} in Appendix \ref{secmisrate} demonstrates a strong positive correlation between the hardness degree of samples and the misclassification rate.
 As ResNet18 architecture achieves strong performance on both datasets at a reasonable computational cost, we use this architecture for target classifiers in the rest of the paper.  We conduct various  model extraction attacks on two CIFAR10 and CIFAR100 target classifiers in the next subsection to depict the hardness degree histogram of their samples.


%% file: content/figure/hardnesshist.tex
\begin{figure}[!t]
	\begin{subfigure}{\linewidth}
		\centering
		\resizebox{\linewidth}{!}{
		\begin{tikzpicture}
\begin{groupplot}[group style={group size= 3 by 1,horizontal sep=0.4cm},height=3.5cm,width=4cm,tick label style = {font = {\fontsize{8 pt}{12 pt}\selectfont}},,ytick align=outside,xtick style={draw=none},tick label style={font=\scriptsize},,ytick pos=left,enlargelimits=true,enlargelimits=0.02,tick label style={font=\scriptsize}, label style={font=\scriptsize},title style={yshift=-0.2cm}]
\nextgroupplot[ymin=0, ymax=2250,	ytick = {0,1000,2000},yticklabels = {0,1K,2K},title=\small{DenseNet121}]
\addplot
[ybar interval,mark=no,fill=purple] 
coordinates
{(0,2050)(1,660)(2,500)(3,588)(4,315)(5,158)(6,108)(7,84)(8,117)(9,64)(10,373)(11,91)(12,144)(13,121)(14,66)(15,33)(16,44)(17,105)(18,76)(19,47)(20,91)(21,71)(22,178)(23,85)(24,71)(25,37)(26,119)(27,64)(28,72)(29,100)(30,68)(31,70)(32,62)(33,41)(34,67)(35,51)(36,66)(37,26)(38,50)(39,53)(40,182)(41,107)(42,61)(43,58)(44,57)(45,51)(46,25)(47,68)(48,82)(49,39)(50,65)(51,63)(52,69)(53,40)(54,66)(55,27)(56,34)(57,53)(58,39)(59,35)(60,59)(61,64)(62,63)(63,26)(64,49)(65,45)(66,69)(67,46)(68,30)(69,46)(70,65)(71,40)(72,29)(73,43)(74,23)(75,33)(76,29)(77,21)(78,46)(79,24)(80,40)(81,22)(82,29)(83,37)(84,16)(85,26)(86,19)(87,19)(88,25)(89,23)(90,38)(91,43)(92,36)(93,24)(94,16)(95,26)(96,52)(97,53)(98,57)(99,142)(100,0)
} 
\closedcycle;
\nextgroupplot[ymin=0,ymax=2250,title=\small{ResNet18},yticklabels={,,}]
\addplot
[ybar interval,mark=no,fill=purple] 
coordinates
{(0,1212)(1,727)(2,414)(3,477)(4,437)(5,204)(6,471)(7,101)(8,109)(9,104)(10,158)(11,109)(12,129)(13,102)(14,208)(15,265)(16,59)(17,103)(18,92)(19,118)(20,362)(21,64)(22,105)(23,97)(24,39)(25,62)(26,103)(27,115)(28,99)(29,77)(30,29)(31,80)(32,116)(33,43)(34,46)(35,50)(36,68)(37,97)(38,80)(39,75)(40,78)(41,69)(42,66)(43,100)(44,75)(45,75)(46,117)(47,91)(48,33)(49,74)(50,92)(51,103)(52,59)(53,58)(54,80)(55,53)(56,67)(57,63)(58,66)(59,41)(60,30)(61,41)(62,71)(63,71)(64,45)(65,41)(66,40)(67,46)(68,42)(69,28)(70,29)(71,32)(72,13)(73,27)(74,34)(75,20)(76,29)(77,21)(78,18)(79,17)(80,10)(81,16)(82,23)(83,26)(84,16)(85,12)(86,8)(87,20)(88,8)(89,15)(90,9)(91,20)(92,14)(93,26)(94,12)(95,16)(96,17)(97,49)(98,54)(99,68)(100,0)
} 
\closedcycle;
\nextgroupplot[ymin=0,ymax=2250,
title=\small{MobileNet},yticklabels={,,}]
\addplot
[ybar interval,mark=no,fill=purple] 
coordinates
{(0,2234)(1,511)(2,410)(3,221)(4,118)(5,91)(6,115)(7,139)(8,80)(9,146)(10,108)(11,73)(12,123)(13,101)(14,68)(15,112)(16,110)(17,59)(18,183)(19,117)(20,107)(21,127)(22,76)(23,43)(24,119)(25,121)(26,89)(27,74)(28,49)(29,81)(30,72)(31,52)(32,52)(33,38)(34,118)(35,81)(36,55)(37,62)(38,94)(39,63)(40,72)(41,60)(42,52)(43,55)(44,38)(45,92)(46,58)(47,80)(48,32)(49,69)(50,86)(51,55)(52,51)(53,73)(54,59)(55,53)(56,83)(57,65)(58,44)(59,37)(60,69)(61,62)(62,41)(63,45)(64,57)(65,54)(66,40)(67,67)(68,43)(69,47)(70,51)(71,48)(72,38)(73,42)(74,49)(75,53)(76,39)(77,30)(78,51)(79,35)(80,36)(81,46)(82,46)(83,42)(84,40)(85,33)(86,36)(87,35)(88,38)(89,41)(90,45)(91,53)(92,35)(93,57)(94,43)(95,71)(96,60)(97,78)(98,96)(99,202)(100,0)
} 
\closedcycle;
	\end{groupplot}
\end{tikzpicture}
}
\centering
\vspace{-0.6cm}
\caption{ \small{CIFAR10}}
\end{subfigure} \\
\begin{subfigure}{\linewidth}
\centering
\resizebox{\linewidth}{!}{
\begin{tikzpicture}
\begin{groupplot}[group style={group size= 3 by 2,horizontal sep=0.4cm},height=3.5cm,width=4cm,tick label style = {font = {\fontsize{8 pt}{12 pt}\selectfont}},,ytick align=outside,xtick style={draw=none},tick label style={font=\scriptsize},,ytick pos=left,enlargelimits=true,enlargelimits=0.02,,tick label style={font=\scriptsize}, label style={font=\scriptsize}]
\nextgroupplot[ymin=0, ymax=1050,	ytick = {0,500,1000},yticklabels = {0,.5K,1K},xlabel = Hardness Degree]
\addplot
[ybar interval,mark=no,fill=purple] 
coordinates
{(0,240)(1,212)(2,341)(3,187)(4,212)(5,182)(6,90)(7,85)(8,79)(9,79)(10,67)(11,99)(12,94)(13,67)(14,83)(15,86)(16,89)(17,115)(18,103)(19,78)(20,73)(21,83)(22,64)(23,75)(24,78)(25,83)(26,67)(27,85)(28,58)(29,78)(30,166)(31,123)(32,96)(33,85)(34,61)(35,95)(36,99)(37,130)(38,103)(39,127)(40,154)(41,109)(42,100)(43,114)(44,133)(45,106)(46,127)(47,119)(48,109)(49,99)(50,110)(51,108)(52,124)(53,115)(54,121)(55,137)(56,132)(57,119)(58,106)(59,100)(60,120)(61,107)(62,87)(63,119)(64,79)(65,92)(66,60)(67,56)(68,81)(69,44)(70,34)(71,64)(72,59)(73,56)(74,53)(75,58)(76,62)(77,30)(78,66)(79,42)(80,57)(81,75)(82,51)(83,48)(84,33)(85,69)(86,56)(87,69)(88,41)(89,61)(90,52)(91,64)(92,61)(93,101)(94,80)(95,104)(96,93)(97,163)(98,201)(99,393)(100,0)
} 
\closedcycle;
\nextgroupplot[ymin=0,ymax=1050,xlabel style={align=center},xlabel = Hardness Degree,yticklabels={,,}]
\addplot
[ybar interval,mark=no,fill=purple] 
coordinates
{(0,162)(1,218)(2,256)(3,310)(4,225)(5,97)(6,95)(7,114)(8,71)(9,201)(10,113)(11,104)(12,88)(13,69)(14,69)(15,84)(16,118)(17,86)(18,86)(19,78)(20,78)(21,95)(22,86)(23,94)(24,88)(25,140)(26,152)(27,80)(28,96)(29,97)(30,141)(31,80)(32,129)(33,207)(34,109)(35,176)(36,166)(37,191)(38,188)(39,164)(40,144)(41,185)(42,198)(43,148)(44,175)(45,152)(46,219)(47,185)(48,143)(49,155)(50,143)(51,128)(52,133)(53,113)(54,119)(55,99)(56,84)(57,71)(58,62)(59,83)(60,62)(61,69)(62,76)(63,39)(64,51)(65,43)(66,55)(67,47)(68,29)(69,37)(70,40)(71,26)(72,38)(73,38)(74,23)(75,24)(76,46)(77,25)(78,26)(79,29)(80,45)(81,27)(82,24)(83,32)(84,32)(85,54)(86,26)(87,41)(88,28)(89,37)(90,30)(91,48)(92,60)(93,41)(94,73)(95,59)(96,101)(97,59)(98,153)(99,267)(100,0)
};
\nextgroupplot[ymin=0,ymax=1050,,	ytick = {0,500,1000},yticklabels={,,},xlabel = Hardness Degree]
\addplot
[ybar interval,mark=no,fill=purple] 
coordinates
{(0,117)(1,169)(2,94)(3,253)(4,183)(5,90)(6,53)(7,71)(8,63)(9,71)(10,76)(11,98)(12,79)(13,73)(14,57)(15,80)(16,62)(17,102)(18,70)(19,42)(20,82)(21,81)(22,67)(23,64)(24,124)(25,86)(26,67)(27,72)(28,47)(29,53)(30,78)(31,75)(32,96)(33,80)(34,77)(35,74)(36,81)(37,78)(38,103)(39,82)(40,68)(41,84)(42,86)(43,72)(44,105)(45,76)(46,104)(47,75)(48,71)(49,66)(50,79)(51,80)(52,67)(53,88)(54,91)(55,88)(56,91)(57,119)(58,93)(59,100)(60,72)(61,81)(62,107)(63,73)(64,153)(65,83)(66,78)(67,104)(68,131)(69,119)(70,126)(71,80)(72,116)(73,106)(74,107)(75,113)(76,85)(77,91)(78,91)(79,107)(80,80)(81,67)(82,98)(83,76)(84,72)(85,86)(86,74)(87,86)(88,85)(89,89)(90,81)(91,74)(92,115)(93,103)(94,118)(95,129)(96,192)(97,190)(98,351)(99,638)(100,0)};
\end{groupplot}
\end{tikzpicture}
}
\centering
\vspace{-0.6cm}
\centering{\caption{ \small{CIFAR100}}}
\end{subfigure}
    \caption{The hardness degree histograms of CIFAR10 and CIFAR100 test samples for DenseNet121, ResNet18, and MobileNet classifiers.}
\label{loginMock}
\end{figure}

%% file: content/attackanalysis.tex
In line with prior work \cite{DBLP:conf/iclr/OrekondySF20,DBLP:conf/cvpr/KariyappaQ20,kariyappa2021protecting}, we select JBDA \cite{10.1145/3052973.3053009}, JBRAND \cite{DBLP:conf/eurosp/JuutiSMA19}, and Knockoff Net \cite{DBLP:conf/cvpr/OrekondySF19} model extraction attacks to evaluate our defense method. These attacks broadly represent two main strategies (synthesis or semantically similar samples) to conduct model extraction attacks. 
Jacobian-Based Dataset Augmentation (JBDA) \cite{10.1145/3052973.3053009} and its improvement (JBRAND) \cite{DBLP:conf/eurosp/JuutiSMA19} assume that the adversary has access to a limited number of samples from the target classifier's training data distribution called seed samples, and they aim to augment seed samples using adversarial examples to increase the \textit{fidelity} of the surrogate classifier to the target classifier. 
\cite{DBLP:conf/cvpr/OrekondySF19} propose Knockoff Net (K.Net) attack that uses large public datasets that is semantically similar to the target classifier dataset to increase the \textit{accuracy} of the surrogate classifier. We consider two versions of K.Net attack, K.Net CIFARX, and K.Net TIN. K.Net CIFARX attack uses CIFAR100 training set to extract CIFAR10 target classifier and vice versa. K.Net TIN employs TinyImageNet training set to extract target classifiers. More details about attacks and their implementations are presented in Appendix \ref{sec:attackdetail}. 

\begin{table}
	\caption{The Accuracy (Acc) and the Fidelity (Fid) of surrogate classifiers being created by four various model extraction attacks on two target classifiers CIFAR10 and CIFAR100. The output type of target classifiers can be Label or Probability Vector (Prob. Vec.).}
	\label{tab:attackaccfid}
	\renewcommand{\arraystretch}{1.5}
	\resizebox{\linewidth}{!}{
		\begin{tabular}{ccccccc}
			\hline
			$f_t$                                                                              & Metric                   &Output type  & JBDA  & JBRAND & K.Net CIFARX & K.Net TIN \\ \hline
			\multirow{4}{*}{\begin{tabular}[c]{@{}c@{}c@{}}CIFAR10 \\ ResNet18 \\ (Acc: 94.36\%)\end{tabular}}  & \multirow{2}{*}{Acc(\%)} & Prob. Vec. & 41.00 & 43.33   & 79.86         & 78.86             \\
			&                          & Label      & 34.57 & 34.35   & 66.88         & 71.29             \\ \cline{3-7} 
			& \multirow{2}{*}{Fid(\%)} & Prob. Vec. & 41.16 & 43.63   & 81.36         & 80.18             \\
			&                          & Label      & 34.86 & 34.45   & 67.98         & 72.43             \\ \hline
			\multirow{4}{*}{\begin{tabular}[c]{@{}c@{}c@{}}CIFAR100 \\ ResNet18 \\ (Acc: 76.38\%)\end{tabular}} & \multirow{2}{*}{Acc(\%)} & Prob. Vec. & 16.44 & 18.78   & 51.09         & 60.36             \\
			&                          & Label      & 8.62  & 8.07    & 23.20         & 32.88             \\ \cline{3-7} 
			& \multirow{2}{*}{Fid(\%)} & Prob. Vec. & 16.90 & 19.13   & 54.59         & 64.90             \\
			&                          & Label      & 8.91  & 8.29    & 24.72         & 34.58       \\
			\hline
		\end{tabular}
	}
\end{table}
To evaluate the performance of model extraction attacks, we use two ResNet18 classifiers being trained on CIFAR10 and CIFAR100 training sets as the target classifiers and conduct all four attacks on them.  The default value of the attack budget in our experiments is B=50000 (same as \cite{kariyappa2021protecting,DBLP:conf/cvpr/KariyappaQ20,DBLP:conf/iclr/OrekondySF20}). Table \ref{tab:attackaccfid} shows the accuracy and the fidelity of surrogate classifiers created by various model extraction attacks on CIFAR10 and CIFAR100 test samples. The results demonstrate that K.Net attacks have significantly better performance than jacobian-based attacks (JBDA and JBRAND), and when the output of target classifiers is probability vector, the performance of attacks is considerably increased.

%% file: content/hbdm.tex
 \input{content/histattackresnet.tex}
\input{content/vis.tex}

Figure \ref{fig:hardnesshistmearesnet} depicts the hardness degree histogram of 50000 samples generated by various attacks for CIFAR10 and CIFAR100 target classifiers. In this experiment, the architecture of target classifiers is ResNet18. We also present the hardness degree histogram of attack samples when the architecture of target classifiers is Densenet121 in Appendix \ref{sec:histattackmeadensenet}.
 Figure \ref{fig:hardnesshistmearesnet} demonstrates that the samples generated by various attacks have a very small number of easy samples, and most samples have medium and high hardness degrees. 
 Figure \ref{fig:vis} displays two-dimensional visualization of CIFAR10 test samples using t-SNE \cite{JMLR:v9:vandermaaten08a}. Figure \ref{fig:visa} uses the logits of the CIFAR10 classifier to visualize CIFAR10 test samples, and the color of each sample is determined by its label. This figure has ten sample clusters where most samples of each cluster are from one class.
 Figure \ref{fig:visb} illustrates the hardness degree of CIFAR10 test samples for CIFAR10 target classifier and demonstrates that most of the easy samples are in the high-density regions inside clusters, and most of the hard samples are in the low-density regions at the borders of clusters. Figure \ref{fig:visc} is similar to Figure \ref{fig:visb}, but the hardness degree of each sample is calculated via CIFAR100 target classifier.
 This figure demonstrates when the training data distribution of the classifier being used to calculate the hardness degree of samples becomes different from the distribution of CIFAR10 test samples, the hardness degree of a high number of samples is changed.
 Figure \ref{fig:visc} shows hard and medium samples are distributed among clusters, and the number of easy samples is very small. 
 Similar to Figure \ref{fig:vis}, we visualize CIFAR100 test samples and their hardness for CIFAR10 and CIFAR100 target classifiers in Appendix \ref{sec:viscifar100}.
 
\textcolor{\changecolor}{Based on our experiments and the findings of previous works \cite{DBLP:conf/icml/JiangZTM21,carlini2019prototypical}, 
easy samples lie on the data manifold that is well supported by the target classifier training data. In other words, easy samples are from the high probability region of the input space. 
Figures \ref{fig:hardnesshistmearesnet} and \ref{fig:hardtlhist} demonstrate that, unlike normal samples, the number of easy samples among model extraction attack samples is very small, which means that attack samples are from the low probability region of the input space and they do not lie on the data manifold that is well supported by the target classifier training data. Furthermore, the experiments demonstrate that the hardness degree of a sample for a classifier pertains to the training data distribution of that classifier. For example, although most CIFAR10 test samples are easy for CIFAR10 target classifier, they are hard for CIFAR100 target classifier because they are not well supported by CIFAR100 data manifold.}
We use histogram rather than hardness degree histogram in the rest of the paper for simplicity.

\subsection{HODA: Hardness-Oriented Detection Approach}
 We propose Hardness-Oriented Detection Approach (HODA) to detect sample sequences of model extraction attacks.
  \textcolor{\changecolor}{We assume that each user of the target classifier has an account that is associated with the email or the phone number of that user, and the queries of each user are separable from the other users' queries.} When new sample $x_i$ from user $u$ arrives, HODA calculates its hardness degree $\phi_{f_t}(x_i)$, and the histogram belongs to that user $H_u$ is updated. 
  HODA requires normal histogram $H_n$ representing the histogram of normal samples. \textcolor{\changecolor}{$H_n$ indicates the hardness degree histogram of samples drawn iid from the target classifier's data distribution.}
  After the number of samples sent by user $u$ reaches a specific number $num_{s}$, HODA calculates Pearson distance between the histograms of normal samples $H_n$ and user samples $H_u$, and if the distance exceeds a threshold $\delta$, the user $u$ is detected as an adversary.
Pearson Distance (PD) between two random variable $X$ and $Y$ is defined as follows:
\begin{equation}
PD(X,Y) = 1 -  \frac{\text{Cov}(X,Y)}{\rho_X\rho_Y}
\end{equation}
where $\text{Cov}(X,Y)$ is the covariance between random variables $X$ and $Y$, and $\rho_X$ is the standard deviation of random variable $X$. The output of Pearson distance is in the range [0,2].
\textcolor{\changecolor}{The output of Pearson distance indicates the inconsistency between samples sent by user $u$ and normal samples.}
To calculate the Pearson distance between two histograms, HODA first transforms histograms into probability vectors by dividing the value of histogram bins by the total number of samples in the histogram ($H_n/sum(H_n)$ and $H_u/sum(H_u)$) and then calculates the Pearson distance between them.

HODA uses normal sample set $S_{HODA}$ to create $H_n$ and calculate $\delta$. 
It randomly selects $num_{seq}$ sample sequences with size $num_s$ from the sample set $S_{HODA}$ and for each sample sequence, produces a histogram and adds it to the histogram set $HistSet$. The normal histogram $H_n$ is the average of all histograms in $HistSet$, and $\delta$ is the maximum Pearson distance between $H_n$ and all histograms in $HistSet$.  
Since $\delta$ is independent of attacks and only relies on normal samples, HODA is not dependent on any attacks. 
Notably, HODA does not need to save samples of each user or their hardness degrees.  It only keeps a vector ($H_u$) that indicates the values of histogram bins for each user.
Algorithms \ref{alg:hd} and \ref{alg:hoda} in Appendix \ref{sec:hodaalg} describes HODA in details.

%% file: content/histattackresnet.tex
\begin{figure}
	\label{fig:hardhist}
	\begin{subfigure}{\linewidth}
		\centering
		\resizebox{\linewidth}{!}{
		\begin{tikzpicture}
\begin{groupplot}[group style={group size= 4 by 1,vertical sep=0.9cm},height=4cm,width=4.8cm,tick label style = {font = {\fontsize{8 pt}{12 pt}\selectfont}},,ytick align=outside,xtick style={draw=none},tick label style={font=\scriptsize},,ytick pos=left,enlargelimits=true,enlargelimits=0.02,,title style={yshift=-0.2cm}]
\nextgroupplot[ymin=0, ymax=10000,title=JBDA,ylabel=\# of samples,ytick={0,2000,4000,6000,8000,10000},yticklabels={0,2K,4K,6K,8K,10K},scaled y ticks = false]
\addplot
[ybar interval,mark=no,fill=purple] 
coordinates
{

(0,400)(1,413)(2,82)(3,70)(4,89)(5,208)(6,91)(7,16)(8,95)(9,12)(10,11)(11,35)(12,7)(13,4)(14,9)(15,29)(16,13)(17,21)(18,27)(19,12)(20,71)(21,12)(22,23)(23,18)(24,135)(25,99)(26,105)(27,168)(28,32)(29,29)(30,13)(31,63)(32,62)(33,33)(34,21)(35,64)(36,24)(37,47)(38,64)(39,42)(40,48)(41,99)(42,20)(43,55)(44,1178)(45,48)(46,4617)(47,298)(48,30)(49,467)(50,1882)(51,88)(52,28)(53,146)(54,368)(55,161)(56,208)(57,5026)(58,6040)(59,1476)(60,81)(61,110)(62,3676)(63,431)(64,883)(65,642)(66,260)(67,274)(68,63)(69,399)(70,981)(71,227)(72,111)(73,3300)(74,498)(75,182)(76,302)(77,1096)(78,111)(79,240)(80,163)(81,275)(82,746)(83,293)(84,143)(85,42)(86,757)(87,30)(88,716)(89,97)(90,116)(91,199)(92,19)(93,230)(94,311)(95,276)(96,1105)(97,604)(98,937)(99,1522)(100,0)
} 
\closedcycle;
\nextgroupplot[ymin=0, ymax=10000,title=JBRAND,ytick={0,2000,4000,6000,8000,10000},yticklabels={0,2K,4K,6K,8K,10K},scaled y ticks = false]
\addplot
[ybar interval,mark=no,fill=purple] 
coordinates
{
(0,135)(1,128)(2,36)(3,23)(4,36)(5,60)(6,34)(7,11)(8,20)(9,2)(10,6)(11,11)(12,8)(13,2)(14,8)(15,12)(16,5)(17,7)(18,10)(19,4)(20,38)(21,4)(22,6)(23,7)(24,66)(25,10)(26,8)(27,13)(28,11)(29,6)(30,3)(31,7)(32,20)(33,3)(34,11)(35,8)(36,9)(37,22)(38,30)(39,17)(40,13)(41,17)(42,7)(43,16)(44,124)(45,11)(46,701)(47,236)(48,33)(49,57)(50,208)(51,56)(52,6)(53,24)(54,113)(55,92)(56,35)(57,5003)(58,9299)(59,3563)(60,41)(61,110)(62,1255)(63,156)(64,429)(65,766)(66,139)(67,94)(68,18)(69,353)(70,477)(71,874)(72,8)(73,4638)(74,364)(75,252)(76,333)(77,1741)(78,78)(79,235)(80,59)(81,521)(82,1712)(83,500)(84,185)(85,71)(86,784)(87,48)(88,302)(89,142)(90,307)(91,939)(92,84)(93,158)(94,362)(95,99)(96,2056)(97,721)(98,1013)(99,2605)(100,0)
} 
\closedcycle;
\nextgroupplot[ymin=0,ymax=4000,title=K.Net CIFARX,ytick={0,1000,2000,3000,4000},yticklabels={0,1K,2K,3K,4K},scaled y ticks = false]
\addplot
[ybar interval,mark=no,fill=purple] 
coordinates
{
	(0,210)(1,213)(2,82)(3,239)(4,105)(5,83)(6,127)(7,60)(8,76)(9,75)(10,87)(11,111)(12,130)(13,98)(14,107)(15,171)(16,68)(17,127)(18,137)(19,115)(20,297)(21,164)(22,177)(23,169)(24,88)(25,121)(26,116)(27,193)(28,167)(29,194)(30,131)(31,216)(32,418)(33,152)(34,177)(35,228)(36,234)(37,432)(38,445)(39,387)(40,465)(41,323)(42,439)(43,468)(44,493)(45,469)(46,696)(47,590)(48,475)(49,591)(50,745)(51,874)(52,743)(53,664)(54,815)(55,706)(56,910)(57,731)(58,942)(59,806)(60,569)(61,773)(62,964)(63,1291)(64,1251)(65,1003)(66,1163)(67,995)(68,995)(69,635)(70,663)(71,782)(72,454)(73,791)(74,790)(75,625)(76,733)(77,657)(78,427)(79,435)(80,447)(81,436)(82,768)(83,664)(84,472)(85,414)(86,325)(87,372)(88,471)(89,425)(90,349)(91,503)(92,306)(93,761)(94,661)(95,811)(96,762)(97,1116)(98,1390)(99,2679)(100,0)
} 
\closedcycle;
\nextgroupplot[ymin=0,ymax=4000,title=K.Net TIN,ytick={0,1000,2000,3000,4000},yticklabels={0,1K,2K,3K,4K},scaled y ticks = false]
\addplot
[ybar interval,mark=no,fill=purple] 
coordinates
{
(0,422)(1,235)(2,112)(3,289)(4,152)(5,102)(6,242)(7,61)(8,80)(9,82)(10,156)(11,111)(12,123)(13,84)(14,124)(15,262)(16,91)(17,173)(18,131)(19,150)(20,360)(21,127)(22,190)(23,195)(24,95)(25,131)(26,136)(27,218)(28,149)(29,208)(30,166)(31,221)(32,287)(33,143)(34,177)(35,221)(36,243)(37,422)(38,423)(39,367)(40,536)(41,327)(42,364)(43,557)(44,466)(45,457)(46,602)(47,670)(48,439)(49,581)(50,691)(51,740)(52,718)(53,703)(54,758)(55,659)(56,788)(57,738)(58,911)(59,756)(60,627)(61,856)(62,930)(63,1196)(64,1131)(65,1060)(66,1076)(67,988)(68,1129)(69,602)(70,633)(71,676)(72,474)(73,756)(74,930)(75,571)(76,715)(77,667)(78,457)(79,422)(80,388)(81,398)(82,678)(83,639)(84,364)(85,318)(86,329)(87,435)(88,429)(89,394)(90,317)(91,476)(92,289)(93,865)(94,543)(95,768)(96,783)(97,1052)(98,1535)(99,2982)(100,0)
} 
\closedcycle;
	\end{groupplot}
\end{tikzpicture}
}
\vspace{-0.5cm}
\caption{ \small{CIFAR10}}
\end{subfigure}
\begin{subfigure}{\linewidth}
\centering
\resizebox{\linewidth}{!}{
\begin{tikzpicture}
\begin{groupplot}[group style={group size= 4 by 1,vertical sep=0.9cm},height=4cm,width=4.8cm,tick label style = {font = {\fontsize{8 pt}{12 pt}\selectfont}},,ytick align=outside,xtick style={draw=none},tick label style={font=\scriptsize},,ytick pos=left,enlargelimits=true,enlargelimits=0.02,]
\nextgroupplot[ymin=0, ymax=10000,ylabel=\# of samples,xlabel = Hardness Degree,ytick={0,2000,4000,6000,8000,10000},yticklabels={0,2K,4K,6K,8K,10K},scaled y ticks = false]
\addplot
[ybar interval,mark=no,fill=purple] 
coordinates
{
	
	(0,31)(1,57)(2,74)(3,36)(4,28)(5,308)(6,7)(7,17)(8,21)(9,16)(10,5)(11,11)(12,7)(13,9)(14,17)(15,11)(16,113)(17,15)(18,17)(19,15)(20,67)(21,427)(22,19)(23,529)(24,16)(25,17)(26,49)(27,45)(28,350)(29,37)(30,36)(31,31)(32,62)(33,105)(34,103)(35,45)(36,26)(37,232)(38,1523)(39,104)(40,432)(41,88)(42,471)(43,2050)(44,2527)(45,714)(46,163)(47,174)(48,7806)(49,1444)(50,851)(51,293)(52,444)(53,1069)(54,786)(55,257)(56,329)(57,525)(58,114)(59,1796)(60,2009)(61,855)(62,336)(63,583)(64,533)(65,285)(66,253)(67,151)(68,440)(69,848)(70,114)(71,265)(72,189)(73,308)(74,338)(75,210)(76,209)(77,150)(78,174)(79,771)(80,124)(81,76)(82,150)(83,769)(84,458)(85,200)(86,191)(87,157)(88,211)(89,329)(90,244)(91,195)(92,468)(93,561)(94,646)(95,534)(96,664)(97,608)(98,1275)(99,6148)(100,0)
} 
\closedcycle;
\nextgroupplot[ymin=0, xlabel =Hardness Degree, ymax=10000,ytick={0,2000,4000,6000,8000,10000},yticklabels={0,2K,4K,6K,8K,10K},scaled y ticks = false]
\addplot
[ybar interval,mark=no,fill=purple] 
coordinates
{

(0,17)(1,26)(2,33)(3,32)(4,26)(5,8)(6,7)(7,17)(8,8)(9,14)(10,4)(11,6)(12,6)(13,9)(14,8)(15,10)(16,15)(17,10)(18,14)(19,10)(20,9)(21,34)(22,9)(23,22)(24,14)(25,9)(26,16)(27,13)(28,17)(29,16)(30,17)(31,14)(32,23)(33,47)(34,29)(35,24)(36,17)(37,89)(38,131)(39,31)(40,74)(41,22)(42,284)(43,577)(44,708)(45,308)(46,56)(47,115)(48,6298)(49,2271)(50,418)(51,203)(52,707)(53,457)(54,280)(55,181)(56,358)(57,420)(58,63)(59,3592)(60,5237)(61,1038)(62,214)(63,958)(64,320)(65,593)(66,229)(67,185)(68,480)(69,750)(70,78)(71,226)(72,133)(73,189)(74,627)(75,277)(76,252)(77,150)(78,186)(79,1322)(80,93)(81,38)(82,162)(83,1052)(84,428)(85,312)(86,222)(87,153)(88,311)(89,328)(90,288)(91,234)(92,731)(93,1030)(94,438)(95,840)(96,539)(97,475)(98,1677)(99,8951)(100,0)
} 
\closedcycle;
\nextgroupplot[ymin=0,ymax=4000,xlabel style={align=center},xlabel = Hardness Degree,ytick={0,1000,2000,3000,4000},yticklabels={0,1K,2K,3K,4K},scaled y ticks = false]
\addplot
[ybar interval,mark=no,fill=purple] 
coordinates
{
	(0,33)(1,29)(2,25)(3,24)(4,161)(5,27)(6,114)(7,61)(8,45)(9,82)(10,22)(11,84)(12,29)(13,16)(14,38)(15,80)(16,125)(17,69)(18,64)(19,52)(20,136)(21,148)(22,68)(23,71)(24,92)(25,143)(26,241)(27,93)(28,233)(29,217)(30,373)(31,170)(32,336)(33,426)(34,261)(35,606)(36,388)(37,606)(38,831)(39,632)(40,584)(41,894)(42,850)(43,921)(44,955)(45,980)(46,1258)(47,1177)(48,1009)(49,1152)(50,1096)(51,1115)(52,1145)(53,893)(54,995)(55,902)(56,701)(57,638)(58,599)(59,645)(60,930)(61,596)(62,635)(63,486)(64,543)(65,468)(66,435)(67,496)(68,355)(69,366)(70,387)(71,325)(72,391)(73,481)(74,445)(75,273)(76,374)(77,347)(78,209)(79,381)(80,337)(81,360)(82,370)(83,303)(84,404)(85,385)(86,524)(87,374)(88,487)(89,485)(90,409)(91,440)(92,562)(93,555)(94,893)(95,893)(96,1019)(97,1017)(98,1891)(99,3609)(100,0)
};
\nextgroupplot[ymin=0,ymax=4000,xlabel = Hardness Degree,ytick={0,1000,2000,3000,4000},yticklabels={0,1K,2K,3K,4K},scaled y ticks = false]
\addplot
[ybar interval,mark=no,fill=purple] 
coordinates
{
	(0,18)(1,48)(2,62)(3,101)(4,82)(5,28)(6,56)(7,71)(8,55)(9,113)(10,49)(11,67)(12,63)(13,47)(14,36)(15,54)(16,123)(17,52)(18,59)(19,67)(20,72)(21,128)(22,114)(23,95)(24,101)(25,187)(26,209)(27,126)(28,178)(29,175)(30,231)(31,200)(32,271)(33,411)(34,279)(35,401)(36,364)(37,614)(38,578)(39,554)(40,548)(41,785)(42,981)(43,829)(44,805)(45,1112)(46,1296)(47,1139)(48,989)(49,1082)(50,1001)(51,1247)(52,1174)(53,1015)(54,1090)(55,855)(56,748)(57,672)(58,667)(59,715)(60,835)(61,542)(62,581)(63,498)(64,532)(65,467)(66,563)(67,512)(68,368)(69,454)(70,409)(71,307)(72,456)(73,433)(74,392)(75,265)(76,404)(77,335)(78,294)(79,346)(80,385)(81,324)(82,299)(83,328)(84,447)(85,473)(86,323)(87,457)(88,334)(89,542)(90,413)(91,490)(92,610)(93,526)(94,952)(95,907)(96,1133)(97,948)(98,2136)(99,3691)(100,0)
};
\end{groupplot}
\end{tikzpicture}
}
\vspace{-0.5cm}
\caption{ \small{CIFAR100}}
\end{subfigure}

\caption{The hardness degree histograms of samples of four various model extraction attacks for CIFAR10 and CIFAR100 target classifiers. The budget of model extraction attacks is 50000. }
\label{fig:hardnesshistmearesnet}
\end{figure}
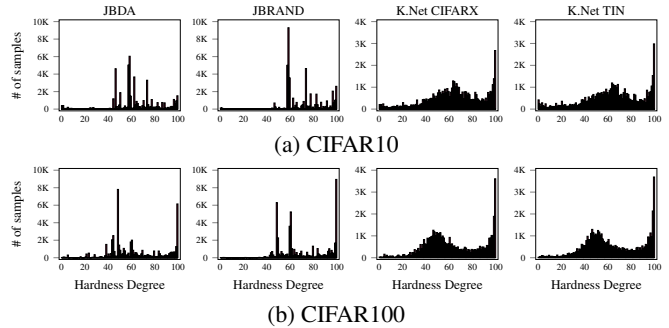

%% file: content/vis.tex
\begin{figure}[!t]
	\begin{subfigure}{0.32\linewidth}
		\resizebox{\linewidth}{!}{
			\begin{tikzpicture}
			\node (myfirstpic) at (0,0) {\includegraphics[scale=4]{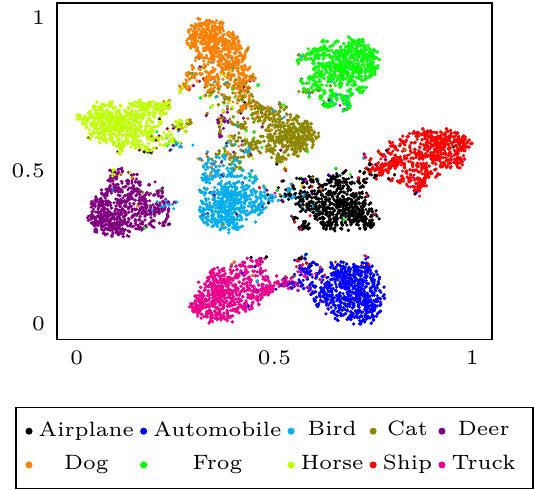}};
			\end{tikzpicture}
		}
		\caption{}
		\label{fig:visa}
	\end{subfigure}
	\begin{subfigure}{0.32\linewidth}
		\resizebox{\linewidth}{!}{
			\begin{tikzpicture}
			\node (myfirstpic) at (0,0) {\includegraphics[scale=2]{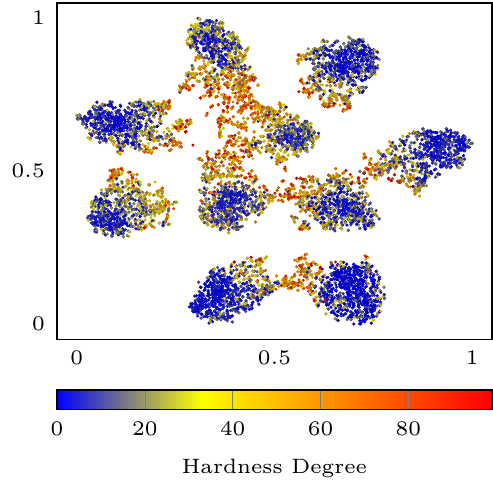}};
			\end{tikzpicture}
		}
		\caption{}
		\label{fig:visb}
	\end{subfigure}
	\begin{subfigure}{0.32\linewidth}
		\resizebox{\linewidth}{!}{
			\begin{tikzpicture}
			\node (myfirstpic) at (0,0) {\includegraphics[scale=2]{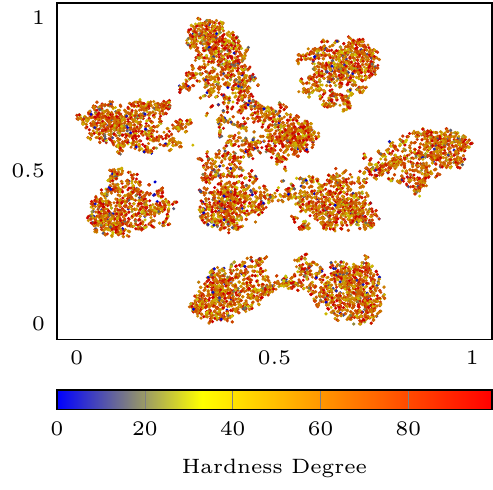}};
			\end{tikzpicture}
		}
		\caption{}
		\label{fig:visc}
	\end{subfigure}
	\caption{ (a) Visualization of CIFAR10 test samples. (b) Hardness of CIFAR10 test samples for CIFAR10 target classifier. (c) Hardness of CIFAR10 test samples for CIFAR100 target classifier. }
	\label{fig:vis}
\end{figure}

%% file: content/evaluation.tex
\input{content/hodadrcifar.tex}

Two normal sample sets $S_{HODA}$ and $S_u$ are required to evaluate the performance of HODA. $S_u$ is used to simulate benign users. We randomly select 40\% and 60\% of test samples of each dataset for  $S_{HODA}$ and $S_u$, respectively. We randomly select $num_{seq} = 40000$ sequences with size $num_s$ from $S_{HODA}$ to create $H_n$ and calculate $\delta$. To evaluate the performance of HODA against model extraction attacks, we simulate 10000 benign users and 10000 adversaries for each attack.
Each benign user sends a sequence of $num_s$ samples randomly selected from $S_u$, and each adversary sends a sequence of $num_s$ samples randomly selected from 50000 samples of attack in the order they were generated. 
So far, we have used 100 subclassifiers to calculate the hardness degree of samples. However, it may not be possible to classify each sample by a high number of subclassifiers in practice. 
\textcolor{\changecolor}{
So in order to reduce the computational cost of HODA, we consider two versions of HODA called HODA-11 and HODA-5.
HODA-11 uses 11 subclassifiers $F_{subclf}=\,<f_t^0, f_t^9, f_t^{19}, f_t^{29}, f_t^{39}, f_t^{49}, f_t^{59}, f_t^{69}, f_t^{79}, f_t^{89}, f_t^{99}>$, and HODA-5 uses 5 subclassifiers $F_{subclf}=\,<f_t^{19}, f_t^{39}, f_t^{59}, f_t^{79}, f_t^{99}>$ to calculate the hardness degree of each sample.
Since the hardness degree domain depends on the number of subclassifiers in $F_{subclf}$, the hardness degree in HODA-11 and HODA-5 is in the ranges [0,10] and [0,4], respectively.}

\textcolor{\changecolor}{We compare HODA with PRADA and VarDetect.}
PRADA \cite{DBLP:conf/eurosp/JuutiSMA19} declares that the histogram of minimum $L_2$ distance between a new sample and all previous samples of a benign user follows a Gaussian distribution. Hence, it uses the Shapiro-Wilk normality test to determine that a sample sequence belongs to a benign user or an adversary. Similar to HODA, PRADA also uses threshold $\delta$ to detect sample sequences of model extraction attacks, and $\delta$ is the only parameter of PRADA. Since PRADA needs to save each user's samples and calculate $L_2$ distance between them, it has a high computational overhead. 
\textcolor{\changecolor}{VarDetect \cite{DBLP:journals/corr/abs-2107-05166} computes MMD between the user samples and normal samples in the latent space of variational autoencoders. We follow the default configurations of VarDetect. However, since VarDetect has no mechanism to compute $\delta$, we use the mechanism of HODA for computing $\delta$ to determine $\delta$ for VarDetect.}

Table \ref{tab:hbdmperfmea} indicates the detection rate, AUC score, and False Positive Rate (FPR) of PRADA, VarDetect, HODA-11, and HODA-5 against four various model extraction attacks on CIFAR10 and CIFAR100 target classifiers. 
We evaluate defenses with various values for $num_s$ to better compare their capabilities. We consider $num_s=100$ as baseline.
All defenses have very low false-positive rates. False-Positive Rate (FPR) indicates the percentage of benign users' sample sequences wrongly detected as an attack. AUC score is independent of $\delta$ and indicates the defense performance to separate adversaries from benign users. 
The results demonstrate that HODA is very effective against model extraction attacks, and it outperforms PRADA and VarDetect by a large margin. HODA-11 is slightly better than HODA-5. 
\textcolor{\changecolor}{We used the test set of target model dataset to simulate benign users. In Appendix \ref{secpearsondisthist}, we show that if benign users use modified STL10 dataset samples to query the CIFAR10 target classifier, the performance of HODA is decent. Since the modified STL10 dataset consists of a subset of CIFAR10 classes, we consider its samples as normal samples for CIFAR10 target classifier.
	 Appendix \ref{secpearsondisthist} also indicates the Pearson distance histogram of benign users and adversaries for all model extraction attacks.
	To compare HODA with the state-of-the-art perturbation-based defense,
	Table \ref{tab:defsuracc} in Appendix \ref{sec:defadv} reports the accuracy of surrogate classifiers for defended target classifiers by HODA and EDM \cite{kariyappa2021protecting}.}

\subsection{Computational Cost Analysis}
\begin{table}
	\caption{The computational cost of PRADA, VarDetect, and HODA to defend CIFAR10 target classifier in $num_s=500$.}
	\label{tab:csd}
	\renewcommand{\arraystretch}{1.1}
	\centering
	\resizebox{\linewidth}{!}{
\begin{tabular}{lccc}
	\hline
	Defense   & \begin{tabular}[c]{@{}c@{}}Avg Runtime\\ of Defense\end{tabular} & \begin{tabular}[c]{@{}c@{}}Avg Memory\\ For Each User\end{tabular} & \begin{tabular}[c]{@{}c@{}}Number of Prediction\\ For each sample\end{tabular} \\ \hline
	PRADA     & 0.470 seconds                                                    & 471 * sample size                                                  & 1                                                                              \\
	VarDetect & 0.010 seconds                                                    & 500 * size of VAs latent space                                     & 1                                                                              \\
	HODA-11   & 0.001 seconds                                                    & Vector of 11 integers                                              & 11                                                                             \\
	HODA-5    & 0.001 seconds                                                    & Vector of 5 integers                                               & 5                                                                              \\ \hline
\end{tabular}
}
\end{table}
\textcolor{\changecolor}{
Table \ref{tab:csd} compares PRADA, VarDetect, HODA-11, and HODA-5 in terms of the average runtime, average memory consumption, and the number of model predictions to defend CIFAR10 target classifier in $num_s=500$ on Tesla K80 GPU. Since HODA does not need to store samples in input space or latent space,  the memory consumption of HODA is significantly less than PRADA and VarDetect for each user. HODA needs to store a vector of integers for each user, representing the hardness degree histogram. The runtime of HODA also is several times smaller than PRADA and VarDetect. Besides, the runtime of PRADA and VarDetect increases by increasing $num_S$, but the runtime of HODA is constant for any $num_s$.
Although HODA-5 (HODA-11) requires the predictions of 5 (11) models to calculate the hardness degree of each sample, there is no sequential relationship between models, and they can predict in parallel, so HODA does not increase the prediction time of target models. 
Overall, the main drawback of HODA is in the number of predictions that can be solved by providing several GPUs. On the other side, memory consumption and runtime of VarDetect and PRADA are extremely more than HODA for each user. Hence, HODA is more scalable than PRADA and VarDetect with respect to the number of users.}

%% file: content/hodadrcifar.tex
\begin{table}
	\caption{The detection rate, AUC score, and False Positive Rate (FPR) of PRADA, VarDetect, HODA-5, and HODA-11 against four various model extraction attacks on CIFAR10 and CIFAR100 target classifiers.}

	\label{tab:hbdmperfmea}
	\renewcommand{\arraystretch}{1.7}
	\resizebox{\linewidth}{!}{
		\begin{tabular}{ccccccccc}
			\hline
			&                        &         &                       &         & \multicolumn{4}{c}{Detection Rate of Attacks \% (AUC Score)} \\ \cline{6-9} 
			$f_t$&          Defense              & $num_s$ & $\delta$ & FPR(\%) & JBDA    & JBRAND   & K.Net CIFARX   & K.Net TIN   \\ \hline
			\multirow{8}{*}{\rotatebox[]{90}{CIFAR10}}  & \multirow{2}{*}{PRADA} & \textbf{100}     & 0.818                 & 0.01    & 0 (0.512)       & 0 (0.494)       & 0 (0.541)             & 0 (0.495)           \\
			&                        & 500     & 0.973                 & 0.05    & 96.7 (0.999)   & 94.2 (0.996)    & 4.4 (0.857)            & 1.6 (0.766)        \\ \cline{3-9} 
			& \multirow{2}{*}{VarDetect}  & \textbf{100}      &    0.283             &  0.15  &   100 (1)   &   100 (1)    &    0.66  (0.698)     &   16.58 (0.943)     \\
			&                        & 500     &      0.101            &  0.13  &   100 (1)  &  100 (1)  &     73.82 (0.995)    &  100 (0.999)      \\ \cline{3-9} 
			& \multirow{2}{*}{HODA-11}  & 50      & 0.290                 & 0.02    & 100 (1)    & 100 (1)      & 99.92 (0.999)         & 99.73 (0.999)     \\
			&                        & \textbf{100}     & 0.154                 & 0.02    & 100 (1)    & 100 (1)    & 100 (1)            & 100 (1)         \\ \cline{3-9} 
			& \multirow{2}{*}{HODA-5}     & \textbf{100}     &     0.044             &   0.01  &  100 (1)  &  100 (1)  &     93.82 (0.998)       &  93.86  (0.998)   \\ 
			&                        & 200     &   0.017              &   0.02  &   100 (1)  &   100 (1)  &     99.27 (0.999)       &     99.15  (0.999)     \\ \hline
			\multirow{8}{*}{\rotatebox[]{90}{CIFAR100}} & \multirow{2}{*}{PRADA} & 500     & 0.550                 & 0.01    & 0 (0.518)      & 0 (0.528)       & 0 (0.319)              & 0 (0.366)       \\
			&                        & 1000    & 0.953                 & 0.03    & 67.3 (0.997)    & 73.5 (0.995)    & 0 (0.126)             & 0  (0.277)         \\ \cline{3-9} 
			& \multirow{2}{*}{VarDetect}  & \textbf{100}      &       0.223          &  0.09  &  100 (1)    &  100 (1)     &    1.76 (0.802)       &    7.38 (0.906)    \\
			&                        & 500     &      0.101            & 0.13   & 100 (1)    &  100 (1)  &     93.08 (0.999)    &    99.91 (0.999)    \\ \cline{3-9} 
			& \multirow{2}{*}{HODA-11}  & 50      & 0.716                 & 0.02    & 94.65 (0.999)  & 100  (1)    & 90.68 (0.999)          & 89.06 (0.999)      \\
			&                        & \textbf{100}     & 0.349                 & 0.02    & 100 (1)     & 100 (1)      & 100 (1)          & 100 (1)         \\ \cline{3-9} 
						& \multirow{2}{*}{HODA-5} & \textbf{100}    &      0.140            & 0.02    & 100 (1) &  100  (1) &         97.08 (0.999)     &        97.79 (0.999)   \\ 	
						&                        & 200    &     0.060             &  0.01   &     100 (1)&  100 (1)  &       99.87 (0.999)     &    99.85  (0.999)     \\ \cline{1-9} 
		\end{tabular}
	}
\end{table}

%% file: content/transferlearning.tex
\input{content/validtransferandnotransfer.tex}

\begin{table}
	\caption{The detection rate, AUC score, and False Positive Rate (FPR) of HODA-5 and HODA-11 against K.Net ILSVRC12 attack.}
	\label{tab:tfhbdmdetection}
	\renewcommand{\arraystretch}{1.1}
	\centering
	\resizebox{0.8\linewidth}{!}{
		\begin{tabular}{cccccc}
			\hline
			\multicolumn{5}{c}{}                                                                 & Detection Rate \% (AUC Score) \\ \cline{6-6} 
			$f_t$&             Defense                    & $num_s$ & $\delta$ & FPR(\%) & K.Net ILSVRC12                \\ \hline
			\multirow{4}{*}{\rotatebox[]{90}{CUB200}}     & \multirow{2}{*}{HODA-11}   & 50      & 0.973    & 0.01    & 97.5  (0.999)                        \\
			&                         & 100     & 0.393    & 0.02    & 100   (1)                       \\ \cline{3-6} 
			& \multirow{2}{*}{HODA-5} & 50      &    0.227      &     0.01    &     99.7 (0.999)                          \\
			&                         & 100     &    0.087      &   0.01      &      100 (1)                         \\ \hline
			\multirow{4}{*}{\rotatebox[]{90}{Caltech256}} & \multirow{2}{*}{HODA-11}   & 50      & 0.838    & 0.01    & 99.98 (0.999)             \\
			&                         & 100     & 0.420    & 0.01    & 100  (1)                         \\ \cline{3-6} 
			& \multirow{2}{*}{HODA-5} & 50      &    0.263      &      0.02   &        97.3  (0.999)                     \\
			&                         & 100     &    0.110      &    0.01     &       99.87 (0.999)                        \\ \hline
		\end{tabular}
		
	}
\end{table}

Transfer learning is a technique that initializes the parameters of the target task classifier using the parameters of a pre-trained source task classifier.
We train two new target classifiers on CUB200 and Caltech256 datasets using transfer learning (details of datasets in Appendix \ref{sec:appa}).
The training process of new target classifiers is the same as  CIFAR10 and CIFAR100 target classifiers (Section \ref{sec:hsbase}). We initialize the parameters of target classifiers from a pre-trained ImageNet \cite{DBLP:conf/cvpr/DengDSLL009} classifier and train all layers of target classifiers. \cite{DBLP:conf/iclr/OrekondySF20} indicate that jacobian-based model extraction attacks have very poor performance on high dimensional datasets. Thereby, we only evaluate the performance of target classifiers against K.Net ILSVRC12 attack.  K.Net ILSVRC12 is the Knockoff Net attack that uses ILSVRC12 dataset as the surrogate classifier's training set. The budget of K.Net ILSVRC12 is 50000, and the output of target classifiers is the entire probability vector. The accuracy of CUB200 target classifier and its surrogate classifier is 73.7\% and 59.3\%, respectively, and the accuracy of Caltech256 target classifier and its surrogate classifier is 77.2\% and 72.2\%, respectively.

Figure \ref{fig:hardtlhist} depicts the hardness degree histogram of CUB200 and Caltech256 test sets on the associated target classifier and also the hardness degree histogram of K.Net ILSVRC12 samples for both target classifiers. The figure demonstrates that the majority number of  K.Net ILSVRC12 attack samples are hard (hardness degree $>$ 70), and the number of easy samples (hardness degree $<$ 30) is very small.
We replicate the experiment of the previous section to evaluate the performance of HODA against K.Net ILSVRC12 attack with the same parameters. Table \ref{tab:tfhbdmdetection} shows the performance of HODA-11 and HODA-5 against K.Net ILSVRC12 attack on both target classifiers. The results demonstrate that even the starting point of target classifiers' parameters is not random, HODA is very effective in detecting K.Net ILSVRC12 attack.

%% file: content/validtransferandnotransfer.tex
\begin{figure}
	\pgfplotsset{every tick label/.append style={font=\tiny}}
	\begin{subfigure}{0.49\linewidth}
		\centering
		\resizebox{\linewidth}{!}{
		\begin{tikzpicture}
\begin{groupplot}[group style={group size= 2 by 2,horizontal sep=0.8cm},height=4cm,width=4.8cm,tick label style = {font = {\fontsize{8 pt}{12 pt}\selectfont}},,ytick align=outside,xtick style={draw=none},tick label style={font=\scriptsize},,ytick pos=left,enlargelimits=true,enlargelimits=0.02,ylabel shift = -8 pt,xlabel = Hardness Degree,,title style={yshift=-0.2cm}]
\nextgroupplot[ymin=0, ymax=1000,title=Test set]
\addplot
[ybar interval,mark=no,fill=purple] 
coordinates
{
(0,14)(1,106)(2,73)(3,181)(4,224)(5,149)(6,115)(7,91)(8,175)(9,124)(10,107)(11,140)(12,71)(13,87)(14,74)(15,192)(16,125)(17,68)(18,84)(19,81)(20,67)(21,47)(22,75)(23,88)(24,91)(25,52)(26,45)(27,37)(28,45)(29,53)(30,39)(31,50)(32,48)(33,36)(34,49)(35,45)(36,46)(37,50)(38,45)(39,23)(40,52)(41,48)(42,40)(43,78)(44,53)(45,36)(46,25)(47,33)(48,30)(49,40)(50,40)(51,36)(52,33)(53,37)(54,48)(55,19)(56,44)(57,58)(58,47)(59,34)(60,32)(61,34)(62,36)(63,35)(64,20)(65,53)(66,28)(67,20)(68,26)(69,29)(70,35)(71,34)(72,30)(73,26)(74,22)(75,29)(76,25)(77,24)(78,38)(79,43)(80,18)(81,26)(82,38)(83,31)(84,27)(85,24)(86,30)(87,36)(88,45)(89,39)(90,33)(91,41)(92,41)(93,50)(94,38)(95,36)(96,64)(97,94)(98,114)(99,177)(100,0)
} 
\closedcycle;
\nextgroupplot[ymin=0,ymax=8000,title=K.Net ILSVRC12,ytick={0,2000,4000,6000,8000},yticklabels={0,2K,4K,6K,8K},scaled y ticks = false,xlabel = Hardness Degree,]
\addplot
[ybar interval,mark=no,fill=purple] 
coordinates
{
	(0,0)(1,4)(2,2)(3,15)(4,23)(5,15)(6,21)(7,11)(8,21)(9,38)(10,28)(11,27)(12,28)(13,30)(14,34)(15,48)(16,63)(17,52)(18,45)(19,52)(20,39)(21,80)(22,79)(23,66)(24,87)(25,83)(26,93)(27,95)(28,106)(29,147)(30,87)(31,133)(32,164)(33,147)(34,119)(35,206)(36,183)(37,259)(38,161)(39,188)(40,268)(41,315)(42,269)(43,411)(44,392)(45,341)(46,222)(47,273)(48,323)(49,317)(50,467)(51,406)(52,370)(53,434)(54,632)(55,475)(56,619)(57,514)(58,335)(59,464)(60,578)(61,551)(62,625)(63,559)(64,516)(65,629)(66,465)(67,552)(68,467)(69,604)(70,797)(71,601)(72,587)(73,598)(74,567)(75,573)(76,666)(77,745)(78,862)(79,758)(80,740)(81,602)(82,524)(83,641)(84,826)(85,834)(86,752)(87,828)(88,1073)(89,914)(90,1020)(91,928)(92,983)(93,1188)(94,1357)(95,1250)(96,1429)(97,2574)(98,3126)(99,5185)(100,0)
} 
\closedcycle;
	\end{groupplot}
\end{tikzpicture}
}
\vspace{-1.5\baselineskip}
\caption{ CUB200}
\end{subfigure}
\begin{subfigure}{0.49\linewidth}
\centering
\resizebox{\linewidth}{!}{
\begin{tikzpicture}
\begin{groupplot}[group style={group size= 4 by 1,,horizontal sep=0.8cm},height=4cm,width=4.8cm,tick label style = {font = {\fontsize{8 pt}{12 pt}\selectfont}},,ytick align=outside,xtick style={draw=none},tick label style={font=\scriptsize},,ytick pos=left,enlargelimits=true,enlargelimits=0.02,,ylabel shift = -8 pt,,title style={yshift=-0.2cm}]
\nextgroupplot[ymin=0,ymax=1000,xlabel style={align=center},xlabel = Hardness Degree,title=Test set]
\addplot
[ybar interval,mark=no,fill=purple] 
coordinates
{
(0,641)(1,226)(2,150)(3,108)(4,155)(5,70)(6,59)(7,37)(8,63)(9,44)(10,66)(11,43)(12,53)(13,63)(14,87)(15,57)(16,47)(17,47)(18,47)(19,32)(20,42)(21,63)(22,50)(23,55)(24,45)(25,31)(26,37)(27,59)(28,37)(29,44)(30,24)(31,78)(32,59)(33,48)(34,54)(35,31)(36,37)(37,46)(38,45)(39,39)(40,24)(41,36)(42,37)(43,40)(44,39)(45,33)(46,47)(47,27)(48,30)(49,46)(50,31)(51,44)(52,46)(53,28)(54,32)(55,34)(56,27)(57,27)(58,38)(59,38)(60,34)(61,23)(62,35)(63,33)(64,24)(65,38)(66,27)(67,37)(68,25)(69,27)(70,34)(71,26)(72,32)(73,30)(74,42)(75,23)(76,39)(77,24)(78,49)(79,32)(80,36)(81,30)(82,38)(83,25)(84,46)(85,47)(86,40)(87,32)(88,46)(89,30)(90,54)(91,68)(92,59)(93,59)(94,71)(95,80)(96,103)(97,110)(98,173)(99,358)(100,0)
};
\nextgroupplot[ymin=0,ymax=8000,xlabel = Hardness Degree,ytick={0,2000,4000,6000,8000},yticklabels={0,2K,4K,6K,8K},scaled y ticks = false,title=K.Net ILSVRC12]
\addplot
[ybar interval,mark=no,fill=purple,] 
coordinates
{
	(0,1)(1,233)(2,113)(3,79)(4,121)(5,86)(6,46)(7,46)(8,54)(9,41)(10,51)(11,79)(12,47)(13,63)(14,70)(15,65)(16,67)(17,47)(18,71)(19,75)(20,91)(21,80)(22,86)(23,128)(24,95)(25,142)(26,132)(27,122)(28,95)(29,170)(30,77)(31,176)(32,203)(33,190)(34,209)(35,151)(36,238)(37,162)(38,305)(39,195)(40,288)(41,207)(42,244)(43,256)(44,231)(45,225)(46,246)(47,242)(48,264)(49,286)(50,338)(51,389)(52,308)(53,292)(54,339)(55,439)(56,323)(57,269)(58,346)(59,434)(60,335)(61,390)(62,371)(63,465)(64,392)(65,469)(66,475)(67,460)(68,463)(69,462)(70,503)(71,390)(72,480)(73,424)(74,640)(75,493)(76,559)(77,621)(78,569)(79,579)(80,651)(81,568)(82,574)(83,588)(84,678)(85,775)(86,783)(87,729)(88,766)(89,816)(90,935)(91,1033)(92,1059)(93,1262)(94,1550)(95,1761)(96,2004)(97,2483)(98,3658)(99,7319)(100,0)
};
\end{groupplot}
\end{tikzpicture}
}
\vspace{-1.5\baselineskip}
\caption{ Caltech256}
\end{subfigure}
\vspace{-0.6\baselineskip}
\caption{The left histograms in subfigures (a) and (b) show the hardness degree histogram of CUB200 and Caltech256 test samples, respectively. The right histogram in each subfigure indicates the hardness degree histograms of K.Net ILSVRC12 attack samples on CUB200 (a) and Caltech256 (b) target classifiers.}
\label{fig:hardtlhist}
\end{figure}
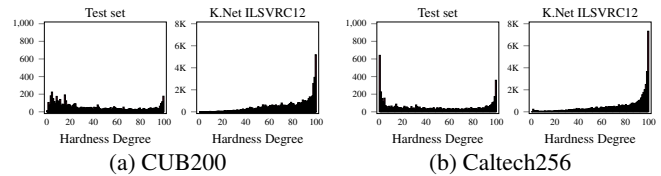

%% file: content/duscussion.tex
\comment{An adaptive adversary who is aware of HODA can potentially modify her attack to evade it. Fundamentally, model extraction attacks can evade HODA if they have access to the samples that come from the target classifier's training data distribution. However, this is a strong threat model, and in most studies, it is supposed that there is no such access \cite{DBLP:conf/cvpr/OrekondySF19,10.1145/3052973.3053009,DBLP:conf/eurosp/JuutiSMA19}. An adaptive adversary must send her queries based on the hardness degree histogram of normal samples to evade HODA. There are two reasons why such attack is hard to conduct.}

An adaptive adversary who is aware of HODA must send her queries based on the hardness degree histogram of normal samples to evade HODA.  We consider two scenarios for an adaptive adversary to conduct model extraction attacks. In the first scenario, the adversary has no access to normal samples, and she only can use synthetic or semantically similar samples to extract the target model. There are two reasons why such attacks are hard to conduct.
First, the adversary needs samples with various degrees of hardness; however, since the adversary has no access to the target classifier, she can not determine the hardness degree of her samples for the target classifier. 
Second, the adversary has no access to the histogram of normal samples to generate her samples based on it.

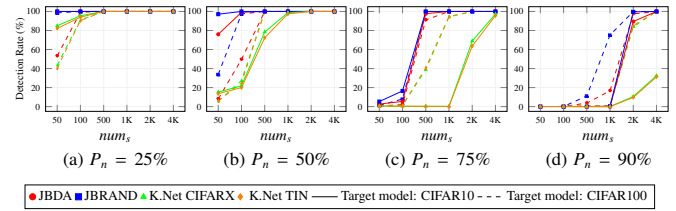
\begin{figure}[!t]
	\pgfplotsset{every tick label/.append style={font=\tiny}}
	\resizebox{\linewidth}{!}{
		\begin{tikzpicture}
		\begin{groupplot}[group style={group name=my plots,group size= 4 by 1,horizontal sep=0.5cm},height=4cm,width=4.8cm,
		xlabel=$num_s$,
		symbolic x coords= {50,100,500,1000,2000,4000},
		xtick={50,100,500,1000,2000,4000},
		xticklabels={50,100,500,1K,2K,4K},
		ytick = {0,20,40,60,80,100},
		ymin = -5,
		ymax = 105,
		grid=major,
		grid style={black!5},ylabel shift=-6pt,]
		\nextgroupplot[xlabel style={align=center,text width=3cm},xlabel=$num_s$\\ \vspace{0.2cm}\large{(a) $P_n = 25\%$},ylabel=\scriptsize{Detection Rate (\%)},]
		\addplot [dashed,red,mark=*,mark size=1] table [x={nums}, y={sev}] {res/HODA_cifar100_jbda_prob_50000.dat};
		\addplot [dashed,blue,mark=square*,mark size=1] table [x={nums}, y={sev}] {res/HODA_cifar100_jbrand_prob_50000.dat};
		\addplot [dashed,green,mark=triangle*,mark size=1] table [x={nums}, y={sev}] {res/HODA_cifar100_k.net_cifarx_prob_50000.dat};
		\addplot [dashed,orange,mark=diamond*,mark size=1] table [x={nums}, y={sev}] {res/HODA_cifar100_k.net_tin_prob_50000.dat};
		
		\addplot [red,mark=*,mark size=1] table [x={nums}, y={sev}] {res/HODA_cifar10_jbda_prob_50000.dat};
		\addplot [blue,mark=square*,mark size=1] table [x={nums}, y={sev}] {res/HODA_cifar10_jbrand_prob_50000.dat};
		\addplot [green,mark=triangle*,mark size=1] table [x={nums}, y={sev}] {res/HODA_cifar10_k.net_cifarx_prob_50000.dat};
		\addplot [orange,mark=diamond*,mark size=1] table [x={nums}, y={sev}] {res/HODA_cifar10_k.net_tin_prob_50000.dat};

		\nextgroupplot[xlabel style={align=center,text width=3cm},xlabel=$num_s$\\ \vspace{0.2cm}\large{(b) $P_n = 50\%$}]
		\addplot [dashed,red,mark=*,mark size=1] table [x={nums}, y={fif}] {res/HODA_cifar100_jbda_prob_50000.dat};
		\addplot [dashed,blue,mark=square*,mark size=1] table [x={nums}, y={fif}] {res/HODA_cifar100_jbrand_prob_50000.dat};
		\addplot [dashed,green,mark=triangle*,mark size=1] table [x={nums}, y={fif}] {res/HODA_cifar100_k.net_cifarx_prob_50000.dat};
		\addplot [dashed,orange,mark=diamond*,mark size=1] table [x={nums}, y={fif}] {res/HODA_cifar100_k.net_tin_prob_50000.dat};
		
		\addplot [red,mark=*,mark size=1] table [x={nums}, y={fif}] {res/HODA_cifar10_jbda_prob_50000.dat};
		\addplot [blue,mark=square*,mark size=1] table [x={nums}, y={fif}] {res/HODA_cifar10_jbrand_prob_50000.dat};
		\addplot [green,mark=triangle*,mark size=1] table [x={nums}, y={fif}] {res/HODA_cifar10_k.net_cifarx_prob_50000.dat};
		\addplot [orange,mark=diamond*,mark size=1] table [x={nums}, y={fif}] {res/HODA_cifar10_k.net_tin_prob_50000.dat};

		\nextgroupplot[xlabel style={align=center,text width=3cm},xlabel=$num_s$\\ \vspace{0.2cm}\large{(c) $P_n = 75\%$},
		legend style={at={(-0.2,-0.70)},
			anchor=north,legend columns=-6},
		]
		\addlegendimage{red, mark = *, ,only marks, mark size= 2pt}
		\addlegendimage{blue, mark = square*, ,only marks, mark size= 2pt}
		\addlegendimage{green, mark = triangle*,,only marks,  mark size= 2pt}
		\addlegendimage{orange,only marks, mark = diamond*, mark size= 2pt}
		\addlegendimage{mark size= 0.5pt}	\addlegendimage{dashed, mark size= 0.5pt}
		
		\addplot [dashed,red,mark=*,mark size=1] table [x={nums}, y={twe}] {res/HODA_cifar100_jbda_prob_50000.dat};
		\addplot [dashed,blue,mark=square*,mark size=1] table [x={nums}, y={twe}] {res/HODA_cifar100_jbrand_prob_50000.dat};
		\addplot [dashed,green,mark=triangle*,mark size=1] table [x={nums}, y={twe}] {res/HODA_cifar100_k.net_cifarx_prob_50000.dat};
		\addplot [dashed,orange,mark=diamond*,mark size=1] table [x={nums}, y={twe}] {res/HODA_cifar100_k.net_tin_prob_50000.dat};
		
		\addplot [red,mark=*,mark size=1] table [x={nums}, y={twe}] {res/HODA_cifar10_jbda_prob_50000.dat};
		\addplot [blue,mark=square*,mark size=1] table [x={nums}, y={twe}] {res/HODA_cifar10_jbrand_prob_50000.dat};
		\addplot [green,mark=triangle*,mark size=1] table [x={nums}, y={twe}] {res/HODA_cifar10_k.net_cifarx_prob_50000.dat};
		\addplot [orange,mark=diamond*,mark size=1] table [x={nums}, y={twe}] {res/HODA_cifar10_k.net_tin_prob_50000.dat};

		\legend{  \footnotesize{JBDA}, \footnotesize{JBRAND}, \footnotesize{K.Net CIFARX},\footnotesize{ K.Net TIN}, \footnotesize{Target model: CIFAR10},\footnotesize{Target model: CIFAR100 } }
		
		\nextgroupplot[xlabel style={align=center,text width=3cm},xlabel=$num_s$\\ \vspace{0.2cm} \large{(d) $P_n = 90\%$}]
		\addplot [dashed,red,mark=*,mark size=1] table [x={nums}, y={ten}] {res/HODA_cifar100_jbda_prob_50000.dat};
		\addplot [dashed,blue,mark=square*,mark size=1] table [x={nums}, y={ten}] {res/HODA_cifar100_jbrand_prob_50000.dat};
		\addplot [dashed,green,mark=triangle*,mark size=1] table [x={nums}, y={ten}] {res/HODA_cifar100_k.net_cifarx_prob_50000.dat};
		\addplot [dashed,orange,mark=diamond*,mark size=1] table [x={nums}, y={ten}] {res/HODA_cifar100_k.net_tin_prob_50000.dat};
		
		\addplot [red,mark=*,mark size=1] table [x={nums}, y={ten}] {res/HODA_cifar10_jbda_prob_50000.dat};
		\addplot [blue,mark=square*,mark size=1] table [x={nums}, y={ten}] {res/HODA_cifar10_jbrand_prob_50000.dat};
		\addplot [green,mark=triangle*,mark size=1] table [x={nums}, y={ten}] {res/HODA_cifar10_k.net_cifarx_prob_50000.dat};
		\addplot [orange,mark=diamond*,mark size=1] table [x={nums}, y={ten}] {res/HODA_cifar10_k.net_tin_prob_50000.dat};
		\end{groupplot}
		\end{tikzpicture}
	}

	\caption{The detection rate of HODA for various percentages of normal samples $P_n$ over different values of $num_s$.}
	\label{fig:adaptiveatt}
\end{figure}

In the second scenario, we assume the adversary has access to a limited number of normal samples, and she can use normal samples to make her hardness degree histogram more similar to the hardness degree histogram of normal samples. To evaluate HODA in this scenario, we suppose that the adversary has access to 1000 normal samples from $S_{user}$ and she sends a sample sequence of which $P_n\%$ is filled by normal samples, and the rest is filled by model extraction attack samples. Notably, when the number of normal samples in the sequence exceeds 1000, the adversary sends duplicate normal samples.
It is important to note that the cost of attack is increased by a factor of $\frac{1}{1-(P_n/100)}$ in this scenario.
Figure \ref{fig:adaptiveatt} shows the detection rate of HODA for various $P_n$ over different values of $num_s$. The false-positive rate of all experiments is less than 0.2\%. The figure demonstrates that increasing $num_s$ improves the detection rate of HODA. Except for K.Net attacks on CIFAR10 target classifier in $P_n=90\%$, HODA can detect all attacks with a high success rate by increasing $num_s$. Due to the dataset limitation, we can not evaluate HODA for $num_s>4000$. However, we think the detection rate of HODA against K.Net attacks on CIFAR10 target classifier in $P_n=90\%$ will be improved for $num_s>4000$.
Altogether, we think the main challenge of an adaptive adversary to evade HODA is to collect easy samples, which are very rare in out-of-distribution samples based on our experiments.  

\comment{
We propose HODA to detect sample sequences of model extraction attacks. However, some other policies can be defined using the hardness degree histograms of users' samples. For example, the price of a user's queries can be determined based on the distance among the hardness degree histograms of the user's samples and normal samples to raise the price of model extraction attacks in MLaaS. There are several studies \cite{DBLP:conf/sp/LeeEMS19,DBLP:conf/iclr/OrekondySF20,DBLP:conf/cvpr/KariyappaQ20,kariyappa2021protecting} that propose to perturb the output of the target classifier to decrease the performance of the surrogate classifier. Our method can be mixed with such methods. For example, the size of perturbation applied to the output of target classifier can be related to the distance among the hardness degree histograms of the user's samples and normal samples to increase the utility of perturbation-based methods.
}

%% file: content/conc.tex
This paper demonstrates that the hardness degree of samples is important in trustworthy machine learning. 
We investigated the hardness degree of samples and demonstrated that the hardness degree histogram of model extraction attack samples is different from the hardness degree histogram of normal samples. Using this observation, we proposed Hardness-Oriented Detection Approach (HODA) to detect sample sequences of model extraction attacks. HODA can detect the sample sequences of model extraction attacks with a high success rate by only monitoring 100 samples of attacks. 

%% file: content/datasets.tex
\label{sec:appa}

\input{content/figure/hardnessconf.tex}

\textbf{CIFAR10} \cite{Krizhevsky2009LearningML}: CIFAR-10 dataset consists of 60K $32\times32$ color images in 10 classes, including airplanes, cars, birds, cats, deer, dogs, frogs, horses, ships, and trucks. It has 6K images per class, where 5K images is in the training set and 1K images is in the test set.

\textbf{CIFAR100} \cite{Krizhevsky2009LearningML}:  CIFAR100 dataset consists of 60K $32\times32$ color images in 100 classes. It has 600 images per class, where 500 images is in the training set and 100 images is in the test set. 

\textbf{TinyImageNet} \cite{Le2015TinyIV}: TinyImageNet is a subset of ILSVRC12 \cite{DBLP:conf/cvpr/DengDSLL009} dataset, and contains 200 image classes. It has 500 training samples and 50 test samples for each class. The size of images is $64\times 64$. We resize all images to $32 \times 32$.

\textbf{CUB200} \cite{WahCUB_200_2011}:
CUB200 dataset contain 200 classes of bird categories. It consists of about 6K training and about 6K test samples. The size of images is $224\times224$.

\textbf{Caltech256} \cite{griffin2007caltech}: 
Caltech256 dataset contain 256 classes of common objects categories. It consists of about 24K training and about 6K test samples. The size of images is $224\times224$.

\textbf{ILSVRC12} \cite{DBLP:conf/cvpr/DengDSLL009}:  ILSVRC12 uses a subset of ImageNet and consists of 1.2 million training images, 50,000 validation images, and 100,000 test images. The dataset has 1000 classes and the size of images is $224\times224$.

\textbf{STL10} \cite{DBLP:journals/jmlr/CoatesNL11}: STL10 dataset consists of 13K $96\times96$ color images in 10 classes, including airplanes, cars, birds, cats, deer, dogs, monkeys, horses, ships, and trucks.  It has 1.3K images per class, where 0.5K images is in the training set and 0.8K images is in the test set. We resize all images to $32 \times 32$.

%% file: content/figure/hardnessconf.tex
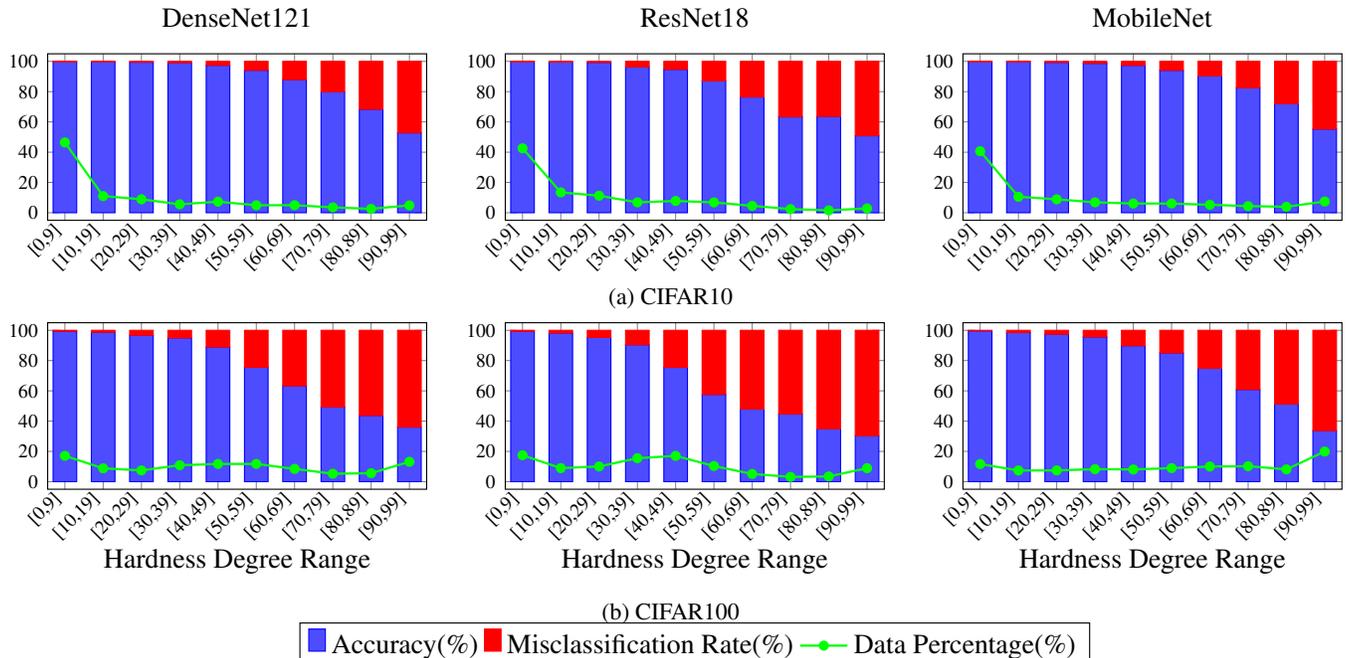
\begin{figure*}
	\begin{subfigure}{\linewidth}
		\centering
		\resizebox{\linewidth}{!}{
		\begin{tikzpicture}
		\begin{groupplot}[group style={group size= 3 by 1,vertical sep=1.2cm},height=3.7cm,width=6.4cm,tick label style = {font = {\fontsize{7pt}{12 pt}\selectfont}},]
		\nextgroupplot[
		enlargelimits=0.05,
		legend style={at={(1,-1)},anchor=north,legend columns=-1},
		ytick={0,20,40,60,80,100},
		symbolic x coords={0-9,10-19,20-29,30-39,40-49,50-59,60-69,70-79,80-89,90-99},
		xticklabels ={{[0,9]},{[10,19]},{[20,29]},{[30,39]},{[40,49]},{[50,59]},{[60,69]},{[70,79]},{[80,89]},{[90,99]}},
		xtick=data,
		x tick label style={rotate=45,anchor=east},
		bar width=0.3cm,
		ymin = 0,
		title= DenseNet121
		]
		\addplot+[ybar stacked,ybar,fill=blue!70,mark=none] plot coordinates {(0-9,99.85)(10-19,99.91)(20-29,99.44)(30-39,99.1)(40-49,97.4)(50-59,94.09)(60-69,87.93)(70-79,79.89)(80-89,68.36)(90-99,52.77)
		};
		\addplot+[ybar stacked,fill=red,ybar,mark=none] plot coordinates {(0-9,0.15)(10-19,0.09)(20-29,0.56)(30-39,0.9)(40-49,2.6)(50-59,5.91)(60-69,12.07)(70-79,20.11)(80-89,31.64)(90-99,47.23)};
		\addplot[green,mark=*,mark size = 0.05cm,thick] 
		coordinates {(0-9,46.44)(10-19,11.0)(20-29,8.88)(30-39,5.54)(40-49,7.3)(50-59,4.91)(60-69,4.97)(70-79,3.53)(80-89,2.56)(90-99,4.87)};

		\nextgroupplot[
		height= 3.7cm,
		width = 6.4cm,
		enlargelimits=0.05,
		legend style={at={(0.5,-0.20)},
			anchor=north,legend columns=-1},
		ytick={0,20,40,60,80,100},
		symbolic x coords={0-9,10-19,20-29,30-39,40-49,50-59,60-69,70-79,80-89,90-99},
				xticklabels ={{[0,9]},{[10,19]},{[20,29]},{[30,39]},{[40,49]},{[50,59]},{[60,69]},{[70,79]},{[80,89]},{[90,99]}},
		xtick=data,
		x tick label style={rotate=45,anchor=east},
		bar width=0.3cm,
		ymin = 0,
		title= ResNet18,
		]
		\addplot+[ybar stacked,ybar,fill=blue!70,mark=none] plot coordinates {
			(0-9,99.88)(10-19,99.55)(20-29,99.29)(30-39,96.35)(40-49,94.73)(50-59,87.1)(60-69,76.48)(70-79,63.33)(80-89,63.64)(90-99,50.88)
		};
		\addplot+[ybar stacked,fill=red,ybar,mark=none] plot coordinates {
			(0-9,0.12)(10-19,0.45)(20-29,0.71)(30-39,3.65)(40-49,5.27)(50-59,12.9)(60-69,23.52)(70-79,36.67)(80-89,36.36)(90-99,49.12)};
		\addplot[green,mark=*,mark size = 0.05cm,thick] 
		coordinates {(0-9,42.56)(10-19,13.43)(20-29,11.23)(30-39,6.84)(40-49,7.78)(50-59,6.82)(60-69,4.55)(70-79,2.4)(80-89,1.54)(90-99,2.85)
		};

		\nextgroupplot[
		tick label style={/pgf/number format/fixed},
		every node near coord/.append style={font=\tiny},
		height= 3.7cm,
		width = 6.4cm,
		enlargelimits=0.05,
		legend style={at={(0.5,-0.20)},
			anchor=north,legend columns=-1},
		ytick={0,20,40,60,80,100},
		symbolic x coords={0-9,10-19,20-29,30-39,40-49,50-59,60-69,70-79,80-89,90-99},
				xticklabels ={{[0,9]},{[10,19]},{[20,29]},{[30,39]},{[40,49]},{[50,59]},{[60,69]},{[70,79]},{[80,89]},{[90,99]}},
		xtick=data,
		x tick label style={rotate=45,anchor=east},
		bar width=0.3cm,
		ymin = 0,
		title= MobileNet
		]
		\addplot+[ybar stacked,ybar,fill=blue!70,mark=none] plot coordinates {(0-9,99.88)(10-19,99.81)(20-29,99.32)(30-39,98.69)(40-49,97.37)(50-59,94.06)(60-69,90.48)(70-79,82.57)(80-89,72.01)(90-99,55.27)
		};
		\addplot+[ybar stacked,fill=red,ybar,mark=none] plot coordinates {
			(0-9,0.12)(10-19,0.19)(20-29,0.68)(30-39,1.31)(40-49,2.63)(50-59,5.94)(60-69,9.52)(70-79,17.43)(80-89,27.99)(90-99,44.73)};
		\addplot[green,mark=*,mark size = 0.05cm,thick] 
		coordinates {(0-9,40.65)(10-19,10.54)(20-29,8.86)(30-39,6.87)(40-49,6.08)(50-59,6.06)(60-69,5.25)(70-79,4.36)(80-89,3.93)(90-99,7.4)
		};
		\end{groupplot}
		\end{tikzpicture}
		}
		\vspace{-0.6cm}
		\caption{ CIFAR10}
	\end{subfigure}
	\begin{subfigure}{\linewidth}
		\centering
		\resizebox{\linewidth}{!}{
		\begin{tikzpicture}
		\begin{groupplot}[group style={group size= 3 by 1,vertical sep=1.2cm},height=3.7cm,width=6.4cm,tick label style = {font = {\fontsize{7pt}{12 pt}\selectfont}},]
		\nextgroupplot[
		enlargelimits=0.05,
		legend style={at={(0.5,-0.20)},
			anchor=north,legend columns=-1},
		ytick={0,20,40,60,80,100},
		symbolic x coords={0-9,10-19,20-29,30-39,40-49,50-59,60-69,70-79,80-89,90-99},
				xticklabels ={{[0,9]},{[10,19]},{[20,29]},{[30,39]},{[40,49]},{[50,59]},{[60,69]},{[70,79]},{[80,89]},{[90,99]}},
		xtick=data,
		x tick label style={rotate=45,anchor=east},
		bar width=0.3cm,
		ymin = 0,
		xlabel style={yshift=0.2cm},
		xlabel=Hardness Degree Range
		]
		\addplot+[ybar stacked,ybar,fill=blue!70,mark=none] plot coordinates {(0-9,99.41)(10-19,98.75)(20-29,96.77)(30-39,95.02)(40-49,88.97)(50-59,75.6)(60-69,63.31)(70-79,49.43)(80-89,43.75)(90-99,36.05)
		};
		\addplot+[ybar stacked,fill=red,ybar,mark=none] plot coordinates {(0-9,0.59)(10-19,1.25)(20-29,3.23)(30-39,4.98)(40-49,11.03)(50-59,24.4)(60-69,36.69)(70-79,50.57)(80-89,56.25)(90-99,63.95)
		};
		\addplot[green,mark=*,mark size = 0.05cm,thick] 
		coordinates {(0-9,17.07)(10-19,8.81)(20-29,7.44)(30-39,10.85)(40-49,11.7)(50-59,11.72)(60-69,8.45)(70-79,5.24)(80-89,5.6)(90-99,13.12)
		};

		\nextgroupplot[
		enlargelimits=0.05,
		legend style={at={(0.5,-0.80)},
			anchor=north,legend columns=-1},
		ytick={0,20,40,60,80,100},
		symbolic x coords={0-9,10-19,20-29,30-39,40-49,50-59,60-69,70-79,80-89,90-99},
				xticklabels ={{[0,9]},{[10,19]},{[20,29]},{[30,39]},{[40,49]},{[50,59]},{[60,69]},{[70,79]},{[80,89]},{[90,99]}},
		xtick=data,
		x tick label style={rotate=45,anchor=east},
		bar width=0.3cm,
		ymin = 0,
		xlabel style={yshift=0.2cm},
		xlabel=Hardness Degree Range,xlabel style={align=center} 
		]
		\addplot+[ybar stacked,ybar,fill=blue!70,mark=none] plot coordinates {(0-9,99.37)(10-19,98.21)(20-29,95.33)(30-39,90.46)(40-49,75.53)(50-59,57.58)(60-69,48.03)(70-79,44.76)(80-89,34.97)(90-99,30.3)
		};
		\addplot+[ybar stacked,fill=red,ybar,mark=none] plot coordinates {
			(0-9,0.63)(10-19,1.79)(20-29,4.67)(30-39,9.54)(40-49,24.47)(50-59,42.42)(60-69,51.97)(70-79,55.24)(80-89,65.03)(90-99,69.7)
			
		};
		\addplot[green,mark=*,mark size = 0.05cm,thick] 
		coordinates {(0-9,17.49)(10-19,8.95)(20-29,10.06)(30-39,15.51)(40-49,17.04)(50-59,10.35)(60-69,5.08)(70-79,3.15)(80-89,3.46)(90-99,8.91)};
		
		\legend{Accuracy(\%), Misclassification Rate(\%), Data Percentage(\%)}
		
		\nextgroupplot[
		height= 3.7cm,
		width = 6.4cm,
		enlargelimits=0.05,
		legend style={at={(0.5,-0.20)},
			anchor=north,legend columns=-1},
		ytick={0,20,40,60,80,100},
		symbolic x coords={0-9,10-19,20-29,30-39,40-49,50-59,60-69,70-79,80-89,90-99},
				xticklabels ={{[0,9]},{[10,19]},{[20,29]},{[30,39]},{[40,49]},{[50,59]},{[60,69]},{[70,79]},{[80,89]},{[90,99]}},
		xtick=data,
		x tick label style={rotate=45,anchor=east},
		bar width=0.3cm,
		ymin = 0,
		xlabel style={yshift=0.2cm},
		xlabel=Hardness Degree Range
		]
		\addplot+[ybar stacked,ybar,fill=blue!70,mark=none] plot coordinates {(0-9,99.66)(10-19,98.51)(20-29,97.44)(30-39,95.63)(40-49,89.84)(50-59,85.04)(60-69,74.93)(70-79,60.86)(80-89,51.29)(90-99,33.7)
		};
		\addplot+[ybar stacked,fill=red,ybar,mark=none] plot coordinates {
			(0-9,0.34)(10-19,1.49)(20-29,2.56)(30-39,4.37)(40-49,10.16)(50-59,14.96)(60-69,25.07)(70-79,39.14)(80-89,48.71)(90-99,66.3)
		};
		\addplot[green,mark=*,mark size = 0.05cm,thick] 
		coordinates {(0-9,11.64)(10-19,7.39)(20-29,7.43)(30-39,8.24)(40-49,8.07)(50-59,8.96)(60-69,10.01)(70-79,10.22)(80-89,8.13)(90-99,19.91)
		};
		\end{groupplot}
		\end{tikzpicture}
		}
		\vspace{-1.5cm}
		\caption{CIFAR100}
	\end{subfigure}
	\vspace{0.6\baselineskip}
	\caption{ Blue and red bars show the percentage of test samples in each range of hardness degrees, which are correctly or wrongly classified, respectively. For each range of hardness degrees, Data Percentage indicates the percentage of test samples whose hardness degrees are in that range.}
	\label{fig:trust}
\end{figure*}

%% file: content/conf.tex
	To assess the relationship between the hardness degree of samples and the misclassification rate, we compute the hardness degree of CIFAR10 and CIFAR100 test samples and then partition them into ten groups based on their hardness degree. Group $i$ consists of samples that their hardness degree is in range $[i\times10, (i+1)\times10]$. Hence, the first hardness group consists of the easiest samples, and the last hardness group consists of the hardest ones. It is important to note that the number of samples in each group is different. 
	Afterward, we calculate the percentage of samples in each group that is classified incorrectly. Figure \ref{fig:trust} demonstrates that the  misclassification rate is increased by increasing the hardness degree of samples. In other words, there is a strong positive correlation between the hardness degree of samples and the misclassification rate. The figure indicates the percentage of samples in each hardness degree range by a green curve.

	For example, the hardness degree of 40.65\% of CIFAR10 test samples (4065 samples) is in the range [0,9] for MobileNet classifier, from which 99.88\% is classified correctly, or the hardness degree of 7.4\% of CIFAR10 test samples (740 samples) is in the range [90,99] for MobileNet classifier, from which 55.27\% is classified correctly.
More than 99\% and 95\% of samples having hardness degree $<$ 30 are correctly classified in CIFAR10 and CIFAR100 test samples, respectively.  On the other side,  less than 55\% and 36\% of samples having hardness degree $\geq$ 90 are correctly classified in CIFAR10 and CIFAR100 test samples, respectively.

%% file: content/classifiers.tex
\label{sec:HDD}

\begin{table}
	\caption{Pearson correlation coefficients between hardness degree of CIFAR10 and CIFAR100 test samples for various pairs of classifiers.}
	\label{tab:corrcls}
	\renewcommand{\arraystretch}{1.2}
	\resizebox{\linewidth}{!}{
\begin{tabular}{lcc}
	& \multicolumn{2}{c}{Pearson Correlation Coefficient} \\ \cline{2-3} 
	& CIFAR10                  & CIFAR100                 \\ \hline
	ResNet18-ResNet18     & 0.784                    & 0.687                    \\
	ResNet18-DenseNet121  & 0.775                    & 0.685                    \\
	ResNet18-MobileNet    & 0.765                    & 0.688                    \\
	DenseNet121-MobileNet & 0.769                    & 0.706                    \\ \hline
\end{tabular}
	}
\end{table}

In this section, we indicate that the hardness of samples is relatively transferable among various classifiers. We use three classifiers created in Section \ref{sec:hsbase} and a new ResNet18 classifier in this experiment.
Table \ref{tab:corrcls} displays the Pearson correlation coefficients between hardness degree of CIFAR10 and CIFAR100 test samples for various pairs of classifiers. The results demonstrate a positive and strong correlation between the hardness degree of samples for various pairs of classifiers. The results indicate the hardness of samples is relatively transferable between different classifiers. On the other side, it implies that the hardness degree of samples is relatively independent of the architecture of classifiers.

%% file: content/detailattack.tex
\label{sec:attackdetail}

\textbf{Jacobian-Based Dataset Augmentation (JBDA)} \cite{10.1145/3052973.3053009}:
The goal of JBDA attack is to increase the fidelity of the surrogate classifier to the target classifier in order to produce adversarial examples for the target classifier in the black-box setting. The authors assume that the adversary has access to a limited number of normal samples called seed samples. 
JBDA augment seed samples using adversarial examples to improve the performance of surrogate model.
The augmentation process is conducted in multiple rounds.
In the first round, surrogate training set $\mathbb{X}_s$ is initialized by seed samples, and surrogate model $f_s$ is trained on $\mathbb{X}_s$. In the next rounds, sample set $S$ with size $\kappa$ is randomly selected from $\mathbb{X}_s$, and for each $x\in S$, adversarial example $x'$ is created using the following equation:
	\begin{equation}
\begin{split}
x ' = x + \lambda.\text{sign}(J_{f_s}[f_t(x)])
\end{split}
\end{equation}
where $\lambda$ is step size and $J$ is the Jacobian function. 
Afterward, new adversarial examples are labeled by the target model, and they are added to $\mathbb{X}_s$. Lastly, surrogate model $f_s$ is trained on $\mathbb{X}_s$.
\comment{ and each round has the following steps:
\begin{enumerate}
	\item In the first round ($\rho=0$), the target classifier labels the seed samples, and then, the labeled samples are added to surrogate training set $\mathbb{X}_s$. In the next rounds ($\rho>0$), new unlabeled perturbed samples ($\mathbb{S}_s^{\rho+1} \mathbin{/}  \mathbb{S}_s^\rho $) are labeled by the target classifier and then, they are added to surrogate training set $\mathbb{X}_s$.
	\item Surrogate classifier $f_s$ is trained on $\mathbb{X}_s$.
	\item For each sample in $\mathbb{S}_s \subset \mathbb{X}_s$ (size $\mathbb{S}_s$ is $\kappa$), a new perturbed sample is created according to the following equation and is added to $\mathbb{S}_s$.
	\begin{equation}
	\begin{split}
	\mathbb{S}_s^{\rho+1} = \mathbb{S}_s^\rho \cup \{x + \lambda_{\rho+1}.\text{sign}(J_{f_s}[f_t(x)]) : x \in \mathbb{S}_s^\rho
	\}
	\end{split}
	\end{equation}
	where $\lambda$ is step size and $J$ is jacobian function. 
\end{enumerate}}
The attack is implemented with  $\lambda=0.1$ and $\kappa=2000$. The seed samples are selected from the test set of datasets. We use 500 (50 for each class) and 1000 (10 for each class) samples of CIFAR10 and CIFAR100 test sets for seed samples, respectively.

\textbf{Jacobian-Based Random Target (JBRAND)} \cite{DBLP:conf/eurosp/JuutiSMA19}:
The goal of JBRAND is to improve the performance of JBDA. It perturbs each sample in multiple iterations to generate more powerful adversarial examples and generates targeted adversarial examples with random targets. We generate three adversarial examples with random targets for each sample and use the same seed samples as JBDA. Each sample is perturbed in five iterations with $\epsilon = \frac{64}{225\times5}$. The attack is implemented with  $\lambda=\frac{64}{255}$ and $\kappa=2000$.

\textbf{Knockoff Net (K.Net)} \cite{DBLP:conf/cvpr/OrekondySF19}: Knockoff Net attack uses large public datasets that are semantically similar to the target model's training samples to create the surrogate model's training set. It has adaptive and random strategies to select the surrogate classifier's training set, which both use semantically similar samples. Since the adaptive strategy has very marginal benefits, we only consider the random strategy to implement this attack. K. Net randomly selects a subset of a public dataset and labels them using the target classifier to create $\mathbb{X}_s$. Finally, it uses $\mathbb{X}_s$ to train surrogate classifier $f_s$.

%% file: content/secattackhistdensenet.tex
\label{sec:histattackmeadensenet}

\begin{figure}[!t]
	\begin{subfigure}{\linewidth}
		\centering
		\resizebox{\linewidth}{!}{
		\begin{tikzpicture}
		\begin{groupplot}[group style={group size= 4 by 2,vertical sep=0.9cm},height=4cm,width=4.8cm,tick label style = {font = {\fontsize{8 pt}{12 pt}\selectfont}},,ytick align=outside,xtick style={draw=none},tick label style={font=\scriptsize},,ytick pos=left,enlargelimits=true,enlargelimits=0.02,]
		\nextgroupplot[ymin=0, ymax=10000,title=JBDA,ylabel=\# of samples,ytick={0,2000,4000,6000,8000,10000},yticklabels={0,2K,4K,6K,8K,10K},scaled y ticks = false]
		\addplot
		[ybar interval,mark=no,fill=purple] 
		coordinates
		{
			
(0,194)(1,25)(2,28)(3,19)(4,20)(5,11)(6,14)(7,3)(8,5)(9,7)(10,114)(11,15)(12,34)(13,12)(14,7)(15,12)(16,2)(17,80)(18,3)(19,3)(20,95)(21,28)(22,46)(23,23)(24,16)(25,21)(26,122)(27,31)(28,47)(29,13)(30,53)(31,7)(32,375)(33,56)(34,109)(35,83)(36,197)(37,43)(38,11)(39,11)(40,129)(41,17)(42,3)(43,99)(44,229)(45,712)(46,5)(47,98)(48,68)(49,17)(50,100)(51,22)(52,60)(53,24)(54,2542)(55,586)(56,92)(57,53)(58,30)(59,29)(60,2058)(61,60)(62,19)(63,63)(64,2309)(65,2242)(66,3979)(67,52)(68,193)(69,350)(70,152)(71,33)(72,112)(73,94)(74,467)(75,1211)(76,790)(77,48)(78,270)(79,319)(80,132)(81,9)(82,5590)(83,1648)(84,491)(85,1675)(86,338)(87,1115)(88,99)(89,1738)(90,806)(91,715)(92,334)(93,1741)(94,747)(95,396)(96,2737)(97,1090)(98,998)(99,3370)(100,0)
		} 
		\closedcycle;
		\nextgroupplot[ymin=0, ymax=10000,title=JBRAND,ytick={0,2000,4000,6000,8000,10000},yticklabels={0,2K,4K,6K,8K,10K},scaled y ticks = false]
		\addplot
		[ybar interval,mark=no,fill=purple] 
		coordinates
		{
(0,127)(1,25)(2,25)(3,18)(4,15)(5,10)(6,12)(7,3)(8,4)(9,4)(10,43)(11,7)(12,15)(13,4)(14,3)(15,1)(16,2)(17,17)(18,2)(19,1)(20,11)(21,4)(22,10)(23,11)(24,5)(25,3)(26,22)(27,7)(28,11)(29,7)(30,19)(31,4)(32,257)(33,44)(34,99)(35,46)(36,97)(37,4)(38,5)(39,10)(40,35)(41,12)(42,1)(43,46)(44,139)(45,40)(46,2)(47,52)(48,24)(49,2)(50,59)(51,20)(52,33)(53,35)(54,299)(55,24)(56,3)(57,12)(58,11)(59,5)(60,2422)(61,11)(62,7)(63,63)(64,330)(65,1416)(66,528)(67,10)(68,126)(69,604)(70,56)(71,21)(72,67)(73,22)(74,571)(75,420)(76,185)(77,18)(78,303)(79,509)(80,98)(81,1)(82,4223)(83,2939)(84,315)(85,2576)(86,326)(87,2517)(88,35)(89,2637)(90,2695)(91,472)(92,191)(93,3166)(94,663)(95,335)(96,4626)(97,1543)(98,1469)(99,5116)(100,0)
		} 
		\closedcycle;
		\nextgroupplot[ymin=0,ymax=6000,title=K.Net CIFARX,ytick={0,2000,4000,6000},yticklabels={0,2K,4K,6K},scaled y ticks = false]
		\addplot
		[ybar interval,mark=no,fill=purple] 
		coordinates
		{
(0,319)(1,160)(2,224)(3,243)(4,149)(5,64)(6,51)(7,41)(8,62)(9,26)(10,161)(11,67)(12,85)(13,97)(14,73)(15,30)(16,35)(17,140)(18,68)(19,76)(20,78)(21,79)(22,227)(23,91)(24,157)(25,89)(26,146)(27,105)(28,144)(29,165)(30,125)(31,150)(32,165)(33,111)(34,116)(35,183)(36,185)(37,95)(38,132)(39,181)(40,655)(41,407)(42,255)(43,197)(44,327)(45,288)(46,159)(47,346)(48,420)(49,380)(50,285)(51,335)(52,556)(53,362)(54,528)(55,289)(56,318)(57,512)(58,419)(59,445)(60,762)(61,639)(62,650)(63,451)(64,668)(65,460)(66,1173)(67,685)(68,602)(69,683)(70,952)(71,749)(72,675)(73,712)(74,460)(75,694)(76,537)(77,737)(78,787)(79,675)(80,911)(81,600)(82,663)(83,891)(84,735)(85,969)(86,647)(87,624)(88,804)(89,806)(90,897)(91,1132)(92,1138)(93,847)(94,772)(95,1277)(96,1730)(97,1319)(98,2038)(99,5071)(100,0)
		} 
		\closedcycle;
		\nextgroupplot[ymin=0,ymax=6000,title=K.Net TIN,ytick={0,2000,4000,6000},yticklabels={0,2K,4K,6K},scaled y ticks = false]
		\addplot
		[ybar interval,mark=no,fill=purple] 
		coordinates
		{
(0,482)(1,211)(2,210)(3,356)(4,139)(5,92)(6,74)(7,45)(8,46)(9,23)(10,240)(11,87)(12,183)(13,84)(14,92)(15,51)(16,44)(17,144)(18,74)(19,79)(20,136)(21,93)(22,247)(23,107)(24,159)(25,84)(26,183)(27,158)(28,176)(29,155)(30,136)(31,135)(32,136)(33,107)(34,184)(35,209)(36,175)(37,99)(38,123)(39,150)(40,700)(41,350)(42,209)(43,224)(44,359)(45,251)(46,131)(47,294)(48,407)(49,271)(50,259)(51,325)(52,504)(53,378)(54,530)(55,262)(56,313)(57,402)(58,399)(59,412)(60,785)(61,595)(62,550)(63,432)(64,617)(65,467)(66,1196)(67,659)(68,541)(69,832)(70,872)(71,672)(72,659)(73,655)(74,533)(75,561)(76,475)(77,666)(78,776)(79,727)(80,844)(81,538)(82,632)(83,970)(84,772)(85,990)(86,670)(87,476)(88,930)(89,900)(90,787)(91,1255)(92,1061)(93,791)(94,648)(95,1447)(96,1857)(97,1285)(98,1810)(99,5379)(100,0)
		} 
		\closedcycle;
		\end{groupplot}
		\end{tikzpicture}
	}
		\vspace{-0.6cm}
		\caption{ CIFAR10}
	\end{subfigure}
	\begin{subfigure}{\linewidth}
		\centering
		\resizebox{\linewidth}{!}{
		\begin{tikzpicture}
		\begin{groupplot}[group style={group size= 4 by 1,vertical sep=0.9cm},height=4cm,width=4.8cm,tick label style = {font = {\fontsize{8 pt}{12 pt}\selectfont}},,ytick align=outside,xtick style={draw=none},tick label style={font=\scriptsize},,ytick pos=left,enlargelimits=true,enlargelimits=0.02,]
		\nextgroupplot[ymin=0, ymax=10000,ylabel=\# of samples,xlabel = Hardness Degree,ytick={0,2000,4000,6000,8000,10000},yticklabels={0,2K,4K,6K,8K,10K},scaled y ticks = false]
		\addplot
		[ybar interval,mark=no,fill=purple] 
		coordinates
		{
(0,24)(1,19)(2,44)(3,16)(4,25)(5,15)(6,9)(7,8)(8,4)(9,9)(10,7)(11,6)(12,11)(13,8)(14,20)(15,8)(16,16)(17,17)(18,11)(19,13)(20,7)(21,25)(22,6)(23,14)(24,14)(25,226)(26,46)(27,55)(28,10)(29,159)(30,43)(31,102)(32,569)(33,125)(34,32)(35,123)(36,458)(37,47)(38,243)(39,950)(40,15)(41,665)(42,135)(43,37)(44,144)(45,161)(46,35)(47,843)(48,28)(49,115)(50,1749)(51,102)(52,463)(53,163)(54,149)(55,85)(56,131)(57,82)(58,317)(59,692)(60,1312)(61,212)(62,581)(63,187)(64,930)(65,241)(66,258)(67,661)(68,430)(69,890)(70,62)(71,143)(72,210)(73,124)(74,1803)(75,1631)(76,201)(77,4001)(78,266)(79,665)(80,84)(81,373)(82,310)(83,150)(84,141)(85,581)(86,243)(87,1376)(88,115)(89,893)(90,2963)(91,803)(92,616)(93,3488)(94,1007)(95,502)(96,744)(97,895)(98,1254)(99,8999)(100,0)
		} 
		\closedcycle;
		\nextgroupplot[ymin=0, xlabel =Hardness Degree, ymax=10000,ytick={0,2000,4000,6000,8000,10000},yticklabels={0,2K,4K,6K,8K,10K},scaled y ticks = false]
		\addplot
		[ybar interval,mark=no,fill=purple] 
		coordinates
		{
(0,24)(1,19)(2,44)(3,16)(4,24)(5,15)(6,9)(7,7)(8,4)(9,9)(10,7)(11,6)(12,11)(13,8)(14,10)(15,8)(16,7)(17,15)(18,11)(19,7)(20,7)(21,13)(22,3)(23,9)(24,12)(25,64)(26,15)(27,22)(28,9)(29,51)(30,26)(31,33)(32,162)(33,52)(34,12)(35,41)(36,123)(37,27)(38,98)(39,317)(40,12)(41,262)(42,45)(43,21)(44,67)(45,64)(46,22)(47,501)(48,20)(49,42)(50,789)(51,38)(52,409)(53,43)(54,58)(55,39)(56,69)(57,56)(58,257)(59,874)(60,2295)(61,98)(62,493)(63,220)(64,1431)(65,385)(66,150)(67,683)(68,436)(69,1746)(70,35)(71,62)(72,165)(73,104)(74,2713)(75,2475)(76,134)(77,5925)(78,223)(79,359)(80,85)(81,229)(82,195)(83,221)(84,113)(85,372)(86,181)(87,1419)(88,41)(89,1154)(90,3965)(91,1057)(92,710)(93,4290)(94,853)(95,392)(96,589)(97,750)(98,865)(99,7341)(100,0)
		} 
		\closedcycle;
		\nextgroupplot[ymin=0,ymax=6000,xlabel style={align=center},xlabel = Hardness Degree,ytick={0,2000,4000,6000},yticklabels={0,2K,4K,6K},scaled y ticks = false]
		\addplot
		[ybar interval,mark=no,fill=purple] 
		coordinates
		{
			(0,7)(1,15)(2,116)(3,46)(4,10)(5,61)(6,12)(7,61)(8,10)(9,107)(10,34)(11,23)(12,23)(13,46)(14,165)(15,33)(16,67)(17,30)(18,20)(19,43)(20,40)(21,115)(22,33)(23,254)(24,70)(25,46)(26,54)(27,84)(28,44)(29,55)(30,130)(31,343)(32,108)(33,102)(34,80)(35,194)(36,172)(37,136)(38,376)(39,400)(40,439)(41,223)(42,241)(43,353)(44,545)(45,354)(46,468)(47,545)(48,492)(49,517)(50,453)(51,509)(52,537)(53,718)(54,699)(55,742)(56,804)(57,1003)(58,890)(59,498)(60,924)(61,980)(62,712)(63,1011)(64,605)(65,821)(66,547)(67,640)(68,684)(69,525)(70,394)(71,672)(72,485)(73,599)(74,510)(75,597)(76,682)(77,439)(78,683)(79,688)(80,471)(81,573)(82,538)(83,598)(84,390)(85,535)(86,631)(87,628)(88,500)(89,720)(90,689)(91,808)(92,793)(93,1009)(94,824)(95,1382)(96,1252)(97,2241)(98,2459)(99,4966)(100,0)
		};
		\nextgroupplot[ymin=0,ymax=6000,xlabel = Hardness Degree,ytick={0,2000,4000,6000},yticklabels={0,2K,4K,6K},scaled y ticks = false]
		\addplot
		[ybar interval,mark=no,fill=purple] 
		coordinates
		{
(0,52)(1,55)(2,123)(3,56)(4,91)(5,51)(6,39)(7,67)(8,55)(9,32)(10,58)(11,93)(12,44)(13,26)(14,63)(15,40)(16,49)(17,54)(18,63)(19,65)(20,41)(21,50)(22,44)(23,96)(24,84)(25,84)(26,80)(27,80)(28,58)(29,116)(30,224)(31,195)(32,105)(33,128)(34,151)(35,122)(36,164)(37,202)(38,190)(39,292)(40,339)(41,237)(42,271)(43,329)(44,421)(45,315)(46,376)(47,489)(48,420)(49,416)(50,424)(51,412)(52,694)(53,589)(54,709)(55,611)(56,825)(57,856)(58,762)(59,558)(60,837)(61,817)(62,798)(63,870)(64,686)(65,706)(66,575)(67,688)(68,755)(69,520)(70,456)(71,546)(72,694)(73,604)(74,515)(75,622)(76,596)(77,465)(78,655)(79,629)(80,484)(81,627)(82,616)(83,554)(84,434)(85,692)(86,621)(87,759)(88,513)(89,770)(90,664)(91,876)(92,866)(93,935)(94,868)(95,1409)(96,1249)(97,1986)(98,2755)(99,5553)(100,0)
		};
		\end{groupplot}
		\end{tikzpicture}
	}
		\vspace{-0.6cm}
		\caption{ CIFAR100}
	\end{subfigure}
		\vspace{-0.6\baselineskip}
	\caption{The hardness degree histograms of samples of four various model extraction attacks on CIFAR10 and CIFAR100 target classifiers. The budget of model extraction attacks is 50000. The architecture of target classifiers is DenseNet121.}
	\label{fig:hardhistmeadensnet}
\end{figure}
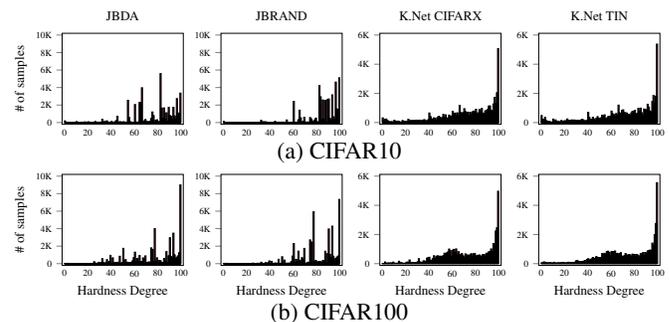

Figure \ref{fig:hardhistmeadensnet} shows the hardness degree histograms of samples of four model extraction attacks for CIFAR10 and CIFAR100 target classifiers. The architecture of target classifiers is DenseNet121. As seen in Figure \ref{fig:hardhistmeadensnet}, the hardness degree histogram of model extraction attack samples is distinguishable from the hardness degree of normal samples (Figure \ref{loginMock}) for DenseNet121 target classifiers.

%% file: content/viscifar100.tex
\begin{figure}[!t]
	\begin{subfigure}{0.32\linewidth}
		\resizebox{\linewidth}{!}{
			\begin{tikzpicture}
			\node (myfirstpic) at (0,0) {\includegraphics[scale=4]{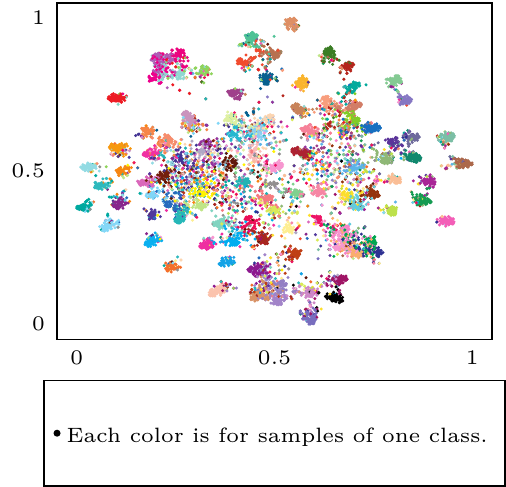}};
			\end{tikzpicture}
		}
		\caption{}
		\label{fig:viscifar100a}
	\end{subfigure}
	\begin{subfigure}{0.32\linewidth}
		\resizebox{\linewidth}{!}{
			\begin{tikzpicture}
			\node (myfirstpic) at (0,0) {\includegraphics[scale=2]{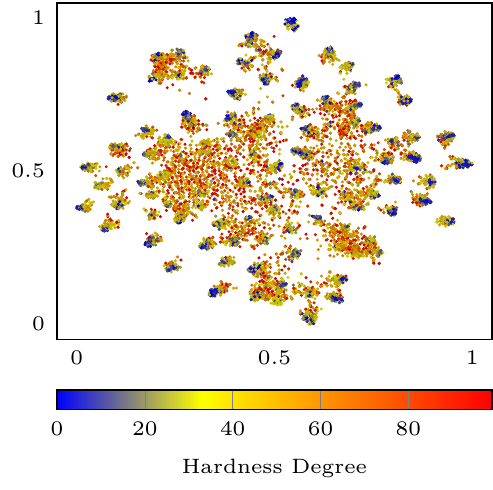}};
			\end{tikzpicture}
		}
		\caption{}
		\label{fig:viscifar100b}
	\end{subfigure}
	\begin{subfigure}{0.32\linewidth}
		\resizebox{\linewidth}{!}{
			\begin{tikzpicture}
			\node (myfirstpic) at (0,0) {\includegraphics[scale=2]{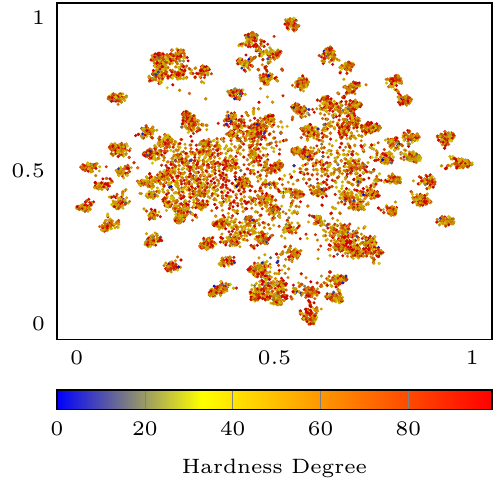}};
			\end{tikzpicture}
		}
		\caption{}
		\label{fig:viscifar100c}
	\end{subfigure}
	\caption{ (a) Visualization of CIFAR100 test samples. (b) Hardness of CIFAR100 test samples for CIFAR100 classifier. (c) Hardness of CIFAR100 test samples for CIFAR10 classifier.}
	\label{fig:viscifar100}
\end{figure}

Figure \ref{fig:viscifar100} displays a two-dimensional visualization of CIFAR100 test samples using t-SNE. Figure \ref{fig:viscifar100a} uses the logits of CIFAR100 classifier to visualize CIFAR100 test samples, and the color of each sample is determined by its label. Figures \ref{fig:viscifar100b}  and \ref{fig:viscifar100c} show the hardness degree of CIFAR100 test samples for CIFAR100  and CIFAR10 target classifiers, respectively.

%% file: content/hodaalgorithm.tex
\label{sec:hodaalg}
\input{content/AlgHD.tex}
\input{content/AlgHODA.tex}

%% file: content/AlgHD.tex
\begin{algorithm}
			\scriptsize
\caption{ Hardness Degree Computation}
\label{alg:hd}
\textbf{Inputs}: $x$ is a sample and $F_{subclf}$ is a sequence of subclassifiers\\
\textbf{Outputs}: $degree$ is the hardness degree of sample $x$
	\begin{algorithmic}
\Function{GetHardnessDegree}{($x,F_{subclf}$)}
\State $label \gets \text{None}$
\For{$i\gets 0, len(F_{subclf})$}	
\State $pred\_vector \gets F_{subclf}[i](x)$ $\quad//$ $F_{subclf}[i]$ is the $i^{th}$ subclassifier 
\State $\quad \quad \quad \quad \quad \quad \quad \quad \quad \quad \quad \quad \quad \quad \quad \quad \;\;$ in sequence $F_{subclf}$
\State $pred\_label \gets \text{argmax}(pred\_vector)$
\If{$pred\_label \neq label$}
\State $degree = i$
\State $label = pred\_label$
\EndIf
\EndFor
	\State \textbf{return} $degree$
\EndFunction
	\end{algorithmic}
\end{algorithm}

%% file: content/AlgHODA.tex
		\begin{algorithm}
			\scriptsize
			\caption{ Hardness-Oriented Detection Approach  (HODA)}
			\label{alg:AWA}
			 \textbf{Inputs}: $S_{HODA}$ is a set of normal samples, $num_s$ is the size of sample sequences, $num_{seq}$ is the number of sample sequences,
			  $F_{subclf}$ is the subclassifier sequence, $NewQuery$ is the newest query being received by the target model, and 
			  $UserID$ is the owner identifier of $NewQuery$.\\
			\textbf{Outputs}: $H_n$ is the histogram of normal samples, $\delta$ is the attack detection threshold, $AttackAlarm$ declares the occurrence of attack.
\begin{algorithmic}[1]
	\Function{PearsonDist}{($H_{n},H_{u}$)}
	\State \textbf{return} $PD(H_n/Sum(H_n),H_{u}/Sum(H_u))$
	\EndFunction
	\Function{HodaInitialization}{($S_{HODA},num_s,num_{seq},F_{subclf}$)}
	\State $ HistSet \gets \emptyset $ 
	\For{$i\gets 0, num_{seq}$}	
	\State $seq \gets $ Randomly select $num_s$ samples from $S_{HODA}$
	\State $ Hist\gets \emptyset $ 
	\For{$s \;\textbf{in} \; seq$}
	\State $HD =$ GetHardnessDegree($s,F_{subclf}$)
	\State $Hist[HD] \; += 1$ 
	\EndFor
	\State $ HistSet \gets HistSet\; \cup\; Hist $
	\EndFor
	\State $H_{n} \gets Avg(HistSet)$
	\State $DistList \gets \emptyset $
	\For{$Hist \;\textbf{in} \; HistSet$} 
	\State $DistList.\text{append}($ \textsc{PearsonDist}$(H_{n},Hist)) $ 
	\EndFor 
	\State $\delta \gets Max(DistList)$ 
	\State \textbf{return} $H_{n} , \delta$
	\EndFunction
	\Function{HODA}{($NewQuery,UserID, H_n,\delta,num_s,F_{subclf}$)}	
	\State $AttackAlarm \gets False$
	\State $H_u \gets$ GetUserHisogram($UserID$)  
	\State $HD \gets$ GetHardnessDegree($NewQuery,F_{subclf}$)  
	\State $H_u[HD] \;+= 1$
	\If{$Sum(H_u) == Num_s$}
	\If{\textsc{PearsonDist}$(H_n,H_u) > \delta$ }
	\State $AttackAlarm \gets True$
	\EndIf 
	\EndIf
	\State SaveUserHistogram($H_u,UserID$)
	\State \textbf{return} $AttackAlarm$
	\EndFunction
\end{algorithmic}
			\label{alg:hoda}
		\end{algorithm}

%% file: content/disthist.tex
\input{content/hodapdhist.tex}

HODA creates a hardness degree histogram for each user, called $H_u$, and calculates Pearson distance between $H_u$ and normal histogram $H_n$ (PD$(H_n, H_u)$). In section \ref{sec:seteva}, we simulated 10000 benign users and 10000 adversaries for each attack. 
Figure \ref{fig:disthist} indicates the histogram of Pearson distance between $H_n$ and benign users' hardness degree histogram and also $H_n$ and adversaries' hardness degree histogram for $num_s=100$ and HODA-11. Notably, Pearson distance is in the range [0,2]. 

We also consider 10000 STL10 users. STL10 users randomly select their samples from STL10 dataset (details in Appendix \ref{sec:appa}). STL10 dataset has been inspired by the CIFAR10 dataset, and its images are obtained from the ImageNet dataset. It has the same classes as the CIFAR10 dataset, except instead of the frog class, it has a monkey class. We remove monkey images from STL10 dataset. Since classes of modified STL10 dataset are a subset of CIFAR10 classes, STL10 users use in-distribution samples for the CIFAR10 target classifier. However, similar to K.Net CIFARX attack on CIFAR100 target classifier, STL10 users use out-of-distribution samples for CIFAR100 target classifier. Since we suppose that only adversaries use out-of-distribution samples to extract a target model, STL10 users are adversaries for CIFAR100 target classifier and are benign users for CIFAR10 target classifier. The detection rate of HODA for STL10 users is 100\% for CIFAR100 target classifier and 5.27\% for CIFAR10 target classifier in $num_s=100$. Hence, the false-positive rate of HODA for STL10 users, which are considered benign users, is 5.27\% for CIFAR10 target classifier. By choosing a larger delta value (e.g., $\delta=0.25$), the false-positive rate of HODA for STL10 users can be reduced to almost zero with almost no change in the detection rate of attacks.

%% file: content/hodapdhist.tex
\begin{figure*}
	\label{fig:hardhist}
	\begin{subfigure}{\linewidth}
		\centering
		\resizebox{\linewidth}{!}{
		\begin{tikzpicture}
\begin{groupplot}[group style={group size= 5 by 2,vertical sep=0.9cm,horizontal sep=0.3cm},height=4cm,width=4.8cm,tick label style = {font = {\fontsize{8 pt}{12 pt}\selectfont}},,ytick align=outside,xtick style={},xtick align=outside,tick label style={font=\tiny},xtick pos=bottom,ytick pos=left,enlargelimits=true,enlargelimits=0.02,,title style={yshift=-0.2cm},xtick={0,0.25,0.5,0.75,1,1.25,1.5,1.75,2},xmax=2,x tick label style={rotate=45,anchor=east},]
\nextgroupplot[ymin=0, ymax=10000,title=JBDA,ylabel=\# of users,ytick={0,2000,4000,6000,8000,10000},yticklabels={0,2K,4K,6K,8K,10K},scaled y ticks = false]
\addplot
[ybar interval,mark=no,fill=purple] 
coordinates
{
( 0.0 , 9262.0 ) ( 0.05 , 681.0 ) ( 0.1 , 53.0 ) ( 0.15 , 4.0 ) ( 0.2 , 0.0 ) ( 0.25 , 0.0 ) ( 0.3 , 0.0 ) ( 0.35 , 0.0 ) ( 0.4 , 0.0 ) ( 0.45 , 0.0 ) ( 0.5 , 0.0 ) ( 0.55 , 0.0 ) ( 0.6 , 0.0 ) ( 0.65 , 0.0 ) ( 0.7 , 0.0 ) ( 0.75 , 0.0 ) ( 0.8 , 0.0 ) ( 0.85 , 0.0 ) ( 0.9 , 0.0 ) ( 0.95 , 0.0 ) ( 1.0 , 0.0 ) ( 1.05 , 0.0 ) ( 1.1 , 0.0 ) ( 1.15 , 0.0 ) ( 1.2 , 0.0 ) ( 1.25 , 0.0 ) ( 1.3 , 0.0 ) ( 1.35 , 0.0 ) ( 1.4 , 0.0 ) ( 1.45 , 0.0 ) ( 1.5 , 0.0 ) ( 1.55 , 0.0 ) ( 1.6 , 0.0 ) ( 1.65 , 0.0 ) ( 1.7 , 0.0 ) ( 1.75 , 0.0 ) ( 1.8 , 0.0 ) ( 1.85 , 0.0 ) ( 1.9 , 0.0 ) ( 1.95 , 0.0 ) 

} 
\closedcycle;
\addplot
[ybar interval,mark=no,fill=blue] 
coordinates
{
( 0.0 , 0.0 ) ( 0.05 , 0.0 ) ( 0.1 , 0.0 ) ( 0.15 , 0.0 ) ( 0.2 , 0.0 ) ( 0.25 , 0.0 ) ( 0.3 , 0.0 ) ( 0.35 , 0.0 ) ( 0.4 , 0.0 ) ( 0.45 , 0.0 ) ( 0.5 , 0.0 ) ( 0.55 , 0.0 ) ( 0.6 , 0.0 ) ( 0.65 , 0.0 ) ( 0.7 , 0.0 ) ( 0.75 , 0.0 ) ( 0.8 , 0.0 ) ( 0.85 , 0.0 ) ( 0.9 , 2.0 ) ( 0.95 , 44.0 ) ( 1.0 , 195.0 ) ( 1.05 , 722.0 ) ( 1.1 , 1873.0 ) ( 1.15 , 3041.0 ) ( 1.2 , 2589.0 ) ( 1.25 , 1193.0 ) ( 1.3 , 289.0 ) ( 1.35 , 48.0 ) ( 1.4 , 3.0 ) ( 1.45 , 1.0 ) ( 1.5 , 0.0 ) ( 1.55 , 0.0 ) ( 1.6 , 0.0 ) ( 1.65 , 0.0 ) ( 1.7 , 0.0 ) ( 1.75 , 0.0 ) ( 1.8 , 0.0 ) ( 1.85 , 0.0 ) ( 1.9 , 0.0 ) ( 1.95 , 0.0 )

} 
\closedcycle;
\nextgroupplot[ymin=0, ymax=10000,title=JBRAND,ytick={},yticklabels={},scaled y ticks = false]
\addplot
[ybar interval,mark=no,fill=purple] 
coordinates
{
( 0.0 , 9262.0 ) ( 0.05 , 681.0 ) ( 0.1 , 53.0 ) ( 0.15 , 4.0 ) ( 0.2 , 0.0 ) ( 0.25 , 0.0 ) ( 0.3 , 0.0 ) ( 0.35 , 0.0 ) ( 0.4 , 0.0 ) ( 0.45 , 0.0 ) ( 0.5 , 0.0 ) ( 0.55 , 0.0 ) ( 0.6 , 0.0 ) ( 0.65 , 0.0 ) ( 0.7 , 0.0 ) ( 0.75 , 0.0 ) ( 0.8 , 0.0 ) ( 0.85 , 0.0 ) ( 0.9 , 0.0 ) ( 0.95 , 0.0 ) ( 1.0 , 0.0 ) ( 1.05 , 0.0 ) ( 1.1 , 0.0 ) ( 1.15 , 0.0 ) ( 1.2 , 0.0 ) ( 1.25 , 0.0 ) ( 1.3 , 0.0 ) ( 1.35 , 0.0 ) ( 1.4 , 0.0 ) ( 1.45 , 0.0 ) ( 1.5 , 0.0 ) ( 1.55 , 0.0 ) ( 1.6 , 0.0 ) ( 1.65 , 0.0 ) ( 1.7 , 0.0 ) ( 1.75 , 0.0 ) ( 1.8 , 0.0 ) ( 1.85 , 0.0 ) ( 1.9 , 0.0 ) ( 1.95 , 0.0 )

} 
\closedcycle;
\addplot
[ybar interval,mark=no,fill=blue] 
coordinates
{
( 0.0 , 0.0 ) ( 0.05 , 0.0 ) ( 0.1 , 0.0 ) ( 0.15 , 0.0 ) ( 0.2 , 0.0 ) ( 0.25 , 0.0 ) ( 0.3 , 0.0 ) ( 0.35 , 0.0 ) ( 0.4 , 0.0 ) ( 0.45 , 0.0 ) ( 0.5 , 0.0 ) ( 0.55 , 0.0 ) ( 0.6 , 0.0 ) ( 0.65 , 0.0 ) ( 0.7 , 0.0 ) ( 0.75 , 0.0 ) ( 0.8 , 0.0 ) ( 0.85 , 0.0 ) ( 0.9 , 0.0 ) ( 0.95 , 0.0 ) ( 1.0 , 0.0 ) ( 1.05 , 11.0 ) ( 1.1 , 198.0 ) ( 1.15 , 1730.0 ) ( 1.2 , 4563.0 ) ( 1.25 , 2940.0 ) ( 1.3 , 529.0 ) ( 1.35 , 29.0 ) ( 1.4 , 0.0 ) ( 1.45 , 0.0 ) ( 1.5 , 0.0 ) ( 1.55 , 0.0 ) ( 1.6 , 0.0 ) ( 1.65 , 0.0 ) ( 1.7 , 0.0 ) ( 1.75 , 0.0 ) ( 1.8 , 0.0 ) ( 1.85 , 0.0 ) ( 1.9 , 0.0 ) ( 1.95 , 0.0 )

} 
\closedcycle;
\nextgroupplot[ymin=0, ymax=10000,title=K.Net CIFARX,ytick={},yticklabels={},scaled y ticks = false]
\addplot
[ybar interval,mark=no,fill=purple] 
coordinates
{
( 0.0 , 9262.0 ) ( 0.05 , 681.0 ) ( 0.1 , 53.0 ) ( 0.15 , 4.0 ) ( 0.2 , 0.0 ) ( 0.25 , 0.0 ) ( 0.3 , 0.0 ) ( 0.35 , 0.0 ) ( 0.4 , 0.0 ) ( 0.45 , 0.0 ) ( 0.5 , 0.0 ) ( 0.55 , 0.0 ) ( 0.6 , 0.0 ) ( 0.65 , 0.0 ) ( 0.7 , 0.0 ) ( 0.75 , 0.0 ) ( 0.8 , 0.0 ) ( 0.85 , 0.0 ) ( 0.9 , 0.0 ) ( 0.95 , 0.0 ) ( 1.0 , 0.0 ) ( 1.05 , 0.0 ) ( 1.1 , 0.0 ) ( 1.15 , 0.0 ) ( 1.2 , 0.0 ) ( 1.25 , 0.0 ) ( 1.3 , 0.0 ) ( 1.35 , 0.0 ) ( 1.4 , 0.0 ) ( 1.45 , 0.0 ) ( 1.5 , 0.0 ) ( 1.55 , 0.0 ) ( 1.6 , 0.0 ) ( 1.65 , 0.0 ) ( 1.7 , 0.0 ) ( 1.75 , 0.0 ) ( 1.8 , 0.0 ) ( 1.85 , 0.0 ) ( 1.9 , 0.0 ) ( 1.95 , 0.0 )

} 
\closedcycle;
\addplot
[ybar interval,mark=no,fill=blue] 
coordinates
{
( 0.0 , 0.0 ) ( 0.05 , 0.0 ) ( 0.1 , 0.0 ) ( 0.15 , 1.0 ) ( 0.2 , 0.0 ) ( 0.25 , 1.0 ) ( 0.3 , 4.0 ) ( 0.35 , 15.0 ) ( 0.4 , 17.0 ) ( 0.45 , 28.0 ) ( 0.5 , 41.0 ) ( 0.55 , 90.0 ) ( 0.6 , 107.0 ) ( 0.65 , 161.0 ) ( 0.7 , 228.0 ) ( 0.75 , 293.0 ) ( 0.8 , 370.0 ) ( 0.85 , 425.0 ) ( 0.9 , 560.0 ) ( 0.95 , 652.0 ) ( 1.0 , 715.0 ) ( 1.05 , 757.0 ) ( 1.1 , 887.0 ) ( 1.15 , 855.0 ) ( 1.2 , 736.0 ) ( 1.25 , 762.0 ) ( 1.3 , 601.0 ) ( 1.35 , 521.0 ) ( 1.4 , 418.0 ) ( 1.45 , 318.0 ) ( 1.5 , 184.0 ) ( 1.55 , 112.0 ) ( 1.6 , 83.0 ) ( 1.65 , 34.0 ) ( 1.7 , 17.0 ) ( 1.75 , 6.0 ) ( 1.8 , 1.0 ) ( 1.85 , 0.0 ) ( 1.9 , 0.0 ) ( 1.95 , 0.0 )

} 
\closedcycle;
\nextgroupplot[ymin=0,ymax=10000,title=K.Net TIN,,ytick={},yticklabels={},scaled y ticks = false]
\addplot
[ybar interval,mark=no,fill=purple] 
coordinates
{
( 0.0 , 9262.0 ) ( 0.05 , 681.0 ) ( 0.1 , 53.0 ) ( 0.15 , 4.0 ) ( 0.2 , 0.0 ) ( 0.25 , 0.0 ) ( 0.3 , 0.0 ) ( 0.35 , 0.0 ) ( 0.4 , 0.0 ) ( 0.45 , 0.0 ) ( 0.5 , 0.0 ) ( 0.55 , 0.0 ) ( 0.6 , 0.0 ) ( 0.65 , 0.0 ) ( 0.7 , 0.0 ) ( 0.75 , 0.0 ) ( 0.8 , 0.0 ) ( 0.85 , 0.0 ) ( 0.9 , 0.0 ) ( 0.95 , 0.0 ) ( 1.0 , 0.0 ) ( 1.05 , 0.0 ) ( 1.1 , 0.0 ) ( 1.15 , 0.0 ) ( 1.2 , 0.0 ) ( 1.25 , 0.0 ) ( 1.3 , 0.0 ) ( 1.35 , 0.0 ) ( 1.4 , 0.0 ) ( 1.45 , 0.0 ) ( 1.5 , 0.0 ) ( 1.55 , 0.0 ) ( 1.6 , 0.0 ) ( 1.65 , 0.0 ) ( 1.7 , 0.0 ) ( 1.75 , 0.0 ) ( 1.8 , 0.0 ) ( 1.85 , 0.0 ) ( 1.9 , 0.0 ) ( 1.95 , 0.0 )

} 
\closedcycle;
\addplot
[ybar interval,mark=no,fill=blue] 
coordinates
{
( 0.0 , 0.0 ) ( 0.05 , 0.0 ) ( 0.1 , 0.0 ) ( 0.15 , 0.0 ) ( 0.2 , 0.0 ) ( 0.25 , 5.0 ) ( 0.3 , 7.0 ) ( 0.35 , 10.0 ) ( 0.4 , 32.0 ) ( 0.45 , 44.0 ) ( 0.5 , 63.0 ) ( 0.55 , 118.0 ) ( 0.6 , 151.0 ) ( 0.65 , 203.0 ) ( 0.7 , 289.0 ) ( 0.75 , 371.0 ) ( 0.8 , 438.0 ) ( 0.85 , 502.0 ) ( 0.9 , 607.0 ) ( 0.95 , 669.0 ) ( 1.0 , 718.0 ) ( 1.05 , 755.0 ) ( 1.1 , 782.0 ) ( 1.15 , 784.0 ) ( 1.2 , 692.0 ) ( 1.25 , 674.0 ) ( 1.3 , 586.0 ) ( 1.35 , 464.0 ) ( 1.4 , 371.0 ) ( 1.45 , 268.0 ) ( 1.5 , 161.0 ) ( 1.55 , 115.0 ) ( 1.6 , 61.0 ) ( 1.65 , 31.0 ) ( 1.7 , 22.0 ) ( 1.75 , 6.0 ) ( 1.8 , 1.0 ) ( 1.85 , 0.0 ) ( 1.9 , 0.0 ) ( 1.95 , 0.0 )

} 
\closedcycle;
\nextgroupplot[ymin=0,ymax=10000,title=STL10,,ytick={},yticklabels={},scaled y ticks = false]
\addplot
[ybar interval,mark=no,fill=purple] 
coordinates
{
( 0.0 , 9262.0 ) ( 0.05 , 681.0 ) ( 0.1 , 53.0 ) ( 0.15 , 4.0 ) ( 0.2 , 0.0 ) ( 0.25 , 0.0 ) ( 0.3 , 0.0 ) ( 0.35 , 0.0 ) ( 0.4 , 0.0 ) ( 0.45 , 0.0 ) ( 0.5 , 0.0 ) ( 0.55 , 0.0 ) ( 0.6 , 0.0 ) ( 0.65 , 0.0 ) ( 0.7 , 0.0 ) ( 0.75 , 0.0 ) ( 0.8 , 0.0 ) ( 0.85 , 0.0 ) ( 0.9 , 0.0 ) ( 0.95 , 0.0 ) ( 1.0 , 0.0 ) ( 1.05 , 0.0 ) ( 1.1 , 0.0 ) ( 1.15 , 0.0 ) ( 1.2 , 0.0 ) ( 1.25 , 0.0 ) ( 1.3 , 0.0 ) ( 1.35 , 0.0 ) ( 1.4 , 0.0 ) ( 1.45 , 0.0 ) ( 1.5 , 0.0 ) ( 1.55 , 0.0 ) ( 1.6 , 0.0 ) ( 1.65 , 0.0 ) ( 1.7 , 0.0 ) ( 1.75 , 0.0 ) ( 1.8 , 0.0 ) ( 1.85 , 0.0 ) ( 1.9 , 0.0 ) ( 1.95 , 0.0 )

} 
\closedcycle;
\addplot
[ybar interval,mark=no,fill=orange] 
coordinates
{
( 0.0 , 3226.0 ) ( 0.05 , 4406.0 ) ( 0.1 , 1768.0 ) ( 0.15 , 449.0 ) ( 0.2 , 123.0 ) ( 0.25 , 20.0 ) ( 0.3 , 6.0 ) ( 0.35 , 2.0 ) ( 0.4 , 0.0 ) ( 0.45 , 0.0 ) ( 0.5 , 0.0 ) ( 0.55 , 0.0 ) ( 0.6 , 0.0 ) ( 0.65 , 0.0 ) ( 0.7 , 0.0 ) ( 0.75 , 0.0 ) ( 0.8 , 0.0 ) ( 0.85 , 0.0 ) ( 0.9 , 0.0 ) ( 0.95 , 0.0 ) ( 1.0 , 0.0 ) ( 1.05 , 0.0 ) ( 1.1 , 0.0 ) ( 1.15 , 0.0 ) ( 1.2 , 0.0 ) ( 1.25 , 0.0 ) ( 1.3 , 0.0 ) ( 1.35 , 0.0 ) ( 1.4 , 0.0 ) ( 1.45 , 0.0 ) ( 1.5 , 0.0 ) ( 1.55 , 0.0 ) ( 1.6 , 0.0 ) ( 1.65 , 0.0 ) ( 1.7 , 0.0 ) ( 1.75 , 0.0 ) ( 1.8 , 0.0 ) ( 1.85 , 0.0 ) ( 1.9 , 0.0 ) ( 1.95 , 0.0 ) 
	
} 
\closedcycle;
	\end{groupplot}
\end{tikzpicture}
}
\vspace{-0.5cm}
\caption{ \small{CIFAR10}}
\end{subfigure}
\begin{subfigure}{\linewidth}
\centering
\resizebox{\linewidth}{!}{
\begin{tikzpicture}
\begin{groupplot}[group style={group size= 5 by 2,vertical sep=0.9cm,horizontal sep=0.3cm},height=4cm,width=4.8cm,tick label style = {font = {\fontsize{8 pt}{12 pt}\selectfont}},,ytick align=outside,xtick style={},xtick align=outside,tick label style={font=\tiny},xtick pos=bottom,ytick pos=left,enlargelimits=true,enlargelimits=0.02,,title style={yshift=-0.2cm},xtick={0,0.25,0.5,0.75,1,1.25,1.5,1.75,2},xmax=2,x tick label style={rotate=45,anchor=east}]
\nextgroupplot[ymin=0, ymax=10000,xlabel= PD${(H_n,H_u)}$,ylabel=\# of users,ytick={0,2000,4000,6000,8000,10000},yticklabels={0,2K,4K,6K,8K,10K},scaled y ticks = false]

\addplot
[ybar interval,mark=no,fill=purple] 
coordinates
{
( 0.0 , 6160.0 ) ( 0.05 , 2991.0 ) ( 0.1 , 626.0 ) ( 0.15 , 156.0 ) ( 0.2 , 47.0 ) ( 0.25 , 14.0 ) ( 0.3 , 4.0 ) ( 0.35 , 2.0 ) ( 0.4 , 0.0 ) ( 0.45 , 0.0 ) ( 0.5 , 0.0 ) ( 0.55 , 0.0 ) ( 0.6 , 0.0 ) ( 0.65 , 0.0 ) ( 0.7 , 0.0 ) ( 0.75 , 0.0 ) ( 0.8 , 0.0 ) ( 0.85 , 0.0 ) ( 0.9 , 0.0 ) ( 0.95 , 0.0 ) ( 1.0 , 0.0 ) ( 1.05 , 0.0 ) ( 1.1 , 0.0 ) ( 1.15 , 0.0 ) ( 1.2 , 0.0 ) ( 1.25 , 0.0 ) ( 1.3 , 0.0 ) ( 1.35 , 0.0 ) ( 1.4 , 0.0 ) ( 1.45 , 0.0 ) ( 1.5 , 0.0 ) ( 1.55 , 0.0 ) ( 1.6 , 0.0 ) ( 1.65 , 0.0 ) ( 1.7 , 0.0 ) ( 1.75 , 0.0 ) ( 1.8 , 0.0 ) ( 1.85 , 0.0 ) ( 1.9 , 0.0 ) ( 1.95 , 0.0 ) 

} 
\closedcycle;
\addplot
[ybar interval,mark=no,fill=blue] 
coordinates
{
( 0.0 , 0.0 ) ( 0.05 , 0.0 ) ( 0.1 , 0.0 ) ( 0.15 , 0.0 ) ( 0.2 , 0.0 ) ( 0.25 , 0.0 ) ( 0.3 , 0.0 ) ( 0.35 , 0.0 ) ( 0.4 , 0.0 ) ( 0.45 , 2.0 ) ( 0.5 , 3.0 ) ( 0.55 , 12.0 ) ( 0.6 , 31.0 ) ( 0.65 , 70.0 ) ( 0.7 , 208.0 ) ( 0.75 , 338.0 ) ( 0.8 , 644.0 ) ( 0.85 , 1062.0 ) ( 0.9 , 1471.0 ) ( 0.95 , 1770.0 ) ( 1.0 , 1720.0 ) ( 1.05 , 1283.0 ) ( 1.1 , 818.0 ) ( 1.15 , 386.0 ) ( 1.2 , 139.0 ) ( 1.25 , 33.0 ) ( 1.3 , 10.0 ) ( 1.35 , 0.0 ) ( 1.4 , 0.0 ) ( 1.45 , 0.0 ) ( 1.5 , 0.0 ) ( 1.55 , 0.0 ) ( 1.6 , 0.0 ) ( 1.65 , 0.0 ) ( 1.7 , 0.0 ) ( 1.75 , 0.0 ) ( 1.8 , 0.0 ) ( 1.85 , 0.0 ) ( 1.9 , 0.0 ) ( 1.95 , 0.0 )

} 
\closedcycle;
\nextgroupplot[ymin=0, ymax=10000,xlabel= PD${(H_n,H_u)}$,ytick={},yticklabels={},scaled y ticks = false]
\addplot
[ybar interval,mark=no,fill=purple] 
coordinates
{
( 0.0 , 6160.0 ) ( 0.05 , 2991.0 ) ( 0.1 , 626.0 ) ( 0.15 , 156.0 ) ( 0.2 , 47.0 ) ( 0.25 , 14.0 ) ( 0.3 , 4.0 ) ( 0.35 , 2.0 ) ( 0.4 , 0.0 ) ( 0.45 , 0.0 ) ( 0.5 , 0.0 ) ( 0.55 , 0.0 ) ( 0.6 , 0.0 ) ( 0.65 , 0.0 ) ( 0.7 , 0.0 ) ( 0.75 , 0.0 ) ( 0.8 , 0.0 ) ( 0.85 , 0.0 ) ( 0.9 , 0.0 ) ( 0.95 , 0.0 ) ( 1.0 , 0.0 ) ( 1.05 , 0.0 ) ( 1.1 , 0.0 ) ( 1.15 , 0.0 ) ( 1.2 , 0.0 ) ( 1.25 , 0.0 ) ( 1.3 , 0.0 ) ( 1.35 , 0.0 ) ( 1.4 , 0.0 ) ( 1.45 , 0.0 ) ( 1.5 , 0.0 ) ( 1.55 , 0.0 ) ( 1.6 , 0.0 ) ( 1.65 , 0.0 ) ( 1.7 , 0.0 ) ( 1.75 , 0.0 ) ( 1.8 , 0.0 ) ( 1.85 , 0.0 ) ( 1.9 , 0.0 ) ( 1.95 , 0.0 ) 

} 
\closedcycle;
\addplot
[ybar interval,mark=no,fill=blue] 
coordinates
{
( 0.0 , 0.0 ) ( 0.05 , 0.0 ) ( 0.1 , 0.0 ) ( 0.15 , 0.0 ) ( 0.2 , 0.0 ) ( 0.25 , 0.0 ) ( 0.3 , 0.0 ) ( 0.35 , 0.0 ) ( 0.4 , 0.0 ) ( 0.45 , 0.0 ) ( 0.5 , 0.0 ) ( 0.55 , 0.0 ) ( 0.6 , 0.0 ) ( 0.65 , 0.0 ) ( 0.7 , 0.0 ) ( 0.75 , 0.0 ) ( 0.8 , 0.0 ) ( 0.85 , 0.0 ) ( 0.9 , 0.0 ) ( 0.95 , 1.0 ) ( 1.0 , 13.0 ) ( 1.05 , 91.0 ) ( 1.1 , 556.0 ) ( 1.15 , 1965.0 ) ( 1.2 , 3567.0 ) ( 1.25 , 2705.0 ) ( 1.3 , 955.0 ) ( 1.35 , 136.0 ) ( 1.4 , 11.0 ) ( 1.45 , 0.0 ) ( 1.5 , 0.0 ) ( 1.55 , 0.0 ) ( 1.6 , 0.0 ) ( 1.65 , 0.0 ) ( 1.7 , 0.0 ) ( 1.75 , 0.0 ) ( 1.8 , 0.0 ) ( 1.85 , 0.0 ) ( 1.9 , 0.0 ) ( 1.95 , 0.0 )

} 
\closedcycle;
\nextgroupplot[ymin=0, ymax=10000,xlabel= PD${(H_n,H_u)}$,ytick={},yticklabels={},scaled y ticks = false,,legend style={at={(.5,-1.1)},
	anchor=south,legend columns=-1}]
\addlegendimage{red, only marks,mark = square*, mark size= 1.5pt}
\addlegendimage{blue, only marks, mark = square*, mark size= 1.5pt}
\addlegendimage{orange,only marks, mark = square*, mark size= 1.5pt}
\legend{  \footnotesize{Benign user}, \footnotesize{Adversary}, \footnotesize{STL10 User} }
\addplot
[ybar interval,mark=no,fill=purple] 
coordinates
{
( 0.0 , 6160.0 ) ( 0.05 , 2991.0 ) ( 0.1 , 626.0 ) ( 0.15 , 156.0 ) ( 0.2 , 47.0 ) ( 0.25 , 14.0 ) ( 0.3 , 4.0 ) ( 0.35 , 2.0 ) ( 0.4 , 0.0 ) ( 0.45 , 0.0 ) ( 0.5 , 0.0 ) ( 0.55 , 0.0 ) ( 0.6 , 0.0 ) ( 0.65 , 0.0 ) ( 0.7 , 0.0 ) ( 0.75 , 0.0 ) ( 0.8 , 0.0 ) ( 0.85 , 0.0 ) ( 0.9 , 0.0 ) ( 0.95 , 0.0 ) ( 1.0 , 0.0 ) ( 1.05 , 0.0 ) ( 1.1 , 0.0 ) ( 1.15 , 0.0 ) ( 1.2 , 0.0 ) ( 1.25 , 0.0 ) ( 1.3 , 0.0 ) ( 1.35 , 0.0 ) ( 1.4 , 0.0 ) ( 1.45 , 0.0 ) ( 1.5 , 0.0 ) ( 1.55 , 0.0 ) ( 1.6 , 0.0 ) ( 1.65 , 0.0 ) ( 1.7 , 0.0 ) ( 1.75 , 0.0 ) ( 1.8 , 0.0 ) ( 1.85 , 0.0 ) ( 1.9 , 0.0 ) ( 1.95 , 0.0 ) 

} 
\closedcycle;
\addplot
[ybar interval,mark=no,fill=blue] 
coordinates
{
( 0.0 , 0.0 ) ( 0.05 , 0.0 ) ( 0.1 , 0.0 ) ( 0.15 , 0.0 ) ( 0.2 , 0.0 ) ( 0.25 , 0.0 ) ( 0.3 , 0.0 ) ( 0.35 , 3.0 ) ( 0.4 , 10.0 ) ( 0.45 , 16.0 ) ( 0.5 , 34.0 ) ( 0.55 , 49.0 ) ( 0.6 , 112.0 ) ( 0.65 , 149.0 ) ( 0.7 , 269.0 ) ( 0.75 , 420.0 ) ( 0.8 , 519.0 ) ( 0.85 , 704.0 ) ( 0.9 , 937.0 ) ( 0.95 , 1149.0 ) ( 1.0 , 1175.0 ) ( 1.05 , 1159.0 ) ( 1.1 , 1003.0 ) ( 1.15 , 877.0 ) ( 1.2 , 588.0 ) ( 1.25 , 409.0 ) ( 1.3 , 239.0 ) ( 1.35 , 107.0 ) ( 1.4 , 44.0 ) ( 1.45 , 17.0 ) ( 1.5 , 9.0 ) ( 1.55 , 2.0 ) ( 1.6 , 0.0 ) ( 1.65 , 0.0 ) ( 1.7 , 0.0 ) ( 1.75 , 0.0 ) ( 1.8 , 0.0 ) ( 1.85 , 0.0 ) ( 1.9 , 0.0 ) ( 1.95 , 0.0 )

} 
\closedcycle;
\nextgroupplot[ymin=0,ymax=10000,xlabel= PD${(H_n,H_u)}$,,ytick={},yticklabels={},scaled y ticks = false]
\addplot
[ybar interval,mark=no,fill=purple] 
coordinates
{
( 0.0 , 6160.0 ) ( 0.05 , 2991.0 ) ( 0.1 , 626.0 ) ( 0.15 , 156.0 ) ( 0.2 , 47.0 ) ( 0.25 , 14.0 ) ( 0.3 , 4.0 ) ( 0.35 , 2.0 ) ( 0.4 , 0.0 ) ( 0.45 , 0.0 ) ( 0.5 , 0.0 ) ( 0.55 , 0.0 ) ( 0.6 , 0.0 ) ( 0.65 , 0.0 ) ( 0.7 , 0.0 ) ( 0.75 , 0.0 ) ( 0.8 , 0.0 ) ( 0.85 , 0.0 ) ( 0.9 , 0.0 ) ( 0.95 , 0.0 ) ( 1.0 , 0.0 ) ( 1.05 , 0.0 ) ( 1.1 , 0.0 ) ( 1.15 , 0.0 ) ( 1.2 , 0.0 ) ( 1.25 , 0.0 ) ( 1.3 , 0.0 ) ( 1.35 , 0.0 ) ( 1.4 , 0.0 ) ( 1.45 , 0.0 ) ( 1.5 , 0.0 ) ( 1.55 , 0.0 ) ( 1.6 , 0.0 ) ( 1.65 , 0.0 ) ( 1.7 , 0.0 ) ( 1.75 , 0.0 ) ( 1.8 , 0.0 ) ( 1.85 , 0.0 ) ( 1.9 , 0.0 ) ( 1.95 , 0.0 ) 

} 
\closedcycle;
\addplot
[ybar interval,mark=no,fill=blue] 
coordinates
{
( 0.0 , 0.0 ) ( 0.05 , 0.0 ) ( 0.1 , 0.0 ) ( 0.15 , 0.0 ) ( 0.2 , 0.0 ) ( 0.25 , 0.0 ) ( 0.3 , 0.0 ) ( 0.35 , 5.0 ) ( 0.4 , 12.0 ) ( 0.45 , 25.0 ) ( 0.5 , 49.0 ) ( 0.55 , 99.0 ) ( 0.6 , 168.0 ) ( 0.65 , 241.0 ) ( 0.7 , 361.0 ) ( 0.75 , 514.0 ) ( 0.8 , 654.0 ) ( 0.85 , 813.0 ) ( 0.9 , 960.0 ) ( 0.95 , 1072.0 ) ( 1.0 , 1115.0 ) ( 1.05 , 1066.0 ) ( 1.1 , 929.0 ) ( 1.15 , 736.0 ) ( 1.2 , 515.0 ) ( 1.25 , 338.0 ) ( 1.3 , 183.0 ) ( 1.35 , 100.0 ) ( 1.4 , 28.0 ) ( 1.45 , 13.0 ) ( 1.5 , 3.0 ) ( 1.55 , 1.0 ) ( 1.6 , 0.0 ) ( 1.65 , 0.0 ) ( 1.7 , 0.0 ) ( 1.75 , 0.0 ) ( 1.8 , 0.0 ) ( 1.85 , 0.0 ) ( 1.9 , 0.0 ) ( 1.95 , 0.0 ) 

} 
\closedcycle;
\nextgroupplot[ymin=0,ymax=10000,xlabel= PD${(H_n,H_u)}$,,ytick={},yticklabels={},scaled y ticks = false]
\addplot
[ybar interval,mark=no,fill=purple] 
coordinates
{
( 0.0 , 6160.0 ) ( 0.05 , 2991.0 ) ( 0.1 , 626.0 ) ( 0.15 , 156.0 ) ( 0.2 , 47.0 ) ( 0.25 , 14.0 ) ( 0.3 , 4.0 ) ( 0.35 , 2.0 ) ( 0.4 , 0.0 ) ( 0.45 , 0.0 ) ( 0.5 , 0.0 ) ( 0.55 , 0.0 ) ( 0.6 , 0.0 ) ( 0.65 , 0.0 ) ( 0.7 , 0.0 ) ( 0.75 , 0.0 ) ( 0.8 , 0.0 ) ( 0.85 , 0.0 ) ( 0.9 , 0.0 ) ( 0.95 , 0.0 ) ( 1.0 , 0.0 ) ( 1.05 , 0.0 ) ( 1.1 , 0.0 ) ( 1.15 , 0.0 ) ( 1.2 , 0.0 ) ( 1.25 , 0.0 ) ( 1.3 , 0.0 ) ( 1.35 , 0.0 ) ( 1.4 , 0.0 ) ( 1.45 , 0.0 ) ( 1.5 , 0.0 ) ( 1.55 , 0.0 ) ( 1.6 , 0.0 ) ( 1.65 , 0.0 ) ( 1.7 , 0.0 ) ( 1.75 , 0.0 ) ( 1.8 , 0.0 ) ( 1.85 , 0.0 ) ( 1.9 , 0.0 ) ( 1.95 , 0.0 ) 

} 
\closedcycle;
\addplot
[ybar interval,mark=no,fill=orange] 
coordinates
{
( 0.0 , 0.0 ) ( 0.05 , 0.0 ) ( 0.1 , 0.0 ) ( 0.15 , 0.0 ) ( 0.2 , 0.0 ) ( 0.25 , 0.0 ) ( 0.3 , 0.0 ) ( 0.35 , 6.0 ) ( 0.4 , 11.0 ) ( 0.45 , 23.0 ) ( 0.5 , 46.0 ) ( 0.55 , 68.0 ) ( 0.6 , 146.0 ) ( 0.65 , 225.0 ) ( 0.7 , 293.0 ) ( 0.75 , 468.0 ) ( 0.8 , 663.0 ) ( 0.85 , 775.0 ) ( 0.9 , 926.0 ) ( 0.95 , 1147.0 ) ( 1.0 , 1174.0 ) ( 1.05 , 1138.0 ) ( 1.1 , 973.0 ) ( 1.15 , 719.0 ) ( 1.2 , 531.0 ) ( 1.25 , 329.0 ) ( 1.3 , 182.0 ) ( 1.35 , 91.0 ) ( 1.4 , 39.0 ) ( 1.45 , 18.0 ) ( 1.5 , 6.0 ) ( 1.55 , 3.0 ) ( 1.6 , 0.0 ) ( 1.65 , 0.0 ) ( 1.7 , 0.0 ) ( 1.75 , 0.0 ) ( 1.8 , 0.0 ) ( 1.85 , 0.0 ) ( 1.9 , 0.0 ) ( 1.95 , 0.0 ) 

} 
\closedcycle;

\end{groupplot}
\end{tikzpicture}
}
\vspace{-1.5cm}
\caption{ \small{CIFAR100}}
\end{subfigure}
\vspace{-0.6\baselineskip}
\vspace{0.6cm}
\caption{The histogram of Pearson distance between $H_n$ and 10000 benign users' hardness degree histogram and $H_n$ and 10000 adversaries' hardness degree histogram for various attacks. STL10 users are benign users for CIFAR10 target classifier and are adversaries for CIFAR100 target classifier.}
\label{fig:disthist}
\end{figure*}
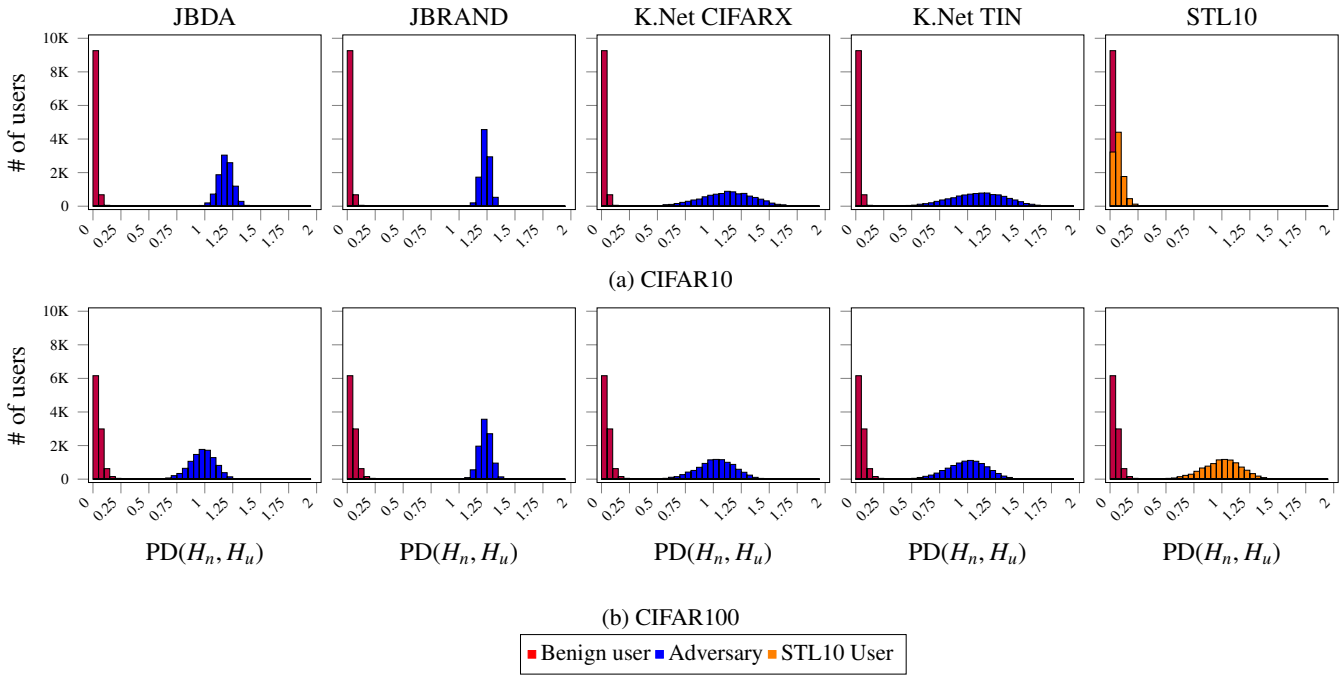

%% file: content/accdefsur.tex
\label{sec:defadv}
\begin{table}
	\caption{The average and standard deviation of the accuracy of surrogate classifiers for defended target models.}
\centering
	\label{tab:defsuracc}
	\renewcommand{\arraystretch}{1.7}
	\resizebox{\linewidth}{!}{
\begin{tabular}{ccccccc}
	\hline
	&         &                & \multicolumn{2}{c}{Acc of $f_s$ for undefended $f_t$ (\%)} & \multicolumn{2}{c}{Acc of $f_s$ for defended $f_t$ (\%)} \\ \cline{4-7} 
	$f_t$                     & Defense & $f_t$ Acc (\%) & K.Net CIFARX                   & K.Net TIN                     & K.Net CIFARX                  & K.Net TIN                    \\ \hline
	\multirow{2}{*}{CIFAR10}  & HODA    & 94.36          &      79.86               & 78.86                  & 17.1 $\pm$ 2.61                   & 17.77 $\pm$ 1.76                 \\ \cline{3-7} 
	& EDM     &      $94.83^*$           &       $85.39^*$                          &           -                    &    $68.5^*$                           &               -               \\ \hline
	\multirow{2}{*}{CIFAR100} & HODA    & 76.38          & 51.09                      & 60.36                     & 2.71 $\pm$ 0.38                     & 3.75 $\pm$ 0.52                    \\ \cline{3-7} 
	& EDM     &       $74.87^*$         &              $53.04^*$                  &              -                 &               $41.16^*$                &           -                   \\ \hline
	&         &                &            \multicolumn{2}{c}{$*$  are from \cite{kariyappa2021protecting}  }                                        &                               &                             
\end{tabular}
	}
\end{table}
\textcolor{\changecolor}{
Table \ref{tab:hbdmperfmea} shows that HODA-11 detects all simulated adversaries by only monitoring 100 samples ($num_s=100$) of each attack. Hence, adversaries only can use the prediction of at most 100 samples to train the surrogate classifier.
Table \ref{tab:defsuracc} compares the performance of HODA with EDM \cite{kariyappa2021protecting} and reports the accuracy of K.Net surrogate classifiers on CIFAR10 and CIFAR100 test sets for defended target models.
For HODA, since simulated adversaries randomly select their samples from 50000 samples of each attack, we simulate ten adversaries for each experiment and report the average and standard deviation of the accuracy of ten trained surrogate classifiers on 100 samples of each attack. The training process and architecture of surrogate classifiers are similar to target classifiers, and the output of target classifiers is a probability vector. For EDM, Table \ref{tab:defsuracc} presents the reported results in \cite{kariyappa2021protecting}. \citet{kariyappa2021protecting} proposed a defense with the same goal as perturbation-based defenses, which does not perturb the target classifier outputs. Their approach employs an ensemble of diverse models to produce discontinuous predictions for OOD samples.}

\textcolor{\changecolor}{
The advantage of Perturbation-based defenses is that they do not need to separate users' sample sequences, and their approach is independent of the sample owner. However, HODA needs to separate the samples of each user. Hence, HODA (and all previous detection-based methods) is vulnerable to Sybil attack. An adversary can spread malicious queries among multiple different user accounts. Although this attack is feasible, the adversary needs to create $\frac{Budget}{num_s} = \frac{50000}{100}= 500$ accounts (email or phone number) to conduct model extraction attacks, which greatly increases the attack cost.
HODA can be complementary to perturbation-based defenses. For example, the size of perturbation can pertain to the value of Pearson distance between $H_n$ and $H_u$ for user $u$. }

%% file: content/interae.tex
\input{content/aehist.tex}

\input{content/newfigattackanalysisprob.tex}

Adversarial examples (AEs) are maliciously crafted inputs that cause the target classifier to misclassify them. 
\comment{Generally, adversarial examples are created by adding small, often imperceptible, perturbations to normal samples to force a classifier to predict incorrectly. Given that large perturbations can change the true class of samples, the magnitude of perturbation is often restricted by $L_p$-norms $\parallel . \parallel_p$.
Suppose that the label of sample $x$ is $y$,  adversarial example $x'$ can be formulated as follows:
\begin{equation}
\begin{split}
x' = x + \eta \quad s.t.  \quad f_t(x') = y',  \; y\ne y', \; \parallel\eta \parallel_p \le \epsilon, x' \in domain(x)
\end{split}
\end{equation}

}There are numerous methods to generate adversarial examples such as L-BFGS \cite{DBLP:journals/corr/SzegedyZSBEGF13}, FGSM \cite{DBLP:journals/corr/GoodfellowSS14},  C\&W \cite{DBLP:conf/sp/Carlini017},  PGD \cite{DBLP:conf/iclr/MadryMSTV18}, and AutoAttack (AA) \cite{DBLP:conf/icml/Croce020a}. 
To investigate the hardness of adversarial examples, we use FGSM ($\epsilon=0.1$), C\&W ($L_2$), PGD ($\epsilon=8/255, \alpha=3/255$), and AA ($L_\infty, \epsilon=8/255$) attacks to generate adversarial examples on CIFAR10 and CIFAR100 test sets. The adversarial examples are created in the white-box setting, and they are untargeted. Figure \ref{fig:aeatthardhist} indicates the hardness degree of adversarial examples on CIFAR10 and CIFAR100 target classifiers. Although the distance between normal samples and adversarial examples is very small, Figure \ref{fig:aeatthardhist} demonstrates that the hardness degree histograms of adversarial examples are very different from normal samples (Figure \ref{loginMock}). Most adversarial examples generated by FGSM and C\&W are harder than adversarial examples  generated by PGD and AA. We think this is because the size of perturbations added by FGSM and C\&W is larger than PGD and AA. An intriguing observation is that almost all adversarial examples being generated by PGD are not hard. AA has relatively the same behavior, and the number of its hard adversarial examples is very small.

%% file: content/aehist.tex
\begin{figure}
	\label{fig:aehardhistres}
	\begin{subfigure}{\linewidth}
		\centering
		\resizebox{\linewidth}{!}{
		\begin{tikzpicture}
		\begin{groupplot}[group style={group size= 4 by 2,vertical sep=0.9cm},height=4cm,width=4.8cm,tick label style = {font = {\fontsize{8 pt}{12 pt}\selectfont}},,ytick align=outside,xtick style={draw=none},tick label style={font=\scriptsize},,ytick pos=left,enlargelimits=true,enlargelimits=0.02,]
		\nextgroupplot[ymin=0, ymax=1100,title=FGSM,ylabel=\# of samples,ytick={0,200,400,600,800,1000},scaled y ticks = false]
		\addplot
		[ybar interval,mark=no,fill=purple] 
		coordinates
		{
(0,131)(1,121)(2,14)(3,15)(4,17)(5,98)(6,33)(7,4)(8,64)(9,2)(10,2)(11,8)(12,1)(13,3)(14,8)(15,2)(16,2)(17,13)(18,7)(19,20)(20,15)(21,1)(22,25)(23,13)(24,36)(25,19)(26,7)(27,16)(28,35)(29,12)(30,4)(31,29)(32,10)(33,16)(34,9)(35,75)(36,9)(37,13)(38,21)(39,45)(40,48)(41,21)(42,11)(43,22)(44,85)(45,21)(46,122)(47,63)(48,20)(49,115)(50,256)(51,72)(52,36)(53,55)(54,133)(55,71)(56,150)(57,187)(58,264)(59,167)(60,88)(61,24)(62,805)(63,324)(64,78)(65,111)(66,175)(67,99)(68,54)(69,144)(70,590)(71,39)(72,56)(73,613)(74,217)(75,42)(76,114)(77,150)(78,55)(79,83)(80,112)(81,190)(82,81)(83,173)(84,38)(85,29)(86,140)(87,11)(88,325)(89,76)(90,79)(91,44)(92,15)(93,105)(94,90)(95,244)(96,302)(97,249)(98,369)(99,573)(100,0)
		} 
		\closedcycle;
		\nextgroupplot[ymin=0, ymax=2500,title=C\&W,ytick={0,500,1000,1500,2000,2500},,scaled y ticks = false]
		\addplot
		[ybar interval,mark=no,fill=purple] 
		coordinates
		{
(0,320)(1,159)(2,111)(3,103)(4,46)(5,20)(6,85)(7,10)(8,9)(9,6)(10,7)(11,4)(12,6)(13,7)(14,10)(15,9)(16,2)(17,1)(18,5)(19,4)(20,6)(21,1)(22,4)(23,2)(24,3)(25,2)(26,5)(27,9)(28,3)(29,10)(30,3)(31,7)(32,14)(33,6)(34,16)(35,6)(36,6)(37,14)(38,24)(39,21)(40,22)(41,17)(42,13)(43,30)(44,26)(45,29)(46,59)(47,43)(48,32)(49,38)(50,47)(51,72)(52,53)(53,54)(54,73)(55,51)(56,85)(57,89)(58,104)(59,59)(60,105)(61,80)(62,111)(63,118)(64,102)(65,124)(66,103)(67,110)(68,160)(69,93)(70,117)(71,101)(72,63)(73,68)(74,119)(75,103)(76,120)(77,107)(78,74)(79,85)(80,93)(81,75)(82,176)(83,119)(84,115)(85,95)(86,115)(87,141)(88,104)(89,127)(90,129)(91,127)(92,91)(93,203)(94,171)(95,328)(96,342)(97,718)(98,835)(99,1751)(100,0)
		} 
		\closedcycle;
		\nextgroupplot[ymin=0,ymax=1000,title=PGD,,ytick={0,200,400,600,800,1000},scaled y ticks = false]
		\addplot
		[ybar interval,mark=no,fill=purple] 
		coordinates
		{
(0,28)(1,21)(2,11)(3,16)(4,33)(5,41)(6,47)(7,49)(8,34)(9,56)(10,75)(11,187)(12,171)(13,148)(14,165)(15,268)(16,122)(17,208)(18,448)(19,199)(20,559)(21,300)(22,439)(23,379)(24,204)(25,266)(26,214)(27,644)(28,172)(29,386)(30,152)(31,169)(32,276)(33,144)(34,369)(35,232)(36,202)(37,287)(38,356)(39,275)(40,174)(41,135)(42,67)(43,181)(44,78)(45,116)(46,231)(47,191)(48,46)(49,87)(50,34)(51,113)(52,43)(53,25)(54,30)(55,15)(56,19)(57,17)(58,20)(59,3)(60,7)(61,3)(62,3)(63,3)(64,2)(65,0)(66,1)(67,0)(68,0)(69,1)(70,0)(71,0)(72,0)(73,0)(74,1)(75,0)(76,1)(77,0)(78,0)(79,0)(80,0)(81,0)(82,0)(83,0)(84,0)(85,0)(86,0)(87,0)(88,0)(89,0)(90,0)(91,0)(92,0)(93,0)(94,0)(95,0)(96,0)(97,0)(98,0)(99,1)(100,0)
		} 
		\closedcycle;
		\nextgroupplot[ymin=0,ymax=1000,title=AA,,ytick={0,200,400,600,800,1000},scaled y ticks = false]
		\addplot
		[ybar interval,mark=no,fill=purple] 
		coordinates
		{
(0,12)(1,0)(2,3)(3,4)(4,17)(5,28)(6,24)(7,19)(8,18)(9,15)(10,50)(11,143)(12,105)(13,85)(14,117)(15,187)(16,90)(17,160)(18,413)(19,236)(20,762)(21,294)(22,393)(23,473)(24,310)(25,243)(26,237)(27,811)(28,313)(29,483)(30,153)(31,112)(32,298)(33,196)(34,562)(35,334)(36,225)(37,234)(38,449)(39,264)(40,94)(41,92)(42,31)(43,87)(44,41)(45,60)(46,127)(47,27)(48,21)(49,21)(50,9)(51,42)(52,14)(53,14)(54,11)(55,8)(56,12)(57,8)(58,11)(59,7)(60,6)(61,6)(62,9)(63,15)(64,10)(65,6)(66,11)(67,19)(68,17)(69,8)(70,12)(71,13)(72,2)(73,11)(74,11)(75,4)(76,14)(77,6)(78,7)(79,8)(80,4)(81,7)(82,6)(83,12)(84,7)(85,5)(86,3)(87,7)(88,2)(89,3)(90,4)(91,9)(92,4)(93,12)(94,6)(95,6)(96,8)(97,29)(98,28)(99,34)(100,0)
		} 
		\closedcycle;
		\end{groupplot}
		\end{tikzpicture}
	}
\vspace{-0.6cm}
		\caption{ CIFAR10}
	\end{subfigure}
	\begin{subfigure}{\linewidth}
		\centering
		\resizebox{\linewidth}{!}{
		\begin{tikzpicture}
		\begin{groupplot}[group style={group size= 4 by 1,vertical sep=0.9cm},height=4cm,width=4.8cm,tick label style = {font = {\fontsize{8 pt}{12 pt}\selectfont}},,ytick align=outside,xtick style={draw=none},tick label style={font=\scriptsize},,ytick pos=left,enlargelimits=true,enlargelimits=0.02,]
		\nextgroupplot[ymin=0, ymax=1100,ylabel=\# of samples,xlabel = Hardness Degree,,ytick={0,200,400,600,800,1000},scaled y ticks = false]
		\addplot
		[ybar interval,mark=no,fill=purple] 
		coordinates
		{
			
(0,0)(1,0)(2,0)(3,0)(4,0)(5,0)(6,0)(7,0)(8,0)(9,0)(10,0)(11,0)(12,1)(13,0)(14,1)(15,1)(16,1)(17,1)(18,0)(19,0)(20,0)(21,4)(22,0)(23,8)(24,2)(25,5)(26,6)(27,6)(28,4)(29,1)(30,2)(31,5)(32,6)(33,24)(34,11)(35,19)(36,19)(37,26)(38,60)(39,15)(40,58)(41,50)(42,52)(43,119)(44,56)(45,165)(46,69)(47,341)(48,303)(49,360)(50,204)(51,134)(52,261)(53,337)(54,247)(55,86)(56,146)(57,258)(58,62)(59,91)(60,467)(61,181)(62,125)(63,68)(64,115)(65,107)(66,185)(67,93)(68,140)(69,58)(70,43)(71,43)(72,114)(73,100)(74,90)(75,232)(76,98)(77,53)(78,64)(79,64)(80,105)(81,31)(82,83)(83,134)(84,76)(85,107)(86,66)(87,38)(88,104)(89,178)(90,290)(91,102)(92,101)(93,104)(94,171)(95,342)(96,229)(97,228)(98,370)(99,1074)(100,0)
		} 
		\closedcycle;
		\nextgroupplot[ymin=0, xlabel =Hardness Degree, ymax=2500,ytick={0,500,1000,1500,2000,2500},scaled y ticks = false]
		\addplot
		[ybar interval,mark=no,fill=purple] 
		coordinates
		{
(0,38)(1,27)(2,33)(3,46)(4,12)(5,3)(6,7)(7,1)(8,3)(9,8)(10,2)(11,5)(12,2)(13,1)(14,3)(15,3)(16,6)(17,1)(18,4)(19,9)(20,2)(21,5)(22,2)(23,7)(24,4)(25,8)(26,6)(27,7)(28,15)(29,9)(30,8)(31,5)(32,11)(33,19)(34,11)(35,20)(36,25)(37,24)(38,36)(39,28)(40,24)(41,34)(42,60)(43,54)(44,45)(45,53)(46,91)(47,101)(48,72)(49,78)(50,85)(51,83)(52,88)(53,93)(54,106)(55,84)(56,66)(57,78)(58,64)(59,83)(60,70)(61,98)(62,74)(63,61)(64,56)(65,66)(66,84)(67,81)(68,60)(69,84)(70,69)(71,58)(72,60)(73,63)(74,64)(75,58)(76,124)(77,67)(78,57)(79,66)(80,106)(81,78)(82,59)(83,81)(84,108)(85,102)(86,96)(87,94)(88,73)(89,152)(90,116)(91,144)(92,168)(93,201)(94,269)(95,373)(96,437)(97,482)(98,1053)(99,2480)(100,0)
		} 
		\closedcycle;
		\nextgroupplot[ymin=0,ymax=1000,xlabel style={align=center},xlabel = Hardness Degree ,ytick={0,200,400,600,800,1000},scaled y ticks = false]
		\addplot
		[ybar interval,mark=no,fill=purple] 
		coordinates
		{	(0,9)(1,19)(2,46)(3,102)(4,120)(5,153)(6,197)(7,278)(8,228)(9,619)(10,364)(11,412)(12,439)(13,381)(14,273)(15,428)(16,564)(17,438)(18,235)(19,602)(20,367)(21,400)(22,283)(23,418)(24,218)(25,376)(26,246)(27,219)(28,207)(29,160)(30,144)(31,134)(32,134)(33,306)(34,75)(35,60)(36,52)(37,76)(38,79)(39,43)(40,25)(41,15)(42,21)(43,5)(44,7)(45,6)(46,6)(47,1)(48,5)(49,1)(50,0)(51,0)(52,0)(53,0)(54,1)(55,0)(56,0)(57,0)(58,0)(59,1)(60,0)(61,0)(62,0)(63,0)(64,0)(65,0)(66,0)(67,1)(68,0)(69,0)(70,0)(71,0)(72,0)(73,0)(74,0)(75,0)(76,0)(77,0)(78,0)(79,0)(80,0)(81,0)(82,0)(83,0)(84,0)(85,0)(86,0)(87,0)(88,0)(89,0)(90,0)(91,0)(92,0)(93,0)(94,0)(95,0)(96,0)(97,0)(98,0)(99,1)(100,0)
		};
		\nextgroupplot[ymin=0,ymax=1000,xlabel = Hardness Degree,,ytick={0,200,400,600,800,1000},scaled y ticks = false]
		\addplot
		[ybar interval,mark=no,fill=purple] 
		coordinates
		{
(0,6)(1,7)(2,25)(3,50)(4,80)(5,141)(6,147)(7,262)(8,181)(9,696)(10,434)(11,398)(12,427)(13,399)(14,205)(15,301)(16,462)(17,394)(18,228)(19,648)(20,242)(21,239)(22,89)(23,408)(24,150)(25,221)(26,182)(27,117)(28,110)(29,81)(30,46)(31,71)(32,51)(33,141)(34,32)(35,32)(36,30)(37,34)(38,48)(39,31)(40,28)(41,32)(42,44)(43,38)(44,32)(45,41)(46,62)(47,59)(48,44)(49,51)(50,51)(51,48)(52,52)(53,45)(54,53)(55,49)(56,34)(57,38)(58,27)(59,42)(60,31)(61,31)(62,36)(63,22)(64,28)(65,24)(66,31)(67,23)(68,14)(69,24)(70,19)(71,11)(72,22)(73,23)(74,11)(75,12)(76,29)(77,16)(78,15)(79,16)(80,36)(81,15)(82,18)(83,24)(84,19)(85,33)(86,19)(87,25)(88,14)(89,22)(90,15)(91,34)(92,37)(93,21)(94,49)(95,41)(96,73)(97,47)(98,109)(99,195)(100,0)
		};
		\end{groupplot}
		\end{tikzpicture}
	}
\vspace{-0.6cm}
		\caption{ CIFAR100}
	\end{subfigure}
\vspace{-0.6\baselineskip}
	\caption{The hardness degree histograms of samples of four various adversarial example attacks on CIFAR10 and CIFAR100 target classifiers. Each attack uses 10000 natural samples in the test set associated with the target classifier dataset to create 10000 adversarial examples.  }
	\label{fig:aeatthardhist}
\end{figure}
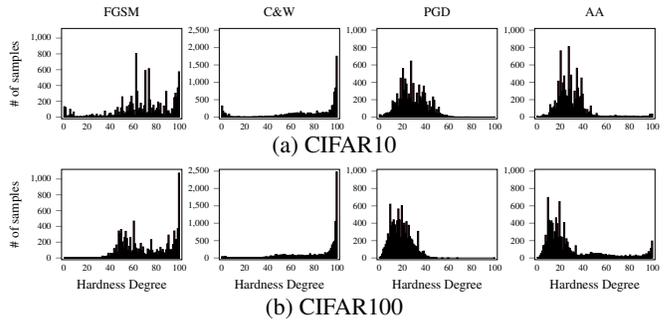

%% file: content/newfigattackanalysisprob.tex
\begin{figure*}[!t]
	\label{fig:analysis}
		\centering
\resizebox{\linewidth}{!}{
	\begin{tikzpicture}
	\begin{groupplot}[group style={group size= 4 by 1,vertical sep=1.2cm},height=4cm,width=4.8cm,tick label style = {font = {\fontsize{7pt}{12 pt}\selectfont}},
	grid,
	grid style={black!5},]
	\nextgroupplot[
	ytick={0,20,40,60,80,100},
	xtick={1,2,3,4,5,6,7,8,9,10},
	ymin = 0,
	title= \small{CIFAR10},
	ylabel = \small{Acc(\%)},
	ylabel shift = -8 pt,
	xlabel = \small{Hardness Group},
	]
	\addplot[cyan, mark = *, mark size= 0.5pt] coordinates { 
(1,99.88)
(2,99.55)
(3,99.29)
(4,96.35)
(5,94.73)
(6,87.1)
(7,76.48)
(8,63.33)
(9,63.64)
(10,50.88)
};
\addplot[blue, mark = diamond, mark size= 0.5pt] coordinates { 
		(1,96.43)
(2,86.82)
(3,79.7)
(4,71.2)
(5,64.65)
(6,57.92)
(7,44.18)
(8,41.25)
(9,37.01)
(10,27.72)
};

	\addplot[red, mark = triangle, mark size= 0.5pt] coordinates { 
	(1,94.83)
	(2,84.21)
	(3,76.94)
	(4,68.27)
	(5,68.64)
	(6,55.28)
	(7,47.91)
	(8,44.17)
	(9,31.17)
	(10,36.84)
};

	\addplot[green, mark = square*, mark size= 0.5pt] coordinates { 
(1,53.62)
(2,38.04)
(3,33.9)
(4,33.85)
(5,29.48)
(6,26.78)
(7,26.61)
(8,25.0)
(9,20.42)
(10,23.88)

};

	\addplot[brown, mark = |, mark size= 0.5pt] coordinates { 
	(1,55.0)
	(2,41.54)
	(3,39.17)
	(4,35.55)
	(5,31.36)
	(6,30.19)
	(7,29.36)
	(8,29.02)
	(9,20.42)
	(10,22.39)
};

\addplot[dashed,blue, mark = diamond, mark size= 0.5pt] coordinates { 
	(1,96.41)
	(2,86.82)
	(3,79.7)
	(4,70.47)
	(5,64.01)
	(6,54.55)
	(7,38.24)
	(8,29.17)
	(9,23.38)
	(10,15.79)
};
\addplot[dashed,red, mark = triangle, mark size= 0.5pt] coordinates { 
	(1,94.81)
	(2,84.14)
	(3,76.85)
	(4,67.69)
	(5,67.99)
	(6,51.32)
	(7,40.88)
	(8,32.92)
	(9,24.03)
	(10,20.7)
};

\addplot[dashed,green, mark = square*, mark size= 0.5pt] coordinates { 
	(1,53.57)
	(2,37.96)
	(3,33.8)
	(4,32.77)
	(5,28.53)
	(6,23.68)
	(7,21.1)
	(8,16.07)
	(9,13.38)
	(10,14.18)
};

\addplot[dashed,brown, mark = |, mark size= 0.5pt] coordinates { 
	(1,55.0)
	(2,41.31)
	(3,38.98)
	(4,34.31)
	(5,30.15)
	(6,28.17)
	(7,22.71)
	(8,20.09)
	(9,12.68)
	(10,12.69)
};

\nextgroupplot[
ytick={0,20,40,60,80,100},
xtick={1,2,3,4,5,6,7,8,9,10},
ymin = 0,
title= CIFAR10,
	ylabel = Fid(\%),
	ylabel shift = -8 pt,
		xlabel = \small{Hardness Group},
]

\addplot[blue, mark = diamond, mark size= 0.5pt] coordinates { 
(1,96.5)
(2,87.27)
(3,80.41)
(4,72.95)
(5,68.25)
(6,62.02)
(7,50.33)
(8,47.08)
(9,35.71)
(10,36.49)
};

\addplot[red, mark = triangle, mark size= 0.5pt] coordinates { 
(1,94.9)
(2,84.51)
(3,77.29)
(4,70.18)
(5,71.85)
(6,58.06)
(7,52.31)
(8,50.83)
(9,41.56)
(10,41.05)
};

\addplot[green, mark = square*, mark size= 0.5pt] coordinates { 
(1,53.59)
(2,38.35)
(3,33.99)
(4,33.85)
(5,30.15)
(6,27.09)
(7,27.06)
(8,26.34)
(9,20.42)
(10,23.88)
};

\addplot[brown, mark = |, mark size= 0.5pt] coordinates { 
(1,55.08)
(2,41.54)
(3,39.36)
(4,35.7)
(5,32.03)
(6,32.51)
(7,27.98)
(8,30.36)
(9,23.24)
(10,23.13)
};

		\nextgroupplot[
ytick={0,20,40,60,80,100},
xtick={1,2,3,4,5,6,7,8,9,10},
	ymin = 0,
	title= CIFAR100,
	ylabel = Acc(\%),
	ylabel shift = -8 pt,
		legend style={at={(-0.2,-0.40)},
		anchor=north,legend columns=-6},
	xlabel = \small{Hardness Group},
	]
	\addlegendimage{cyan, mark = *, mark size= 0.5pt}
	\addlegendimage{green, mark = square*, mark size= 0.5pt}
	\addlegendimage{brown, mark = |, mark size= 0.5pt}
	\addlegendimage{blue, mark = diamond, mark size= 0.5pt}
	\addlegendimage{red, mark = triangle, mark size= 0.5pt}	\addlegendimage{dashed, mark size= 0.5pt}
	\addplot[cyan, mark = *, mark size= 0.5pt] coordinates { 
(1,99.37)
(2,98.21)
(3,95.33)
(4,90.46)
(5,75.53)
(6,57.58)
(7,48.03)
(8,44.76)
(9,34.97)
(10,30.3)
	};

	\addplot[red, mark = triangle, mark size= 0.5pt] coordinates { 
(1,95.94)
(2,88.38)
(3,78.53)
(4,64.93)
(5,48.06)
(6,35.27)
(7,33.07)
(8,27.3)
(9,28.32)
(10,26.26)
	};

\addplot[dashed,red, mark = triangle, mark size= 0.5pt] coordinates { 
	(1,95.94)
	(2,88.38)
	(3,78.13)
	(4,63.7)
	(5,44.31)
	(6,28.12)
	(7,20.47)
	(8,16.83)
	(9,13.58)
	(10,11.22)
};

\addplot[blue, mark = diamond, mark size=0.5pt] coordinates { 
(1,85.31)
(2,77.43)
(3,62.72)
(4,53.77)
(5,40.73)
(6,27.92)
(7,27.36)
(8,24.13)
(9,23.99)
(10,19.98)
};

\addplot[green, mark = square*, mark size= 0.5pt] coordinates { 
(1,36.21)
(2,18.01)
(3,17.31)
(4,14.22)
(5,12.27)
(6,8.74)
(7,10.79)
(8,7.75)
(9,4.89)
(10,7.06)

};

\addplot[brown, mark = |, mark size= 0.5pt] coordinates { 
(1,38.96)
(2,24.97)
(3,18.87)
(4,16.31)
(5,14.33)
(6,10.34)
(7,11.89)
(8,7.39)
(9,8.14)
(10,7.93)
};

\addplot[dashed,blue, mark = diamond, mark size= 0.5pt] coordinates { 
	(1,85.25)
	(2,77.32)
	(3,62.23)
	(4,52.74)
	(5,36.68)
	(6,21.06)
	(7,17.52)
	(8,14.29)
	(9,10.4)
	(10,8.75)
};

\addplot[dashed,green, mark = square*, mark size= 0.5pt] coordinates { 
	(1,36.21)
	(2,17.89)
	(3,16.98)
	(4,13.29)
	(5,10.72)
	(6,5.44)
	(7,5.95)
	(8,3.52)
	(9,1.3)
	(10,2.6)
};

\addplot[dashed,brown, mark = |, mark size= 0.5pt] coordinates { 
	(1,38.9)
	(2,24.72)
	(3,18.42)
	(4,15.3)
	(5,12.2)
	(6,6.82)
	(7,5.95)
	(8,4.58)
	(9,2.61)
	(10,2.85)
};

\legend{  \footnotesize{$f_t$}, \footnotesize{$f_s$: JBDA}, \footnotesize{$f_s$: JBRAND},\footnotesize{ $f_s$: K.Net CIFARX}, \footnotesize{$f_s$: K.Net TIN},\footnotesize{Correctly classified by both $f_t$ and $f_s$ } }

\nextgroupplot[
ytick={0,20,40,60,80,100},
xtick={1,2,3,4,5,6,7,8,9,10},
ymin = 0,
title= CIFAR100,
ylabel = Fid(\%),
ylabel shift = -8 pt,
xlabel = \small{Hardness Group},
]
\addplot[red, mark = triangle, mark size= 0.5pt] coordinates { 
(1,96.57)
(2,90.17)
(3,81.71)
(4,69.57)
(5,56.46)
(6,43.29)
(7,37.01)
(8,38.41)
(9,31.21)
(10,29.85)
};

\addplot[blue, mark = diamond, mark size= 0.5pt] coordinates { 
(1,85.76)
(2,78.77)
(3,65.51)
(4,57.12)
(5,46.24)
(6,35.27)
(7,31.89)
(8,31.11)
(9,24.86)
(10,23.57)
};

\addplot[green, mark = square*, mark size= 0.5pt] coordinates { 
(1,36.4)
(2,18.39)
(3,17.87)
(4,14.3)
(5,13.49)
(6,9.06)
(7,9.03)
(8,8.8)
(9,5.86)
(10,8.18)
};

\addplot[brown, mark = |, mark size= 0.5pt] coordinates { 
(1,39.09)
(2,25.09)
(3,19.31)
(4,16.38)
(5,14.98)
(6,10.87)
(7,10.35)
(8,10.21)
(9,6.84)
(10,9.42)
};

	\end{groupplot}
\end{tikzpicture}
}
\vspace{-0.7\baselineskip}
\caption{The accuracy and the fidelity of four surrogate classifiers over various hardness groups. The test set of each dataset is partitioned into 10 hardness groups so that hardness group 1 consists of the easiest samples and hardness group 10 consists of the hardest samples. The dashed lines indicate the percentage of samples being correctly classified by both target classifier $f_t$ and surrogate classifier $f_s$.}
\label{fig:hardnessanalysisprobacccfida}
\end{figure*}
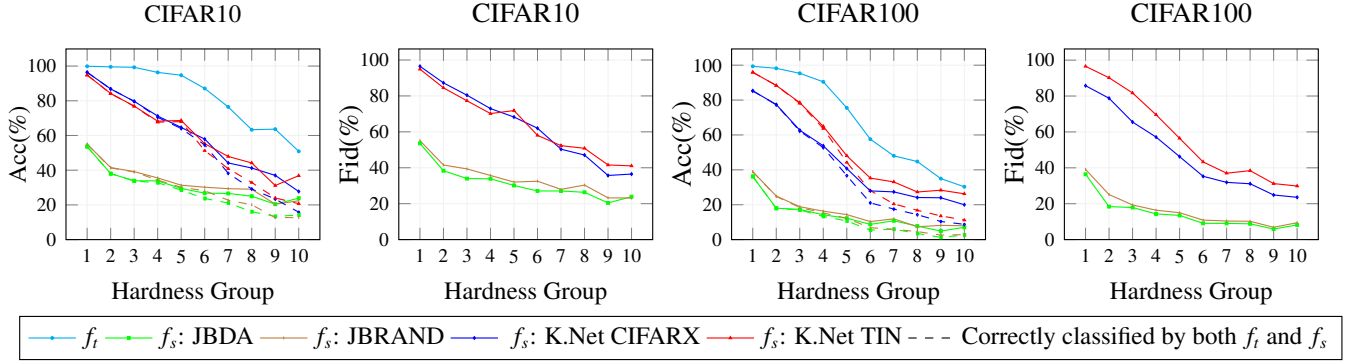

%% file: content/hardnessanalysisofattacks.tex
To give new insight into model extraction attacks, we investigate the performance of model extraction attacks on normal samples with various levels of hardness. For this purpose, the test sets of CIFAR10 and CIFAR100 datasets are partitioned into 10 hardness groups based on hardness degree of samples.
The hardness group $i$ consists of samples that their hardness degree is in range $[i\times10, (i+1)\times10]$. Hence, the first hardness group consists of the easiest samples, and the last hardness group consists of the hardest ones.
Figure \ref{fig:hardnessanalysisprobacccfida} shows the accuracy and the fidelity of attacks over 10 hardness groups when the output of target classifier is the entire probability vector. The results demonstrate that the accuracy and the fidelity of all attacks are decreased as the hardness of samples is increased. We know from Figure \ref{fig:trust} that the accuracy of target classifiers is decreased by increasing the hardness of samples. Figure \ref{fig:hardnessanalysisprobacccfida} indicates the surrogate classifiers also have the same behavior. 

The results demonstrate that the distance between the accuracy of target classifiers and surrogate classifiers (specially K.Net attacks) is increased by increasing the hardness of samples in the first hardness groups. However, the accuracy of the K.Net surrogate classifiers approaches the accuracy of target classifiers on the last two hardness groups. To investigate this observation, Figure \ref{fig:hardnessanalysisprobacccfida} shows the percentage of samples being classified correctly by both surrogate classifier and target classifier for all attacks over various hardness groups with dashed lines. The results indicate that all samples correctly classified by surrogate classifiers are also correctly classified by target classifiers in the first two hardness groups. However, by increasing the hardness of samples, the surrogate classifiers correctly classify some samples that are not correctly classified by the target classifier, and the number of such samples is increased by increasing the hardness of samples. \cite{DBLP:conf/uss/JagielskiCBKP20} demonstrate that labels from the target classifier are more informative than dataset labels. We think the information in the labels that come from the target model causes the surrogate classifiers to correctly classify hard samples that are not correctly classified by the target classifier.

The fidelity of all surrogate classifiers is decreased by increasing the hardness of samples, which means that the disagreement among surrogate classifiers and target classifiers is raised on harder samples. An intriguing observation is that the fidelity of surrogate classifiers to the target classifiers on correctly classified samples by target classifiers is much more than wrongly classified samples.

%% file: content/rw.tex
In the following, we briefly review the most prominent model extraction attacks and defenses presented so far.

\subsection{Model Extraction Attacks}

For the first time, \cite{10.1145/1081870.1081950} demonstrate the possibility of stealing simple linear machine learning models through only interaction with them. \cite{DBLP:conf/uss/TramerZJRR16} show the feasibility of model extraction attacks on commercial MLaaS. \cite{10.1145/3052973.3053009} and \cite{DBLP:conf/eurosp/JuutiSMA19} investigate stealing DNN-based classifiers and propose jacobian-based model extraction attacks for creating a surrogate classifier in order to generate adversarial examples in the black-box setting. \cite{DBLP:conf/uss/ChandrasekaranC20} explore the connection between active learning and model extraction attacks. They implement two query synthesis active learning algorithms to extract machine learning models, such as decision trees. \cite{DBLP:conf/uss/JagielskiCBKP20} use semi-supervised learning methods to improve the performance of model extraction attacks.
Knockoff Net \cite{DBLP:conf/cvpr/OrekondySF19}, ActiveThief \cite{pal2019framework}, and Copycat CNN \cite{DBLP:conf/ijcnn/SilvaBBSO18} use a semantically similar dataset to the target classifier's training set to create the surrogate classifier's training set. They employ different strategies for selecting samples from attack datasets to extract more information from the target classifier. \cite{DBLP:conf/ndss/YuYZTHJ20} employ active learning, transfer learning, and a new method for generating adversarial examples to improve model extraction attacks efficiency.
 A line of studies \cite{Truong_2021_CVPR,kariyappa2020maze,NEURIPS2020_e8d66338} use synthetic data to create the training set of surrogate classifiers. Although their methods do not need to have access to natural samples, they send a high number of queries to the target classifier, which makes their methods impractical. For example, \cite{Truong_2021_CVPR} and \cite{kariyappa2020maze} send millions of queries to extract a CIFAR10 target classifier. While most model extraction attacks have focused on the vulnerabilities of image classifiers, recent studies demonstrate the vulnerability of NLP \cite{DBLP:conf/iclr/KrishnaTPPI20}, Graph DNN \cite{263820}, and Reinforcement learning  \cite{10.1145/3433210.3453090} models against model extraction attacks. Another type of model extraction attacks use hardware side-channel vulnerabilities to extract a target classifier \cite{263804,236204,hong2020security,244042}. However, these attacks have a very strong threat model and suppose that the adversary has access to the hardware that hosts the target classifier.

\subsection{Defenses against Model Extraction Attacks}

Existing defense methods against model extraction attacks generally distribute into two branches: perturbation-based and detection-based defenses. 
Perturbation-based defenses \cite{DBLP:conf/sp/LeeEMS19,DBLP:conf/iclr/OrekondySF20,DBLP:conf/cvpr/KariyappaQ20} attempt to prevent adversaries from producing high-quality surrogate classifiers by adding perturbation to the output of target classifier. These methods generate the perturbation with various strategies to minimize the accuracy of surrogate classifiers. Recently, \cite{kariyappa2021protecting} proposed a new defense with the same goal as perturbation-based defenses, which does not perturb the output of target classifiers. Their approach employs an ensemble of diverse models to produce discontinuous predictions for out-of-distribution samples.
\comment{Maximizing Angular Deviation (MAD) method perturbs the output of target classifier in order to deviate the surrogate classifier's gradient signal. Ali et al. propose methods to modify the output of target classifiers for out-of-distribution samples.}
Proposed detection-based defenses \cite{10.1145/3274694.3274740,DBLP:conf/eurosp/JuutiSMA19} attempt to detect the occurrence of model extraction attacks by observing successive input queries to the target classifier. \cite{10.1145/3274694.3274740} propose a method to measure adversary perceived knowledge from target classifier, but this method only works for Decision Tree models. PRADA \cite{DBLP:conf/eurosp/JuutiSMA19} is the first proposed detection-based defense for DNN models. PRADA uses the histogram of the minimum $L_2$ distance among a new sample and all previous samples to detect model extraction attacks. Aside from its high computational overhead, it has been shown that PRADA is unable to detect model extraction attacks when an adversary uses natural samples \cite{pal2019framework}. 
Watermarking neural networks \cite{272262,10.1145/3196494.3196550,szyller2020dawn,217591} is another type of defense against model extraction attacks. These methods prove ownership of a surrogate classifier instead of preventing the occurrence of model extraction attacks.

\citet{10.1007/978-3-030-62144-5_4} demonstrate that several OOD detection approaches, such as Baseline \cite{DBLP:conf/iclr/HendrycksG17} and ODIN \cite{DBLP:conf/iclr/LiangLS18}, have poor performance in detecting Knockoff Net attack samples. Hence, they propose a new OOD detection approach that leverages a classifier to detect OOD samples. However, their approach only rejects OOD  samples, and it does not have any detection mechanism to detect adversaries. Besides, the OOD detector is trained on samples from the same distribution used by the adversary to conduct Knockoff Net attacks, which is an unrealistic assumption in practice.  SEAT \cite{10.1145/3474369.3486863} aims to detect model extraction attacks that use several similar samples to extract a target model, such as jacobian-based attacks \cite{10.1145/3052973.3053009,DBLP:conf/eurosp/JuutiSMA19}. Hence,  SEAT is ineffective when an adversary uses natural samples that are not similar to each other, such as Knockoff Net attack. 
VarDetect \cite{DBLP:journals/corr/abs-2107-05166} uses Variational Autoencoders (VAs) and Maximum Mean Discrepancy (MMD) to detect model extraction attacks. 